\documentclass[runningheads]{llncs}
\usepackage{graphicx}

\usepackage{tikz}
\usepackage{comment}
\usepackage{amsmath,amssymb} 
\usepackage{color}

\usepackage{booktabs}
\usepackage{multirow,bigstrut}
\usepackage{tabu}
\usepackage{makecell}
\usepackage{xspace}
\usepackage{lmodern}

\usepackage[british,american]{babel}

\usepackage[pagebackref=true,breaklinks=true,colorlinks,bookmarks=false,citecolor=citecolor, linkcolor=linkcolor]{hyperref}
\definecolor{citecolor}{HTML}{0071BC}
\definecolor{linkcolor}{HTML}{ED1C24}

\newcommand{\ours}{OpenSeg}
\newcommand{\imiou}{Grounding mIoU} 
\newcommand{\plus}[1]{
{\begin{scriptsize}\textcolor{blue}{(+#1)}\end{scriptsize}}
}
\newcommand{\minus}[1]{
{\begin{scriptsize}\textcolor{red}{(-#1)}\end{scriptsize}}
}

\newcommand{\pasc}{\texttt{PC-59}} 
\newcommand{\pascfull}{\texttt{PC-459}} 
\newcommand{\ade}{\texttt{A-150}} 
\newcommand{\adefull}{\texttt{A-847}} 

\usepackage{cuted}
\usepackage{capt-of}

\usepackage{pifont}
\newcommand{\cmark}{\ding{51}}%
\newcommand{\xmark}{\ding{55}}%

\DeclareMathOperator*{\argmax}{argmax} 

\newcommand{\supertiny}{\fontsize{4}{2}\selectfont}

\makeatletter
\DeclareRobustCommand\onedot{\futurelet\@let@token\@onedot}
\def\@onedot{\ifx\@let@token.\else.\null\fi\xspace}

\def\eg{\emph{e.g}\onedot} 
\def\ie{\emph{i.e}\onedot} 
 
\def\etc{\emph{etc}\onedot} \def\vs{\emph{vs}\onedot}
 
\def\etal{\emph{et al}\onedot}
\makeatother

\makeatletter\renewcommand\paragraph{\@startsection{paragraph}{4}{\z@}
  {.2em \@plus.0ex \@minus.2ex}{-.5em}{\normalfont\normalsize\bfseries}}\makeatother

\begin{document}
\pagestyle{headings}
\mainmatter
\def\ECCVSubNumber{5240}  

\title{Scaling Open-Vocabulary Image Segmentation with Image-Level Labels} 

\titlerunning{Scaling Open-Vocabulary Image Segmentation with Image-Level Labels}
%
%
\authorrunning{G. Ghiasi et al.}
%
\author{
Golnaz Ghiasi \and Xiuye Gu \and Yin Cui \and
Tsung-Yi Lin\thanks{Work done while at Google.}
}
\institute{Google Research \\ 
\email{\{golnazg, xiuyegu, yincui\}@google.com tsungyil@nvidia.com}}
\maketitle

\begin{abstract}
We design an open-vocabulary image segmentation model to organize an image into meaningful regions indicated by arbitrary texts.
Recent works (CLIP and ALIGN), despite attaining impressive open-vocabulary classification accuracy with image-level caption labels, are unable to segment visual concepts with pixels.
We argue that these models miss an important step of visual grouping, which organizes pixels into groups before learning visual-semantic alignments.
We propose OpenSeg to address the above issue while still making use of scalable image-level supervision of captions.
First, it learns to propose segmentation masks for possible organizations.
Then it learns visual-semantic alignments by aligning each word in a caption to one or a few predicted masks.
We find the mask representations are the key to support learning image segmentation from captions, making it possible to scale up the dataset and vocabulary sizes.  OpenSeg significantly outperforms the recent open-vocabulary method of LSeg by +19.9 mIoU on PASCAL dataset, thanks to its scalability.
\end{abstract}

\begingroup
\setlength\tabcolsep{0.5pt}
\renewcommand{\arraystretch}{0.1}
\centering%
\begin{tabular}{c@{\hskip 0.1in}rcrc}%
\centering%
\includegraphics[height=0.17\linewidth]{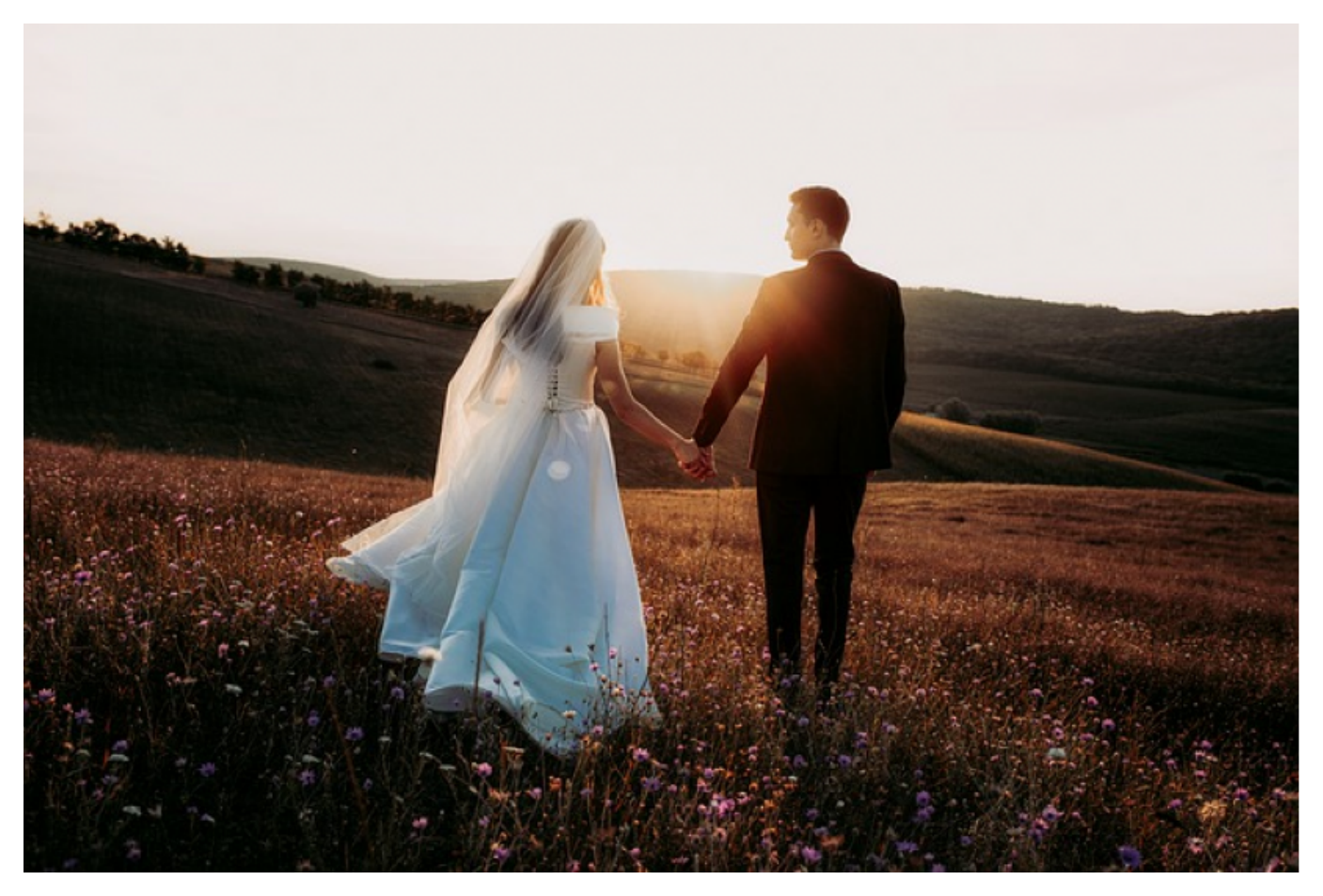}&%
\includegraphics[height=0.17\linewidth]{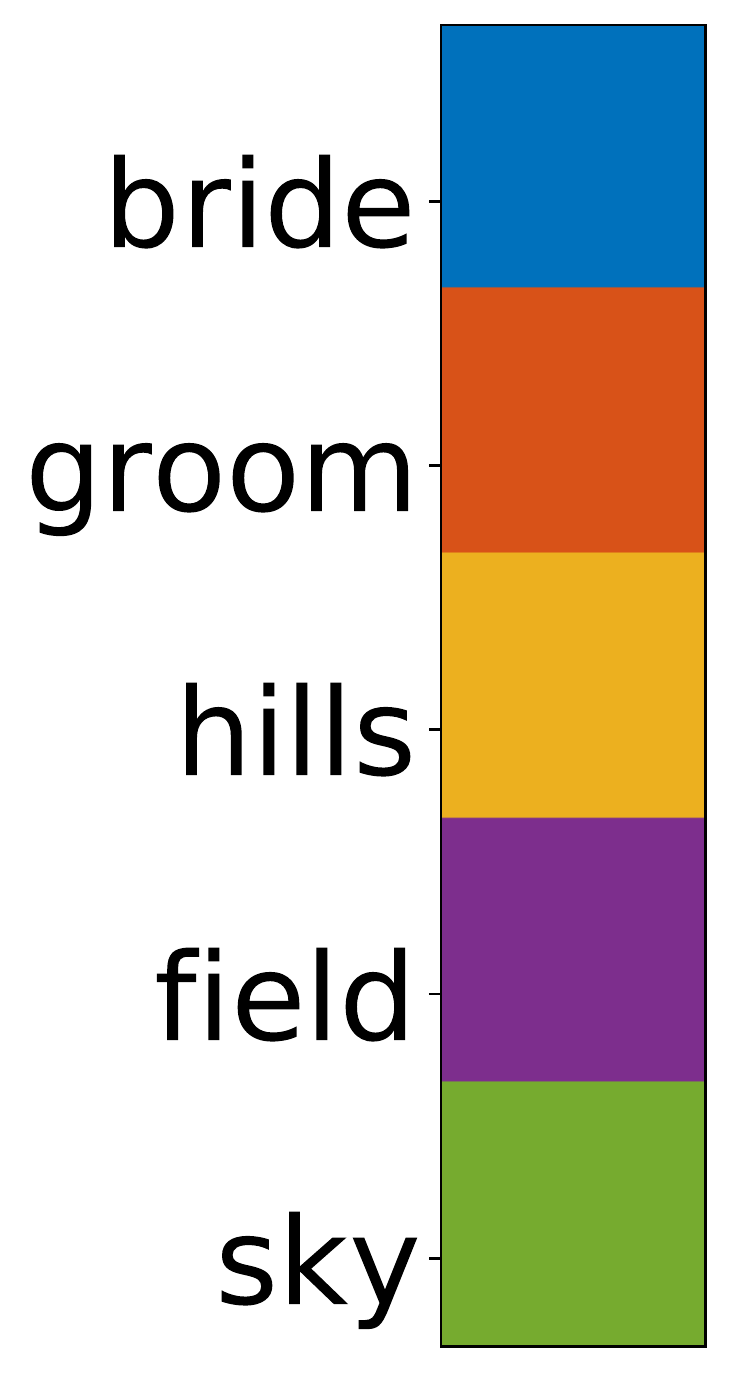}&%
\includegraphics[height=0.17\linewidth]{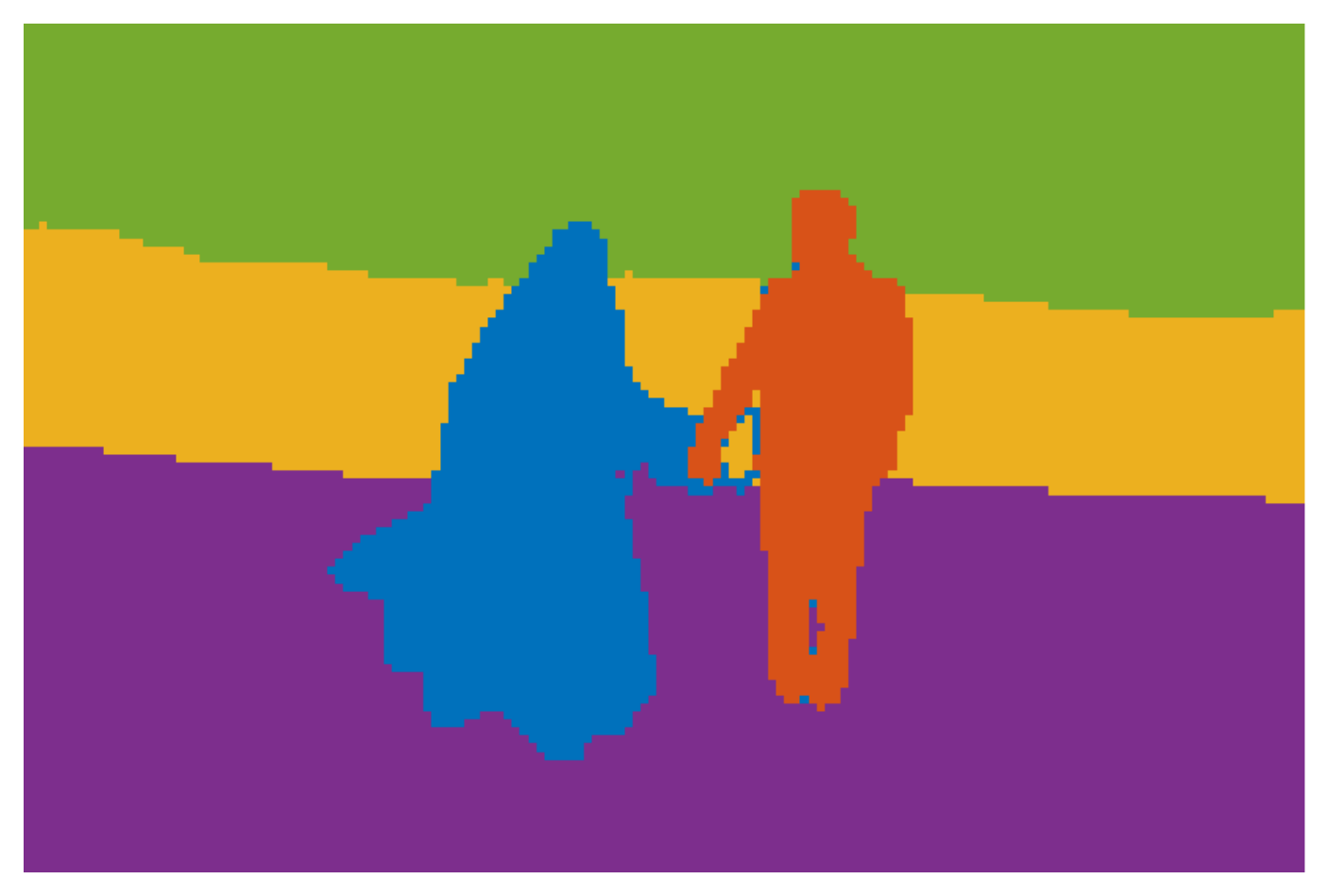}&%
\includegraphics[height=0.17\linewidth]{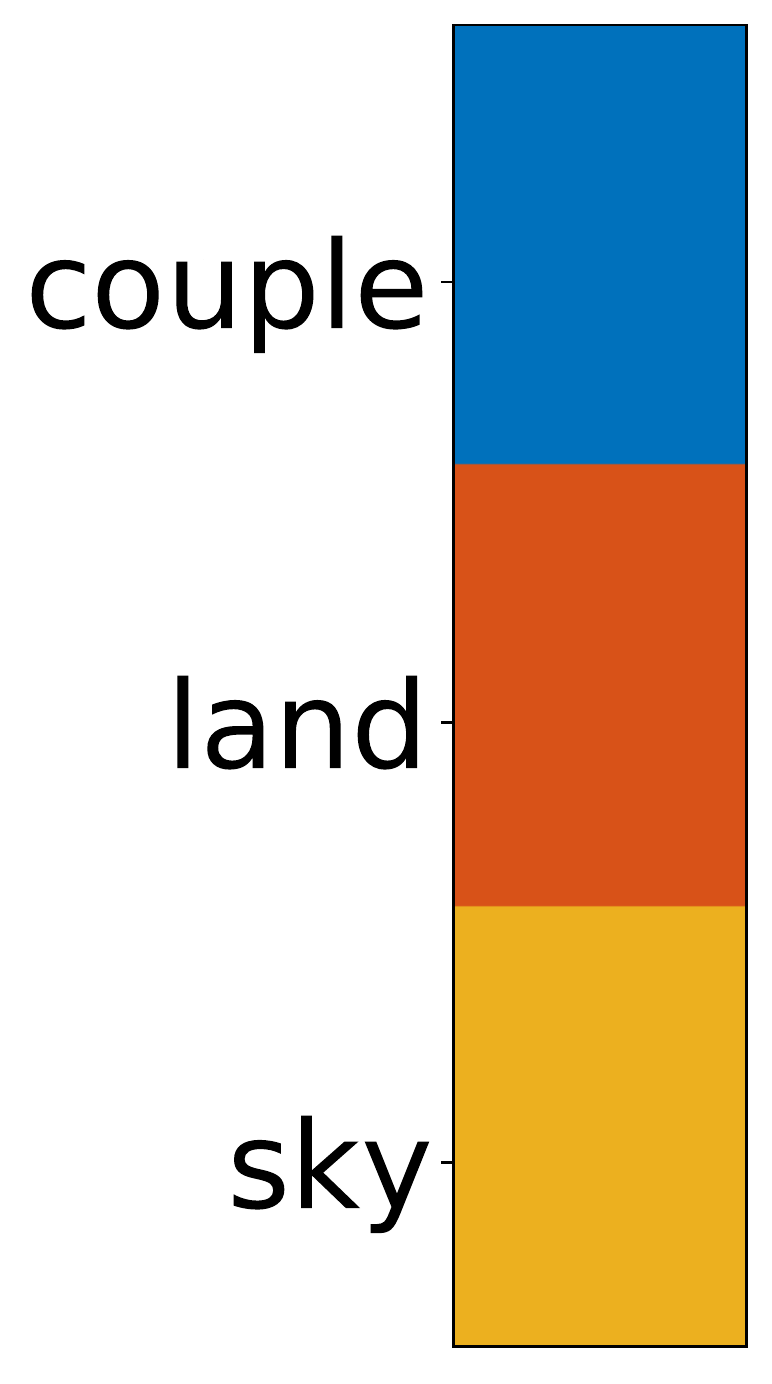}&%
\includegraphics[height=0.17\linewidth]{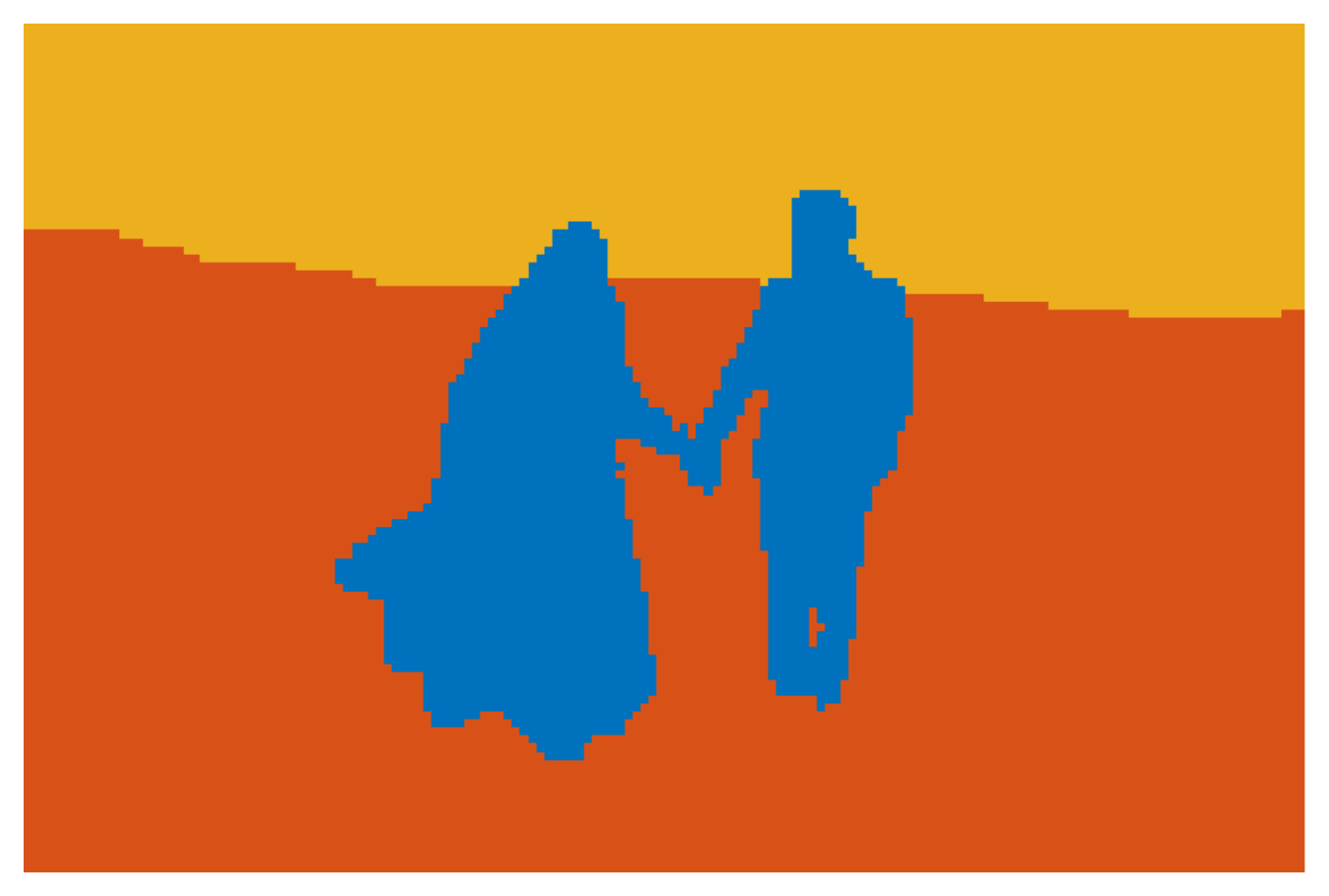}\\%
\includegraphics[height=0.172\linewidth]{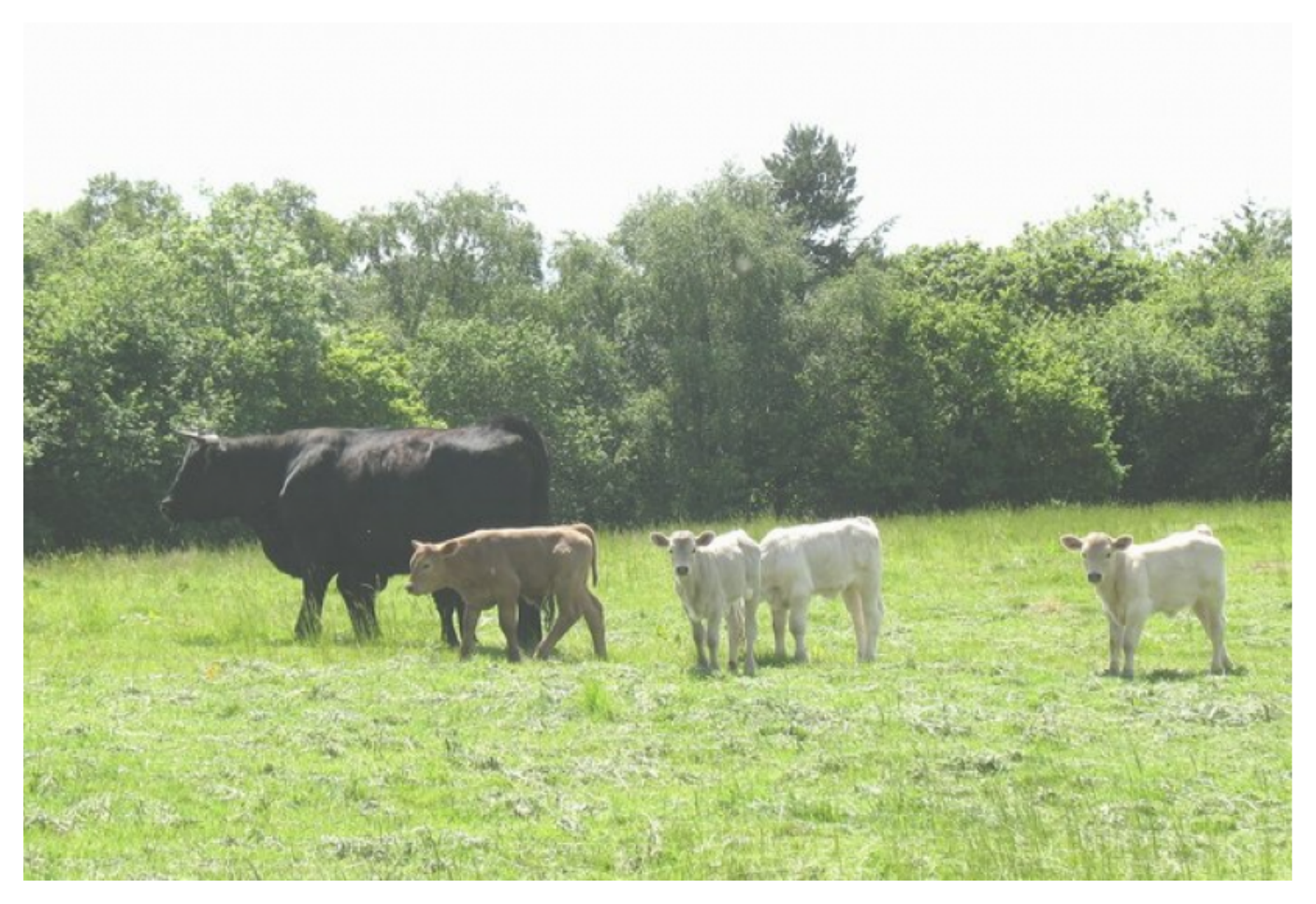}&%
\includegraphics[height=0.17\linewidth]{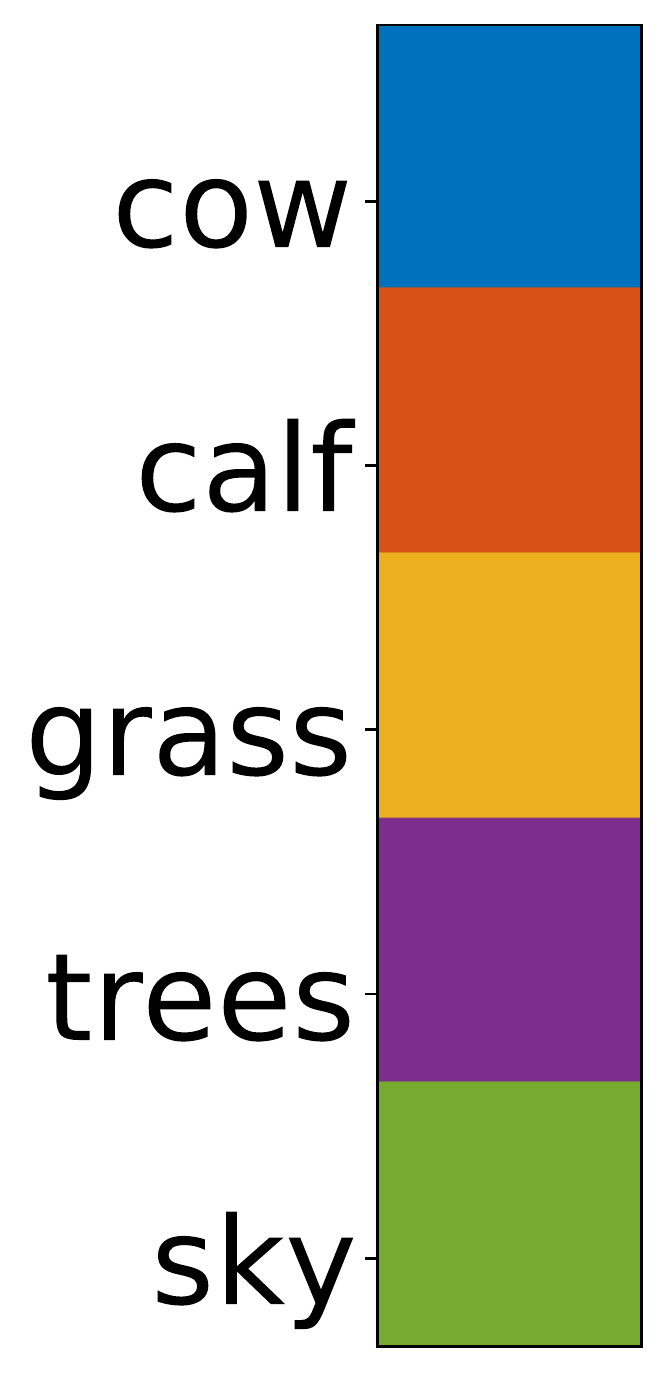}&%
\includegraphics[height=0.172\linewidth]{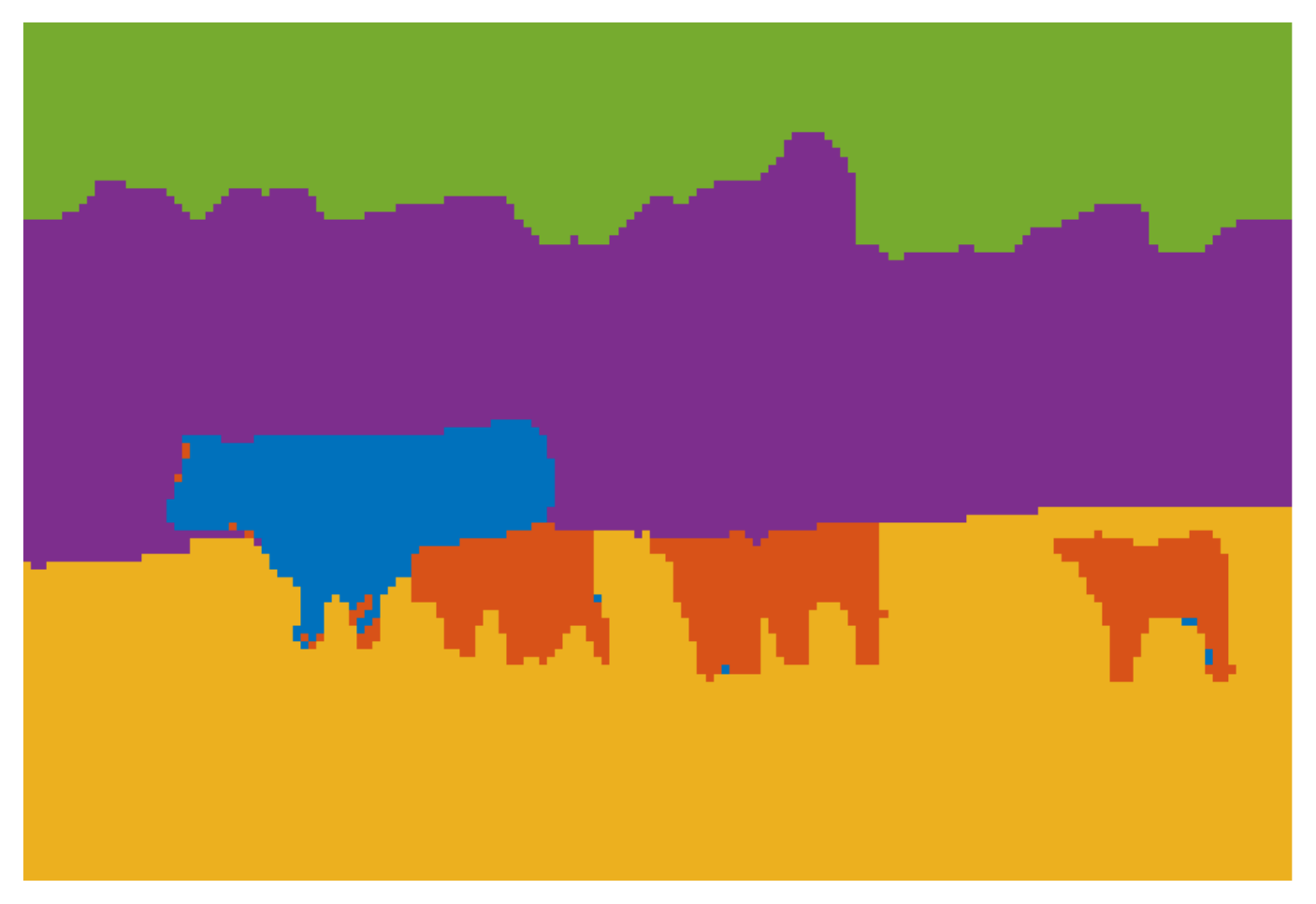}&%
\includegraphics[height=0.17\linewidth]{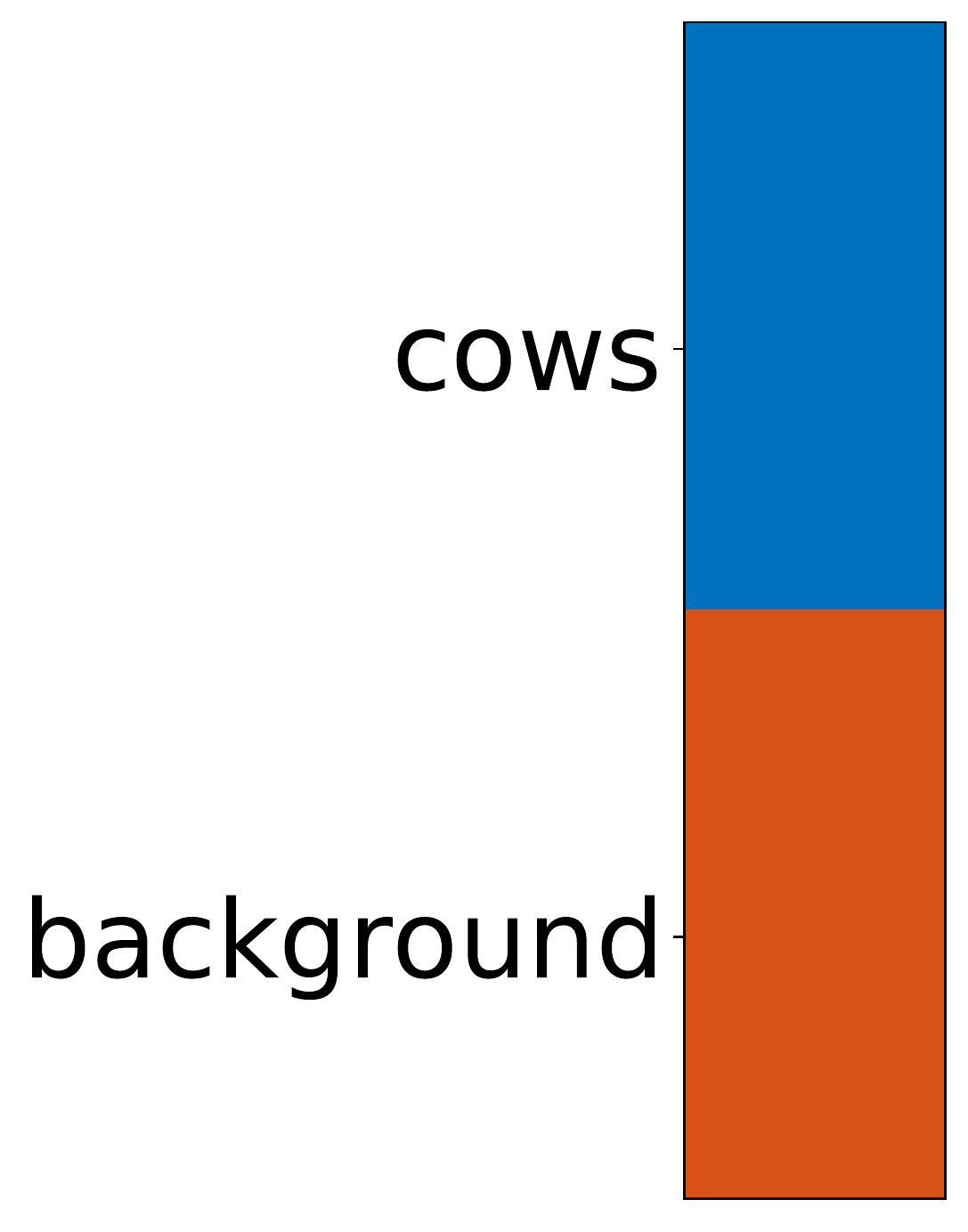}&%
\includegraphics[height=0.172\linewidth]{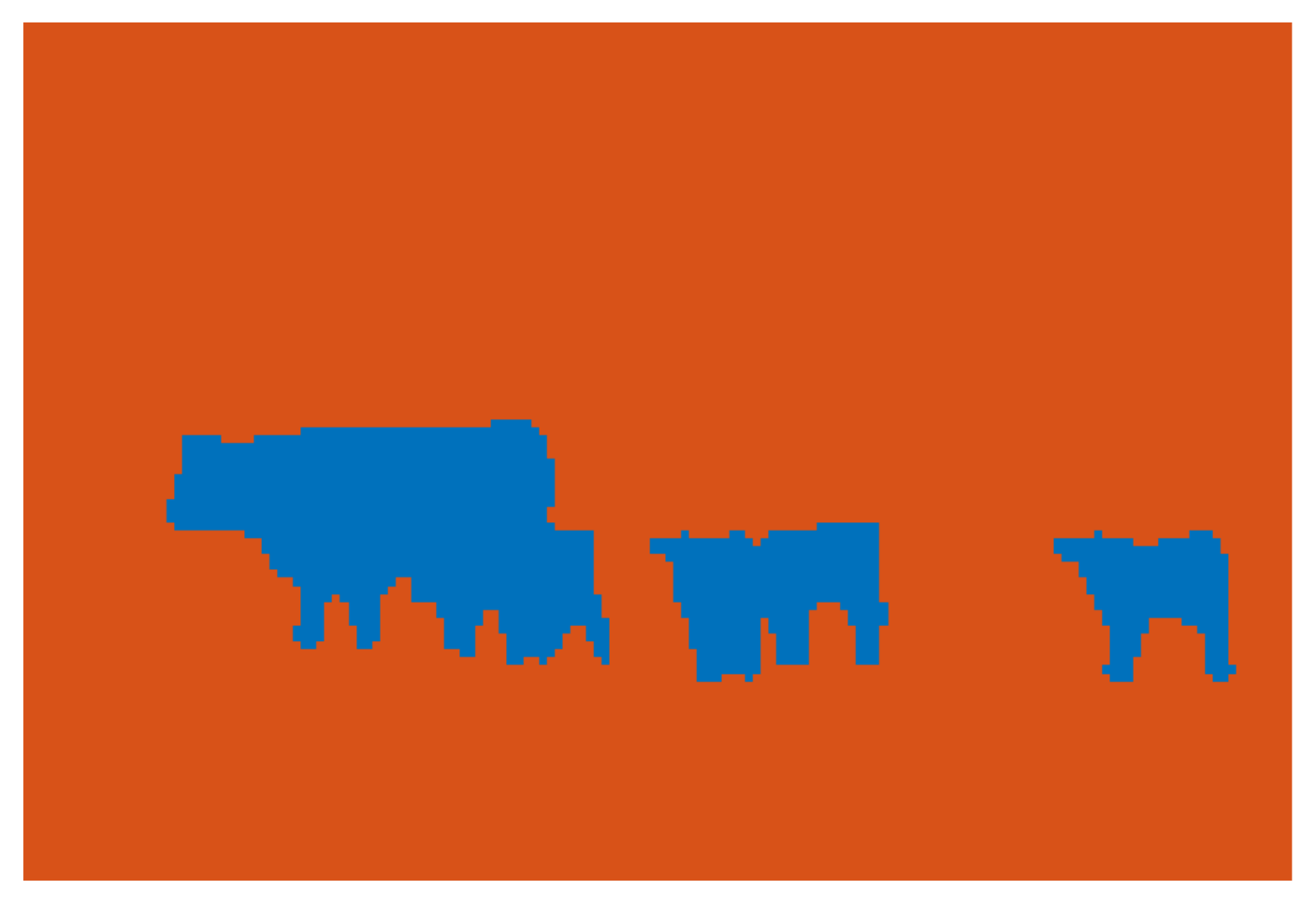}\\%
\end{tabular}
\endgroup
\captionof{figure}
{\footnotesize{\textbf{Examples of image segmentation with arbitrary text queries.} We propose a model, called \textbf{\ours}, that can organize pixels into meaningful regions indicated by texts. In contrast to segmentation models trained with close-vocabulary categories, \ours~can handle arbitrary text queries. For example, the model segments out a region for `couple' and two regions for `bride' and `groom'.
}}
\label{fig:open_vocabulary}

\section{Introduction}
\sloppy

Image segmentation is an important step to organize an image into a small number of regions in order to understand \textit{``what''} and \textit{``where''} are in an image. Each region represents a semantically meaningful entity, which can be a thing (\eg, a chair) or stuff (\eg, floor). 
Language is a natural interface to describe what is in an image. However, semantic segmentation algorithms often only learn with closed-set categories, and thus are unable to recognize concepts outside labeled datasets. Figure~\ref{fig:open_vocabulary} shows examples of image segmentation driven by language. The segmentation model takes text queries as inputs and produces segmented regions accordingly. In this work, we aim to learn open-vocabulary models which can segment an image and indicate regions with arbitrary text queries.

\begin{figure}[t]
    \centering
    \setlength\tabcolsep{1.5pt}
    \begin{tabular}{rrrr}%
    \includegraphics[height=0.18\linewidth]{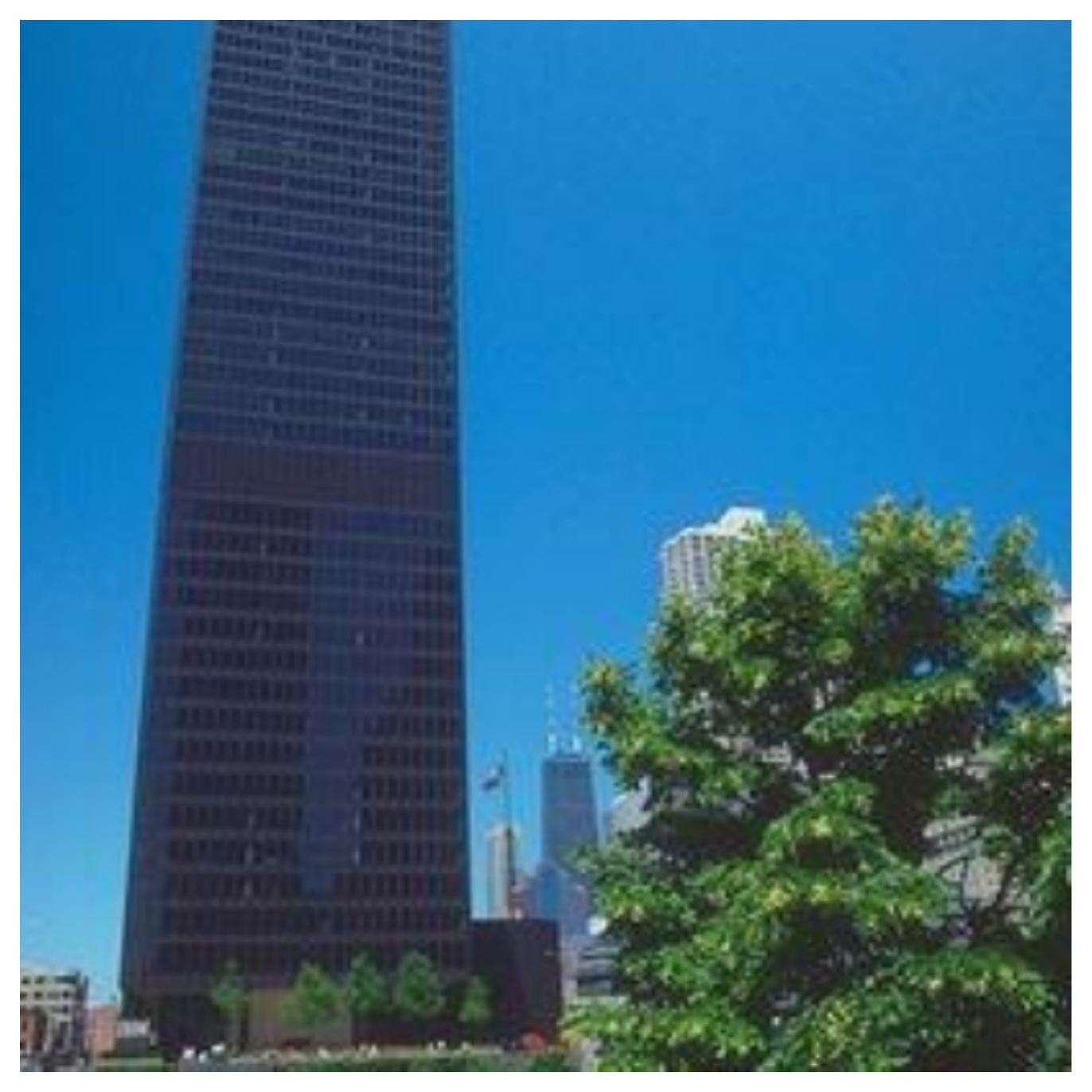}&%
    \includegraphics[height=0.18\linewidth]{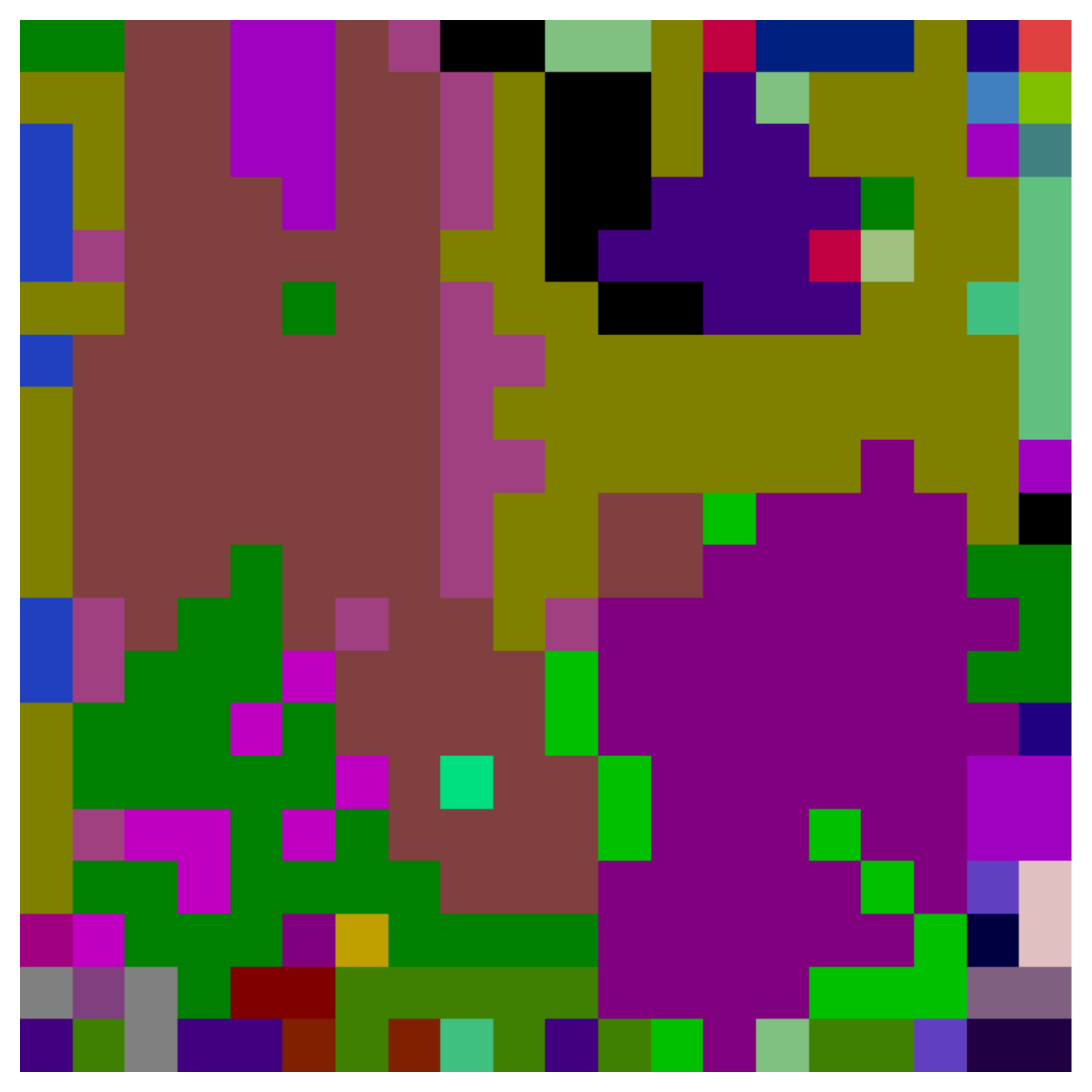}&%
    \includegraphics[height=0.18\linewidth]{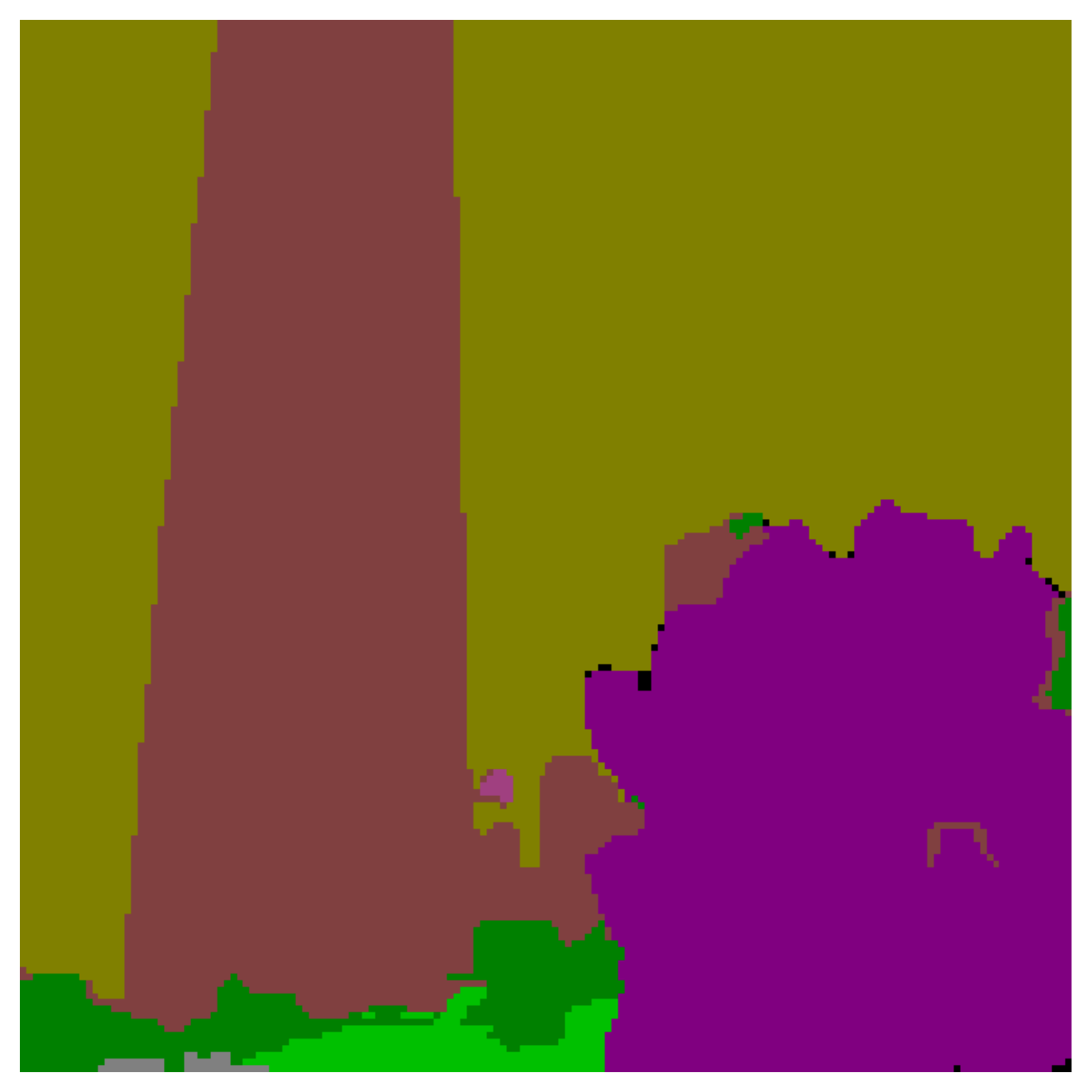}&%
    \includegraphics[height=0.18\linewidth]{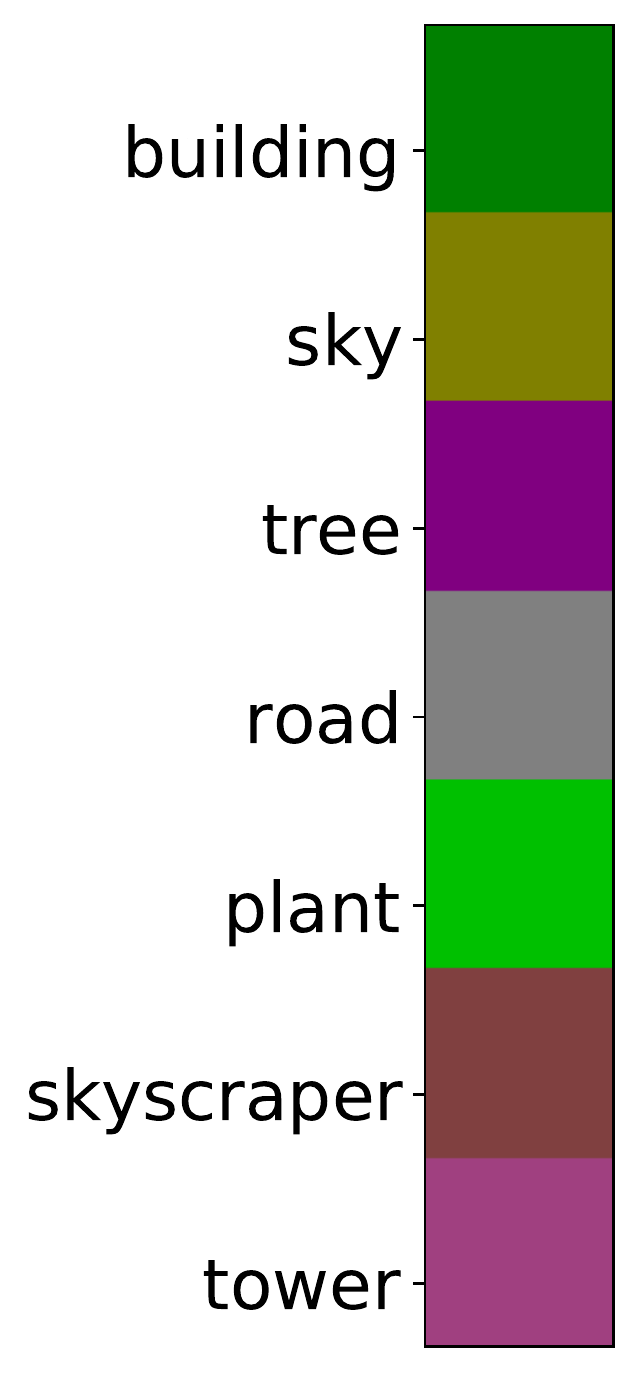}\\%
    \includegraphics[height=0.24\linewidth]{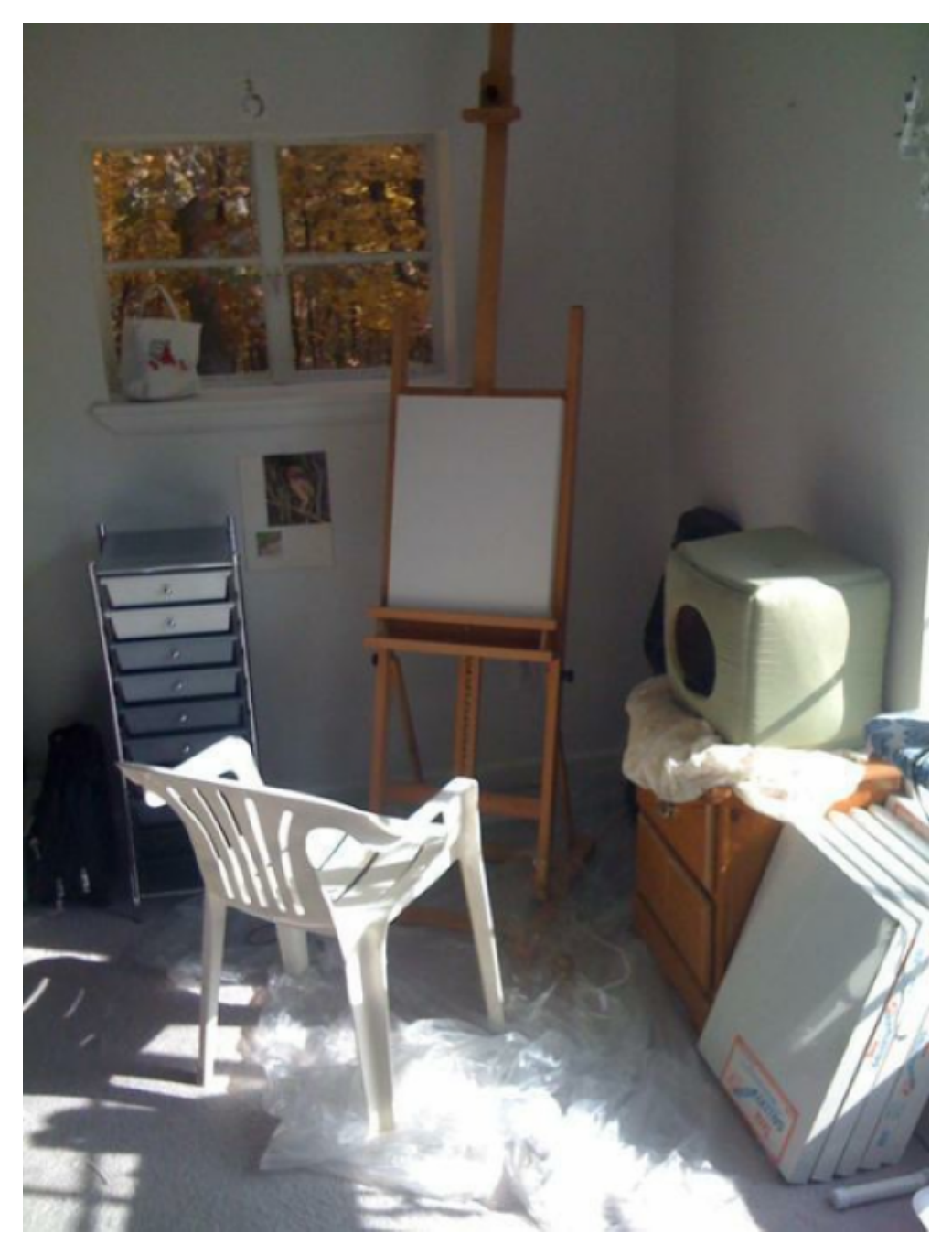}&%
    \includegraphics[height=0.24\linewidth]{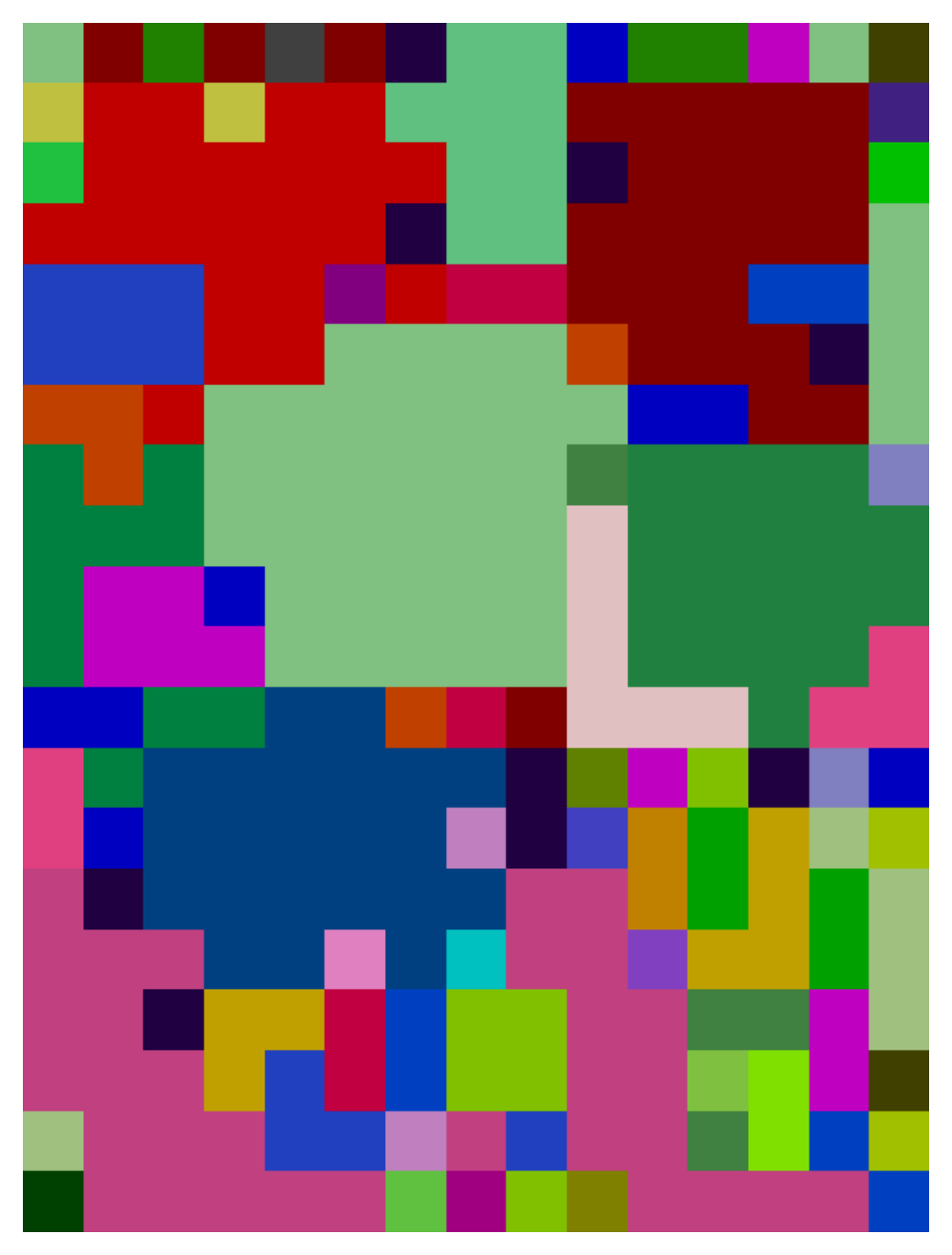}&%
    \includegraphics[height=0.24\linewidth]{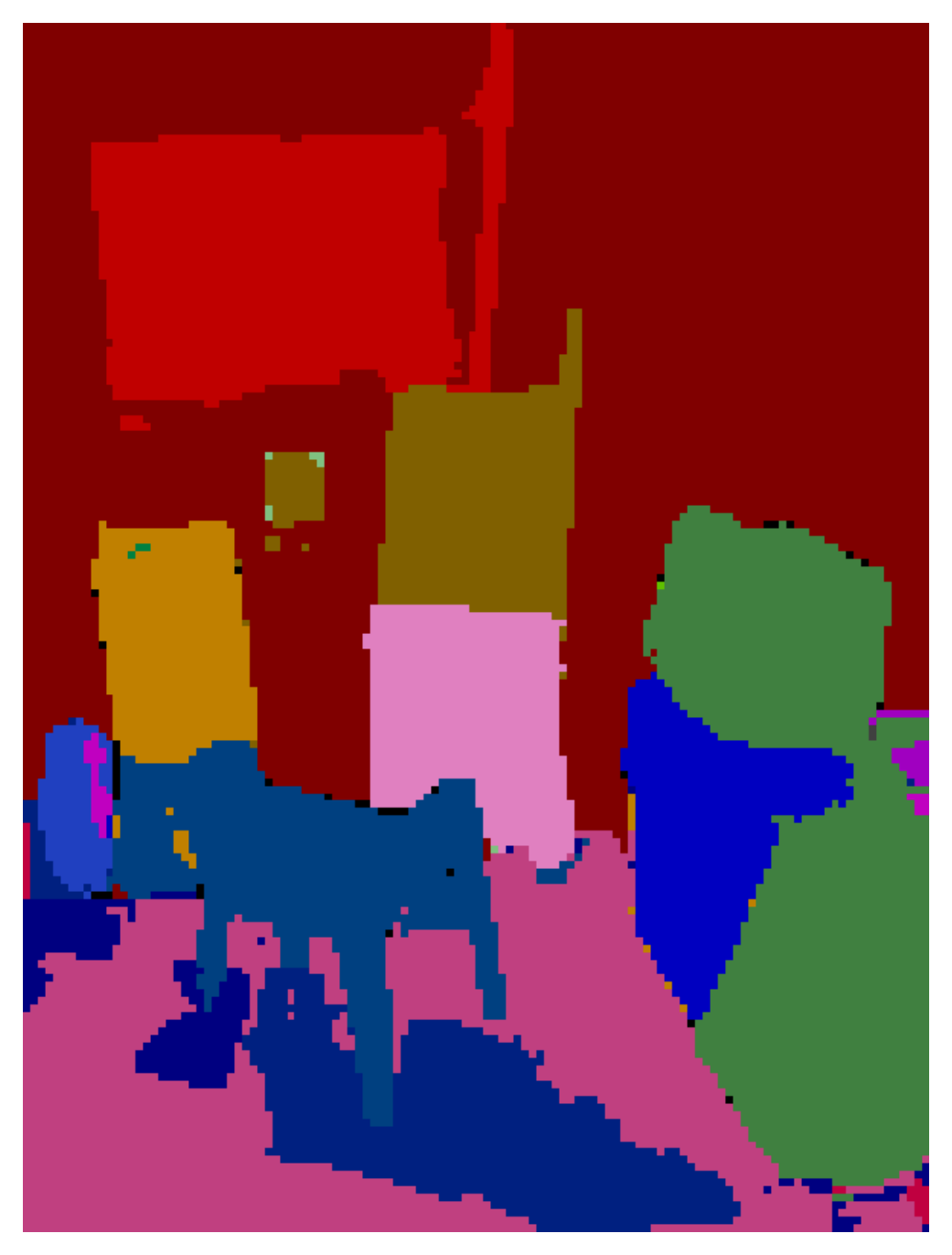}&%
    \includegraphics[height=0.24\linewidth]{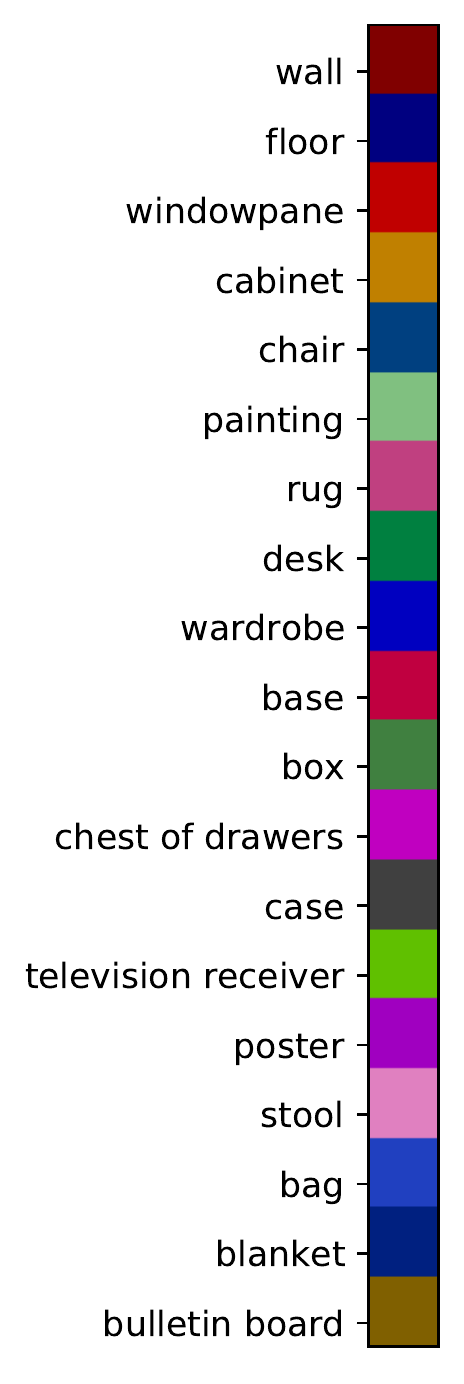}\\%
    \end{tabular}
\caption{\textbf{ALIGN (middle) can only roughly localize text queries onto the image.} In contrast, \ours~(right) can localize visual concepts with accurate segmentation. Moreover, ALIGN predicts more false positives not present in the image.}
\label{fig:teaser}
\end{figure}

Recently, CLIP~\cite{radford2021clip} and ALIGN~\cite{jia2021scaling} learn with billion-scale image-text training examples to understand \textit{``what''} are in an image with arbitrary text queries. These models demonstrate impressive results when directly evaluated on downstream image-text retrieval or classification tasks. However, localizing text queries to understand \textit{``where''} these visual concepts are in an image is still challenging. For example, Figure~\ref{fig:teaser} shows the segmentation predictions of a pre-trained ALIGN~\cite{jia2021scaling} model using class activation maps~\cite{zhou2015cnnlocalization}.

We argue that what is missing in these state-of-the-art open-vocabulary classification models are mid-level representations from visual groupings~\cite{wertheimer1923}, which organize an image into a small set of segmentation masks. Furthermore, visual-semantic alignments should perform after grouping to align texts to segmentation regions. However, these models represent an image with a single feature vector, inevitably losing much location information.

Recently, Li \etal~\cite{li2022language} introduce an open-vocabulary segmentation method using pre-trained CLIP~\cite{radford2021clip} text-encoders. 
It trains an image encoder to predict pixel embedding aligned with the text embedding of its pixel label.

However, the issue with this approach is in the scalability of training data. It is costly to annotate pixel-wise class labels, and thus requires generalization to unseen visual concepts from limited class labels.
We show that the visual-semantic alignments of image segmentation can be learned from scalable image caption labels.

In this work, we represent an image with a set of segmentation masks and their features.
We implement a class-agnostic segmentation module with region-to-image cross-attention~\cite{carion2020detr,huiyu2021max_deeplab,cheng2021maskformer} and train it with class-agnostic segmentation masks.
In contrast to the works using similar architectures~\cite{huiyu2021max_deeplab,cheng2021maskformer}, we do not predict the ``no object'' label $\varnothing$ to indicate if a predicted mask is a valid group of pixels. Considering the training data is only annotated with one possible organization of an image, we allow our model to predict other possible organizations beyond the annotations present in the training data.

Next, we learn visual-semantic alignments based on the predicted masks, which provide two major benefits in training. First, we perform mask-based feature pooling to aggregate pixels inside the predicted mask to generate location-aware region features. Second, the small number of predicted masks makes it easier to learn weakly-supervised alignments between regions to words in an image caption. The ability to learn from weak labels is important for scaling up training data and increasing vocabulary sizes. We call our method \textbf{\ours}, standing for open-vocabulary image segmentation.

To evaluate our method, we measure performances on holdout image segmentation datasets. We want to promote the framework where the model is trained with a large scale supervised/weakly-supervised data to learn \textit{generalist} models transferable to other datasets. Such a framework has been recently 
introduced
for image classification~\cite{radford2021clip,jia2021scaling} and object detection~\cite{zareian2021openvocabulary,gu2021vild}.  
To our knowledge, \ours~is the first work in image segmentation to demonstrate zero-shot transfer
results
across datasets using language. This is in stark contrast to the existing evaluation protocol which measures performances of \textit{specialist} models trained and tested using limited labeled data from the same dataset distribution.

In our experiments, we train the mask prediction model using class-agnostic mask annotations in the Panoptic COCO dataset~\cite{kirillov2019panoptic}. We show that the model can generalize well to other datasets, reaching superior performances compared with prior works on segmentation proposals~\cite{Arbelaez2014MCG,Maninis2018COB}. Then, we report mean intersection over union (mIoU) metrics for measuring both localization quality and accuracy of open-vocabulary semantic recognition. 
We compare  \ours~to the recent open-vocabulary method of LSeg. 
Thanks to the scalability of \ours, our best model significantly outperforms strongest LSeg model by 19.9 on PASCAL dataset.
We also compare \ours~to a version of LSeg implemented in our framework, trained on a larger semantic segmentation dataset of COCO (LSeg+). \ours~with ResNet-101 backbone outperforms LSeg+ models with similar backbone by 2.7 mIoU on PASCAL-Context (459 categories) and 1.9 mIoU on ADE-20k (847 categories). \ours~achieves this improvement mainly because of its ability to make use of image caption data which enables us to train it on a larger set of vocabularies and also a larger set of training examples.

\section{Related Work}
\paragraph{Grouping for visual recognition:}
Grouping has been a core research area in mid-level visual representations. 
The importance of grouping for human perception was pointed out almost a hundred years ago~\cite{wertheimer1923}. 
In machine perception, early works~\cite{comaniciu1997robust,shi2000normalized_cuts} group pixels based on local affinities.
Arbelaez~\etal~\cite{arbelaez2010contour} find contour detection and multiscale information helpful to generate segmentation and use it to predict object candidates~\cite{Arbelaez2014MCG}.
COB~\cite{Maninis2018COB} improves the efficiency and performance by leveraging deep nets.
These mid-level region representations are then used for semantic segmentation~\cite{Maninis2018COB} and object detection~\cite{uijlings2013selective_search}.
Recently, Qi~\etal~\cite{qi2021open} propose to segment all visual entities without considering semantic labels and show generalization to unseen domains.
In contrast to~\cite{qi2021open}, our work not only predicts segmentation, but also understands the semantics of segmented regions by open-vocabulary visual-semantic alignments.

\paragraph{Fully-supervised segmentation:}
To understand the semantics of pixels, several datasets have been developed with an increasing number of images and categories~\cite{everingham2010pascal,mottaghi2014role,zhou2018ade,caesar2018cocostuff}.
Models trained on these datasets can only learn to recognize the pre-defined classes,  which are at most in the order of hundreds for standard benchmarks.
Also, the classes across datasets are not transferable.
MSeg~\cite{lambert2020mseg} points out the ambiguity of class definitions, and manually resolves
it
to learn a transferable model across datasets. But the model still can not transfer to new visual concepts not present in the dataset.
\ours~overcomes these drawbacks.

\paragraph{Semantic segmentation with less supervisions:}
Weakly-supervised semantic segmentation trains with image-level labels~\cite{pinheiro2015image,wang2020self,jo2021puzzle,li2021pseudo,xu2021leveraging}, of which refining CAMs~\cite{zhou2015cnnlocalization} is a popular techniques.
Our model also adopts weak image-caption supervision, and it is different in that it has access to a set of class-agnostic segmentation annotations. Furthermore, it can transfer to arbitrary classes while these methods can not.
Zero-shot semantic segmentation methods~\cite{bucher2019zeroshotseg,xian2019semantic,hu2020uncertainty,li2020consistent,baek2021exploiting} aim to segment images with unseen visual concepts using language embeddings. These approaches learn with pixel-wise class labels which are expensive to scale up due to the long-tailed nature mentioned in the previous paragraph.
In contrast, we leverage cheap image caption data that covers a wide range of concepts, to achieve better and more practical performance on arbitrary categories.
In addition, we evaluate on datasets with much larger number of categories to verify the zero-shot transfer capability.

\paragraph{Open-vocabulary segmentation:}
Open-vocabulary segmentation aims to overcome the limitation of closed-set vocabulary in previous segmentation works.
Zhao~\etal~\cite{zhao2017openvocabulary_scene_parsing} is the pioneering work that learns a joint pixel and word concept embedding space; however, its vocabulary and knowledge is limited to WordNet and can not take arbitrary texts as input.

In a recent work, Li~\etal~\cite{li2022language} train an image encoder to encode pixel embeddings and use CLIP~\cite{radford2021clip} text embeddings as the per-pixel classifier.
Both Zhao~\etal~\cite{zhao2017openvocabulary_scene_parsing} and Li~\etal~\cite{li2022language} need per-pixel semantic supervision which is expensive to scale up. On the contrary, OpenSeg makes use of cheap image-level supervision such as captions, which allows scaling up training data.
There are multiple works concurrently developed with \ours:
GroupVit~\cite{xu2022groupvit} learn segmentation masks from text supervision.
Zabari and Hoshen~\cite{zabari2021semantic} use model interpretability to obtain pixel-level pseudo-labels from CLIP to supervise single-image segmentation methods; it's different from all other works as it does not need any training images, but the method is slow.
Zhou~\etal~\cite{zhou2021denseclip} adapt CLIP for segmentation, and use pseudo per-pixel labels and self-training to boost the performance; similar to Li~\etal~\cite{li2022language}, it utilizes per-pixel semantic supervision.
Xu~\etal~\cite{xu2021simple} first generate mask proposals, and then leverage CLIP for classification of the proposals.  In contrast, we learn visual-semantic alignment from image captions, which is no longer limited by image classification models (\eg, CLIP).

\paragraph{Visual grounding:}
Image captioning and image text datasets~\cite{plummer2015flickr30k,coco_captions,krishna2016visualgenome} enable research on the interplay of captions and grounded visual concepts~\cite{feng2015caption_and_back,rohrbach2016grounding,fukui2016multimodal,gupta2020contrastive,kamath2021mdetr}. However, these methods often rely on an object detector to predict object bounding boxes for grounding. Therefore, they are not able to handle stuff and can not generate a single segmentation map for everything.
Our method also uses captions as semantically-rich supervision. We draw inspiration from these works and expand the model's ability to ground visual concepts of both things and stuff to pixels with our mask representations.
\paragraph{Referring image segmentation:}
The goal of this task is to compute a binary mask localizing a referring expression. 
Since there are multiple supervised datasets (\eg, RefCOCO~\cite{yu2016modeling}) for this task, previously developed methods are usually fully supervised~\cite{hu2016segmentation,yu2018mattnet,ding2021vision,hu2020bi,ye2019cross}. As a result, the training data for these methods are not scalable.
\begin{figure}[t]
\includegraphics[width=1.0\linewidth]{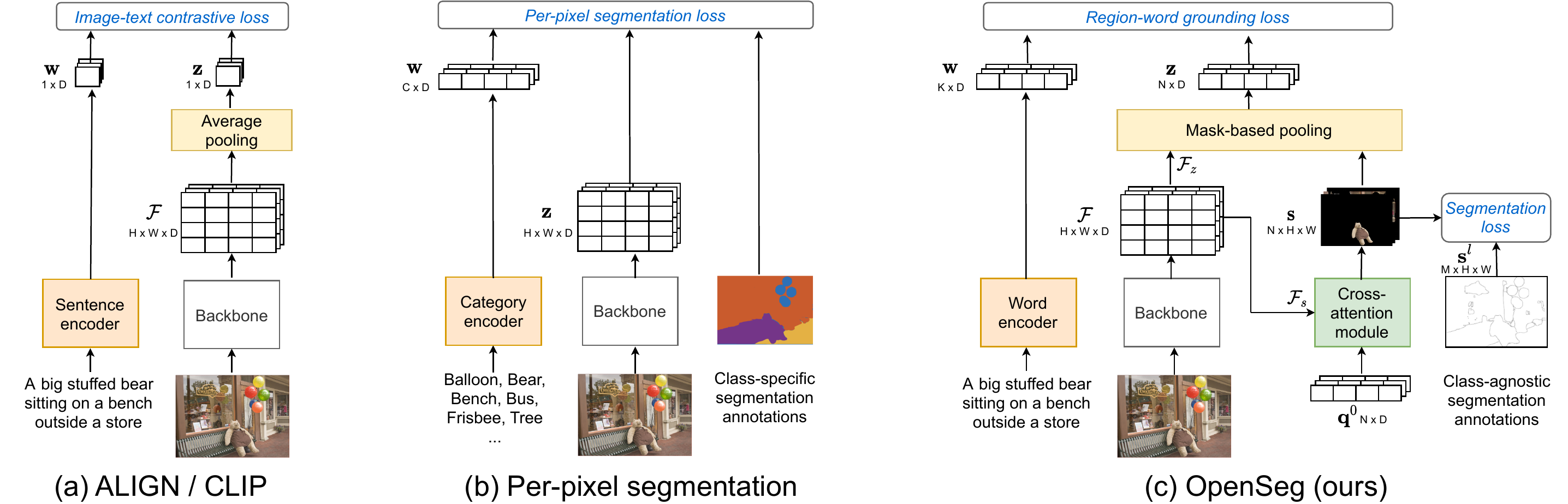}
\caption{\textbf{An overview of our approach}. We compare \ours~with ALIGN / CLIP~\cite{jia2021scaling,radford2021clip} and per-pixel segmentation models such as LSeg~\cite{li2022language}. The major differences are in the image and text representations $\textbf{z}$ and $\textbf{w}$. ALIGN / CLIP has $\textbf{z} \in \mathbb{R}^{1 \times D}$, losing location information. Per-pixel segmentation represents an image with $\textbf{z} \in \mathbb{R}^{H \times W \times D}$, requiring class-specific mask annotations for training. \ours~represents an image with a set of $N$ segmentation regions $\textbf{z} \in \mathbb{R}^{N \times D}$, facilitating weakly-supervised learning using captions.}
\label{fig:overview}
\end{figure}

\section{Method}
Figure~\ref{fig:overview} shows an overview of our approach. In contrast to approaches that represent an image with a vector $\mathcal{Z} \in \mathbb{R}^{1 \times D}$ or a feature map $
\mathcal{Z} \in \mathbb{R}^{H \times W \times D}$, \ours~represents an image with $N$ proposal masks with their features $\mathcal{Z} \in \mathbb{R}^{N \times D}$.
Our mask representations support learning precise image segmentation with image captions by weakly-supervised learning.
In Section~\ref{sec:grouping}, we describe the learning of predicting mask proposals from an image.  In Section~\ref{sec:alignment}, we describe the feature representations of proposal and the learning of region-word alignments. In the following sections, We use a bold symbol to indicate an array of elements $\textbf{x} = \{x_1, x_2, ..., x_n\}$, where the first dimension indicates the number of elements.

\subsection{Learning Segmentation Masks}
\label{sec:grouping}
We design a model architecture which consists of a feature pyramid network (FPN)~\cite{fpn} for multi-scale feature extraction and a cross-attention module for segmentation region proposal. We fuse FPN features into $P_2$ resolution as described in~\cite{must} to generate image features $\mathcal{F}$.
From $\mathcal{F}$, we obtain $\mathcal{F}_s \in \mathbb{R}^{H \times W \times D}$ by convolution and fc layers. Then we augment image features by adding learnable position embeddings $PE$: $\mathcal{F}_s^{PE} = \mathcal{F}_s + PE$. We use a cross-attention module taking inputs as $\mathcal{F}_s^{PE}$ and a randomly initialized queries $\textbf{q}^0 \in \mathbb{R}^{N \times D}$ to generate mask queries $\textbf{q}  \in \mathbb{R}^{N \times D}$. Then, we compute the dot product of mask queries and position-augmented image features to predict masks $\textbf{s} = Sigmoid(dot(\textbf{q}, \mathcal{F}_s^{PE})) \in\mathbb{R}^{N\times H\times W}$. This architecture is conceptually similar to  Max-deeplab~\cite{huiyu2021max_deeplab} and MaskFormer~\cite{cheng2021maskformer}. The details of the architecture are in Appendix~\ref{sec:cross_attention_arch}.

We compute Dice coefficient~\cite{milletari2016vnet_dice} between predicted masks $\mathbf{s}$ and class-agnostic labeled masks $\textbf{s}^{l} \in \mathbb{R}^{M \times H \times W}$ and maximize the Dice coefficient of the best matched mask for each labeled mask.
\begin{equation}
\mathcal{L_S} = 
\frac{1}{M}
\sum_{j=1}^{M} 
(1 - \max_i Dice(s_i, s_j^{l}))
\end{equation}
Typically, $N > M$ for each training image. Therefore, a subset of proposal masks are optimized to best match labeled masks.
The rest of proposals can still segment out unlabeled regions without being penalized.
One predicted mask may match to multiple labeled masks in the early training stage when their overlaps are low. But this does not prevent learning masks that highly overlap with labeled masks in the latter training stage.

\subsection{Visual-Semantic Alignment with Masks}
\label{sec:alignment}
We use a pair of image $I_b$ and caption $C_b$ to learn visual-semantic alignments. We break $I_b$ into regions (Section~\ref{sec:grouping}) and $C_b$ into words by extracting list of nouns and adjectives from the caption. We randomly drop each word with the probability of $1-kp$, where $kp$ is the keep probability of words extracted from captions.
We generate image features $\mathcal{F}_z$ using the same architecture as $\mathcal{F}_s$.
For each region, we compute its feature by pooling image features with the mask $\textbf{z}[n, d] = \sum_{ij} \textbf{s}[n, i, j] \cdot \mathcal{F}_z[i, j, d]$.
We feed each word to a pre-trained text encoder to compute the word feature $w$. 

We follow the grounding loss in prior works~\cite{gupta2020contrastive,zareian2021openvocabulary} to learn region-word alignments. We first define the notation for Softmax on an array $\textbf{x}$ to get the normalized score at the $i$-th element:

\begin{equation}
\sigma(\textbf{x})_i =
\frac{e^{x_i / \tau}}
{\sum_{j}e^{x_j / \tau}}
\end{equation}
where $\tau$ is a learnable scalar for the temperature.
The similarity score of a region $i$ and a word $j$ is defined by its cosine similarity $\langle z_i, w_j \rangle = \frac{z_i \cdot w_j}{\|z_i\| \|w_j\|}$.  Then we define the similarity of all regions $\textbf{z}$ to a word $w_j$ as: $g(\textbf{z}, w_j) = [
\langle z_1, w_j \rangle,
...,
\langle z_N, w_j \rangle] \in \mathbb{R}^{N \times 1}$. We compute the similarity of an image $I_b$ and its caption $C_b$ by:
\begin{equation}
    G(I_b, C_b) = \frac{1}{K}
    \sum_{j=1}^{K}
    \sum_{i=1}^{N}
    \sigma(g(\mathbf{z}, w_j))_i
    \cdot
    \langle z_i, w_j \rangle
\end{equation}
The above similarity function encourages each word to be grounded to one or a few regions. Also, it avoids penalizing regions that can not find any similar word. Next, a grounding loss is defined for a given mini-batch $B$, where each example contains an image-caption pair. We define the similarity scores of all images in a batch $\textbf{I}$ to a caption $C_b$ by $G(\mathbf{I}, C_b) = [G(I_1, C_b), ..., G(I_{|B|}, C_b)] \in \mathbb{R}^{|B| \times 1}$ and similarly $G(I_b, \mathbf{C}) = [G(I_b, C_1), ..., G(I_b, C_{|B|})] \in \mathbb{R}^{|B| \times 1}$. The grounding loss aims at maximizing the normalized score of a labeled image-caption pair $\langle I_b, C_b \rangle$ over all images and all captions in a mini-batch.
\begin{equation}
    \mathcal{L_G} = -\frac{1}{|B|}
    \sum_{b=1}^{|B|}\Big(
    \log \sigma\big(G( \mathbf{I}, C_b)\big)_b + 
    \log \sigma\big(G( I_b, \mathbf{C})\big)_b\Big)
\end{equation}
 To train \ours, we simply sum the two losses with a weight $\alpha$:
\begin{equation}
    \mathcal{L} = \mathcal{L_G} + \alpha \mathcal{L_S}
\end{equation}
When setting $\alpha=0$, the model learns without labeled class-agnostic segmentation, and thus needs to induce mask predictions with the visual-semantic grounding loss. We find this setting leads to a poor performance, suggesting class-agnostic mask annotations are critical for learning mask predictions.

\subsection{Learning from Caption Only Data}
Since annotating images with segmentation is expensive, to scale up the training data we need to learn from images with only caption annotations. We follow MuST~\cite{must} and first train a teacher model on a segmentation dataset with only the segmentation loss $\mathcal{L_S}$. Then we annotate a large image-text dataset with pseudo segmentation labels using the teacher model. Lastly, the \ours~model is trained with a mix of human and pseudo labels.

\subsection{Inference}
\label{sec:inference}
Up to this point, we learn a vision model that predicts segmentation masks $\textbf{s} \in \mathbb{R}^{N \times H \times W}$ and corresponding features $\textbf{z} \in \mathbb{R}^{N \times D}$.
Given an evaluation segmentation dataset, we encode its categories using the text encoder. If a category is defined by more than one word, we simply include all word embeddings for that category. We obtain $K$ word embeddings $\textbf{w} \in \mathbb{R}^{K \times D}$ representing all categories.
The region logits are obtained by taking the cosine similarity between words and regions $\langle \textbf{w}, \textbf{z} \rangle \in \mathbb{R}^{K \times N}$.
We multiply the region logits and segmentation masks to obtain segmentation logits at each pixel $\textbf{y} = \langle \textbf{w}, \textbf{z} \rangle \cdot \textbf{s} \in \mathbb{R}^{K \times H \times W}$.
Then the category prediction at each pixel is an argmax of segmentation logits along the word dimension:

\begin{equation}
    pred[i, j] = \argmax_k \textbf{y}[k, i, j]
\end{equation}

\section{Experiments}
\subsection{Experimental Settings}
\label{sec:experiment_setting}

\paragraph{Architecture.}
We use EfficientNet-B7~\cite{tan2019efficientnet} (and ResNet101 in Table~\ref{tab:sota}) as the backbone architecture and employ FPN~\cite{fpn} for multi-scale feature fusion. We use pyramid levels from $P_2$ to $P_5$ with feature dimension 640, upsample all feature levels to $P_2$, and then merge them by a sum operation to obtain $\mathcal{F}$.
To compute $\mathcal{F}_z$ and $\mathcal{F}_s$, we apply a fc layer followed by 3 layers of $3 \times 3$ convolutions with 640 channels after $\mathcal{F}$.
For text encoder we use the frozen pre-trained BERT-Large model in ALIGN~\cite{jia2021scaling}.

\paragraph{Training Parameters.}
All models are trained with an image size of $640\times640$. We apply multi-scale jittering with a random scale between $[0.8, 1.2]$ (\ie, small scale jittering in~\cite{ghiasi2021simple}). The weight decay is set to 1e-05 and we use a learning rate 0.005 with the cosine learning rate schedule. Unless otherwise mentioned, we initialize the backbone of the model from the ALIGN checkpoint~\cite{jia2021scaling}. We train \ours~on COCO dataset for 30k steps. For training on COCO and Localized Narrative datasets, we sample examples from the datasets with equal probability and we train the model for 60k steps. We set
$kp$ (keep probability of words extracted from captions) to 0.5.
We train models with global batch size of 1024 and local batch size of 16 (we have 64 Cloud TPU v3 cores). Unless otherwise stated, for each core we compute the loss over the local batch of examples (See Appendix~\ref{sec:batch_size_ablation} for the comparison between sync and unsync contrastive loss over the cores and also comparison of training with smaller batch sizes).
\subsubsection{Training Datasets}
\paragraph{COCO:} We use the panoptic segmentation~\cite{kirillov2019panoptic} and caption~\cite{coco_captions} annotations in the 2017 splits which include 118k/5k train/val images. We utilize the panoptic segmentation annotations in a class-agnostic manner. When evaluating on COCO Panoptic, we treat it as a semantic segmentation dataset and our model only predicts the semantic class for each pixel.

\paragraph{Localized Narrative (Loc.\ Narr.):} Localized Narrative~\cite{locnar} contains detailed natural language descriptions along with mouse traces for multiple datasets (COCO, Flickr, Open Images, ADE20k). We don’t train on the ADE20k portion to keep its image distribution unseen. The remaining 652k images are used for training.

\subsubsection{Evaluation Datasets}

\paragraph{PASCAL Context:}
PASCAL Context~\cite{mottaghi2014role} includes per-pixel segmentation annotations of object and stuff on 5k/5k train/val images from various indoor and outdoor senses.
The full version (\textbf{\pascfull}) includes 459 classes.
The version with the most frequent 59 classes (\textbf{\pasc}) is widely used in the existing literature.

\paragraph{PASCAL VOC:} PASCAL VOC 2012~\cite{everingham2010pascal} includes 20 object classes and a background class with 1.5k/1.5k train/val images.
Since the text ``background'' is ambiguous, we assign the background class to the pixels predicted as \textbf{\pasc} categories that are not in PASCAL VOC.

\paragraph{ADE20k:}
ADE20k~\cite{zhou2019semantic} includes 20k/2k train/val images with 
segmentation annotations and covers a wide variety of indoor and outdoor scenes.
The full version has annotations in an open-vocabulary setting and includes 2693 object and stuff classes.
We follow~\cite{cheng2021maskformer} and evaluate on the version with 847 classes (\textbf{\adefull}).
We also test on the widely-used version with 150 frequent categories (\textbf{\ade}).

\subsection{Predicting Masks Across Datasets}
We train the segmentation proposal model on COCO and evaluate on COCO and \pasc~with recalls at IoU 50\%, 70\%, and 90\% as metrics. Table~\ref{tab:proposals} shows performance comparisons with MCG~\cite{Arbelaez2014MCG} and COB~\cite{Maninis2018COB} using their pre-computed proposals.
\ours~shows significantly superior performances. We perform additional cross-dataset evaluation using datasets in MSeg~\cite{lambert2020mseg} in Appendix~\ref{sec:mseg_experiments}.
Figure~\ref{fig:proposals} shows 6 manually selected proposals to demonstrate our model can organize images into semantically meaningful regions. Particularly, the underwater scene is not present in our training dataset COCO, but the model can still organize pixels into regions for ocean, coral, diver, goggles, \etc. The full 128 proposals are included in Appendix~\ref{sec:visualization_of_full_proposals}.

\begin{table}[t]
\scriptsize
\centering
\caption{\textbf{Recall of segmentation mask proposals} on COCO and PASCAL-Context datasets. All methods use 128 proposals.}
\begin{tabular}{l|rrr|rrr}
& \multicolumn{3}{c|}{COCO} & \multicolumn{3}{c}{PASCAL Context-59} \\
& R50 & R70 & R90 & R50 & ~~~~~R70 & R90 \\
\hline
MCG~\cite{Arbelaez2014MCG} &  41.1 & 21.4 & 4.6 & 57.8 & 31.7 & 8.7 \\
COB~\cite{Maninis2018COB} &  46.0 & 24.8 & 4.9 & 62.9 & 37.6 & 12.1 \\
\hline
\ours & \textbf{68.9} & \textbf{48.1} & \textbf{16.9} & \textbf{84.5} & \textbf{65.1} & \textbf{29.1}\\
\end{tabular}
\label{tab:proposals}
\end{table}
%

\begin{figure}[t]%
\centering%
\begin{tabular}{c|cccccc}%
\includegraphics[width=0.135\linewidth]{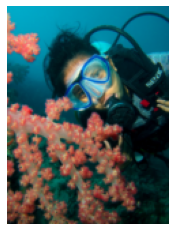} &%
\includegraphics[width=0.135\linewidth]{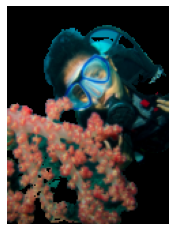}%
\includegraphics[width=0.135\linewidth]{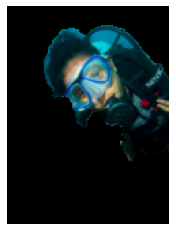}%
\includegraphics[width=0.135\linewidth]{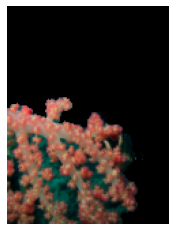}%
\includegraphics[width=0.135\linewidth]{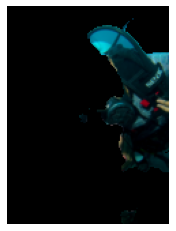}%
\includegraphics[width=0.135\linewidth]{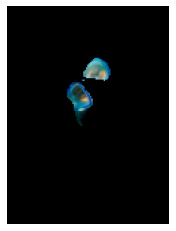}%
\includegraphics[width=0.135\linewidth]{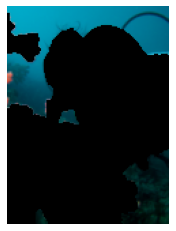}\\%
\end{tabular}
\caption{\textbf{Examples of predicted segmentation masks in an unseen scene.} \ours~is able to segment an image into meaningful regions. These regions may be overlapping and indicate concepts of foreground (diver and coral) \vs background (ocean), and whole (diver) \vs parts (scuba and goggles). Notably, \ours~is trained on COCO which does not include underwater scenes.}
\label{fig:proposals}
\end{figure}

\subsection{Open-vocabulary Image Segmentation}
\label{sec:open_vocab_segm}
In this section, we first describe open-vocabulary baselines and our evaluation metrics. Then we discuss the experimental results with our open-vocabulary baselines and state-of-the-art open-vocabulary and zero-shot methods.

\paragraph{ALIGN baseline:} Although ALIGN~\cite{jia2021scaling} is trained for open-vocabulary classification, it can still roughly localize objects and stuff with arbitrary text queries (see Figure~\ref{fig:teaser}). Since we initialize the backbone of \ours~from ALIGN's pre-trained checkpoint, we use ALIGN as a baseline. 
We follow the CAM~\cite{zhou2015cnnlocalization} method for segmentation prediction. We compute the activation map before the average pooling layer of the image encoder. Then for each spatial location we compute its cosine similarity with the text embeddings of all input categories. We assign the class with the highest similarity to each location.

\paragraph{LSeg baseline:}
Recently,~\cite{li2022language} introduce an open-vocabulary segmentation method which trains an image encoder to encode pixel embeddings and use CLIP~\cite{radford2021clip} text embeddings as the per-pixel classifier.
Figure~\ref{fig:overview}(b) illustrates the model of this approach.
For a fair comparison, we also construct LSeg in our codebase as follows.
We add FPN and introduce a high resolution map in the same approach in Section~\ref{sec:experiment_setting}. We embed class names into text embeddings using ALIGN~\cite{jia2021scaling} text-encoder and use them as per-pixel classifiers. We fine-tune the pre-trained image encoder and FPN layers on COCO dataset using a per-pixel cross-entropy loss to align pixel embeddings with text embeddings. We call this model LSeg+.

\paragraph{ALIGN w/proposal baseline:}
The ALIGN, LSeg and LSeg+ baselines are methods that perform visual-semantic alignments without explicit visual grouping. Since our method uses visual grouping, we also compare our method to ALIGN w/proposal baseline which leverage proposals generated by OpenSeg at inference. We use the ALIGN model to classify each proposal and then similarly to \ours~we aggregate all proposals to compute the final segmentation map.

\paragraph{Evaluation metrics:}
We use two metrics, \textit{mIoU} and \textit{\imiou}, for evaluation. Both metrics are calculated using the standard mIoU formula~\cite{everingham2010pascal} and only differ in the text queries for each image.
The mIoU is commonly used in literature.
It measures the performance of image segmentation with fixed text queries, \eg, 847 classes when evaluated for all images in \adefull.
The \imiou~evaluates concept grounding. An example scenario is interactive segmentation where users can specify a set of concepts in an image for the model to segment.
It only uses the ground-truth classes in an image, \eg, 7 classes are used as text queries for the example in the second row of Figure~\ref{fig:main_result}.
We find that predictions in the mIoU and \imiou~settings can look quite differently and sometimes mIoU does not correctly reflect the prediction quality due to class ambiguity.
For example, building, brick, house are all correct visual concepts to describe the object in Figure~\ref{fig:main_result} but the ground-truth label is building.

\newcommand{\moretiny}{\fontsize{5.5}{2}\selectfont}

\begin{figure}[tb]%

\begin{tabular}{@{}c@{}}
\begingroup
\centering%
\setlength\tabcolsep{.5pt}
\renewcommand{\arraystretch}{0.1}
\begin{tabu}{>{\supertiny}c>{\supertiny}c>{\supertiny}c>{\supertiny}c>{\supertiny}c>{\supertiny}c>{\supertiny}c}%
&\bf{Image/Ground-truth}&\bf{ALIGN}&\bf{LSeg+}&\bf{\ours}&\bf{\ours~w/Loc.Narr.}&\bf{\ }\\
\rotatebox{90}{\bf{\makecell{Dataset Classes \\(mIoU)}}}&%
\includegraphics[width=0.176\linewidth]{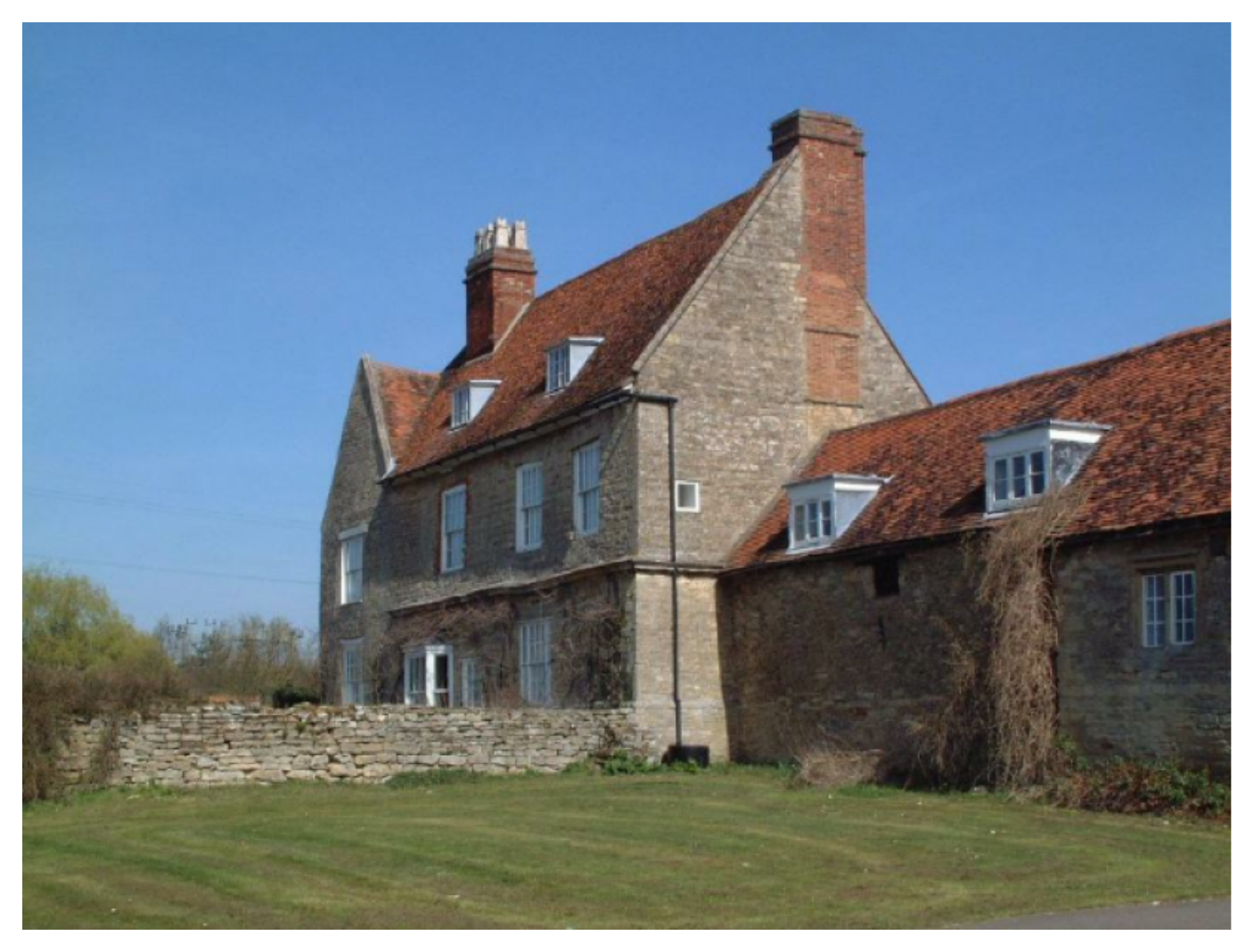}&%
\includegraphics[width=0.176\linewidth]{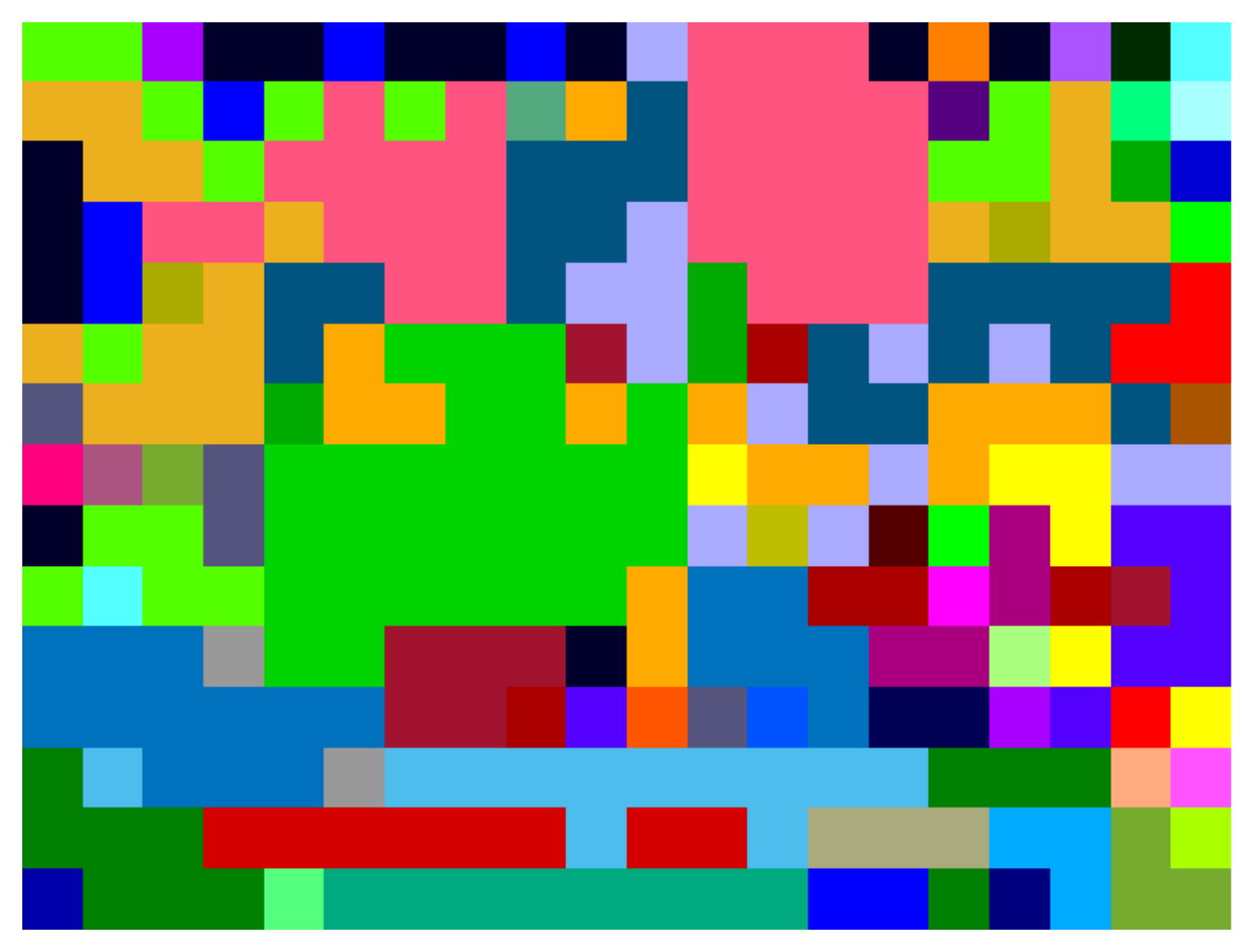}&%
\includegraphics[width=0.176\linewidth]{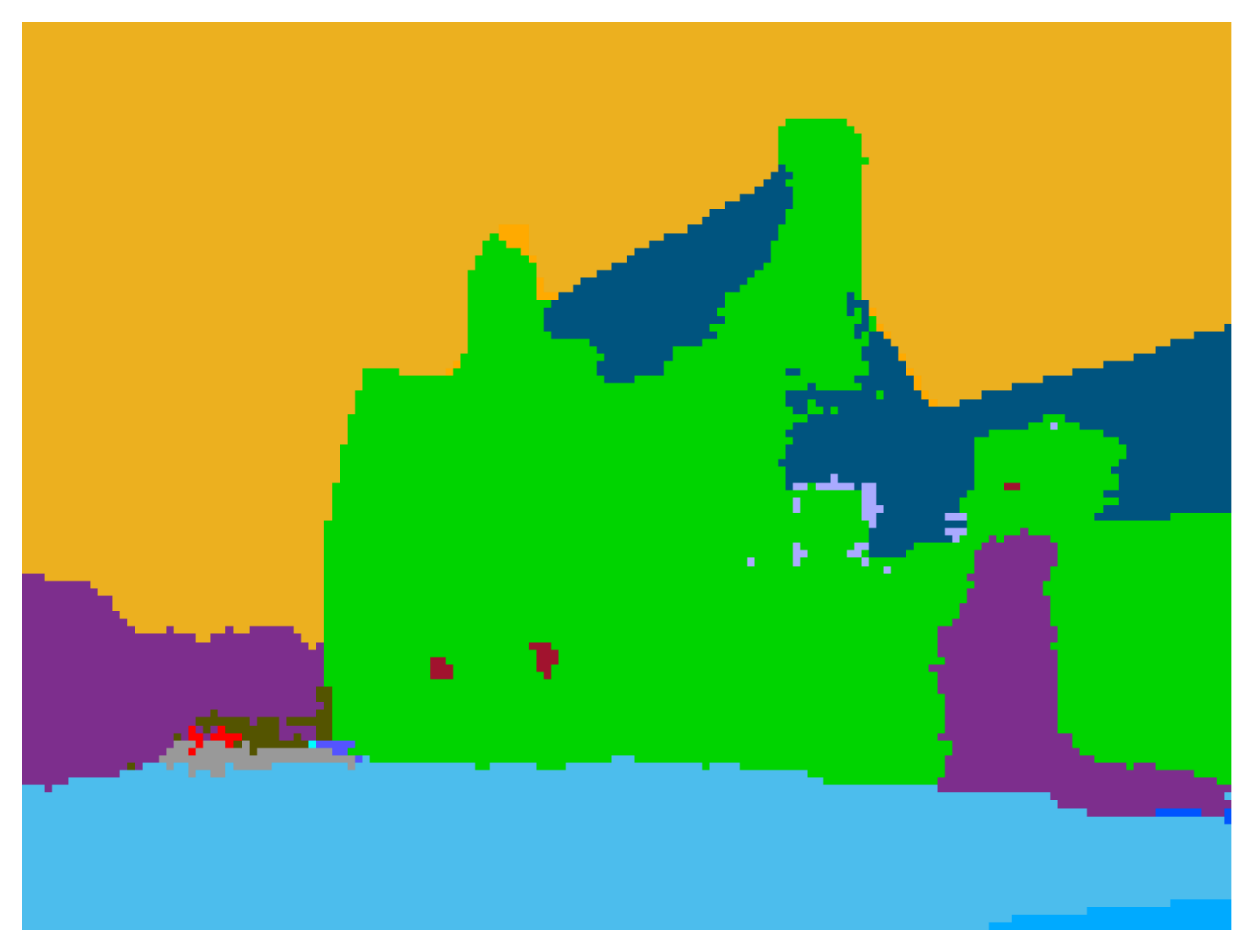}&%
\includegraphics[width=0.176\linewidth]{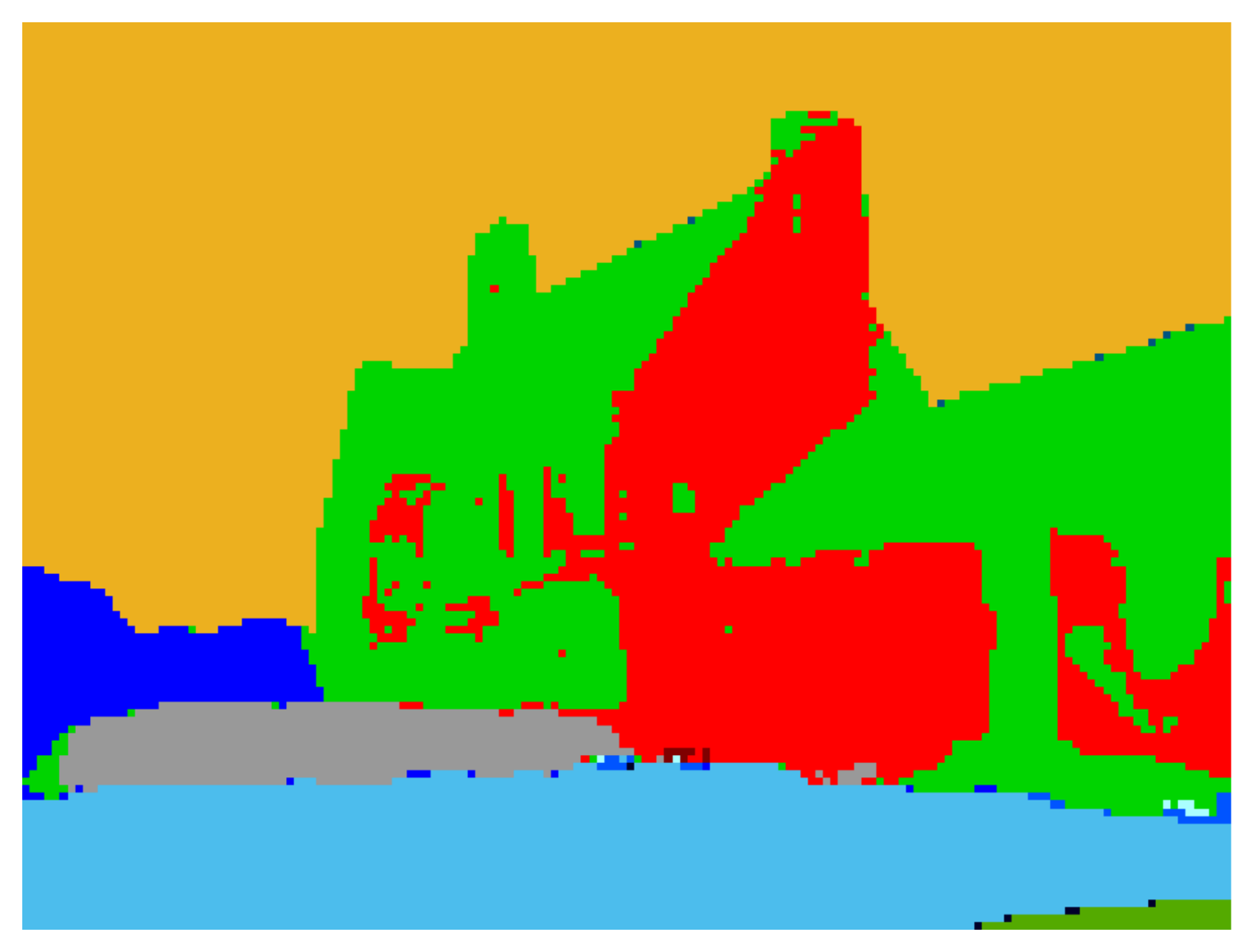}&%
\includegraphics[width=0.176\linewidth]{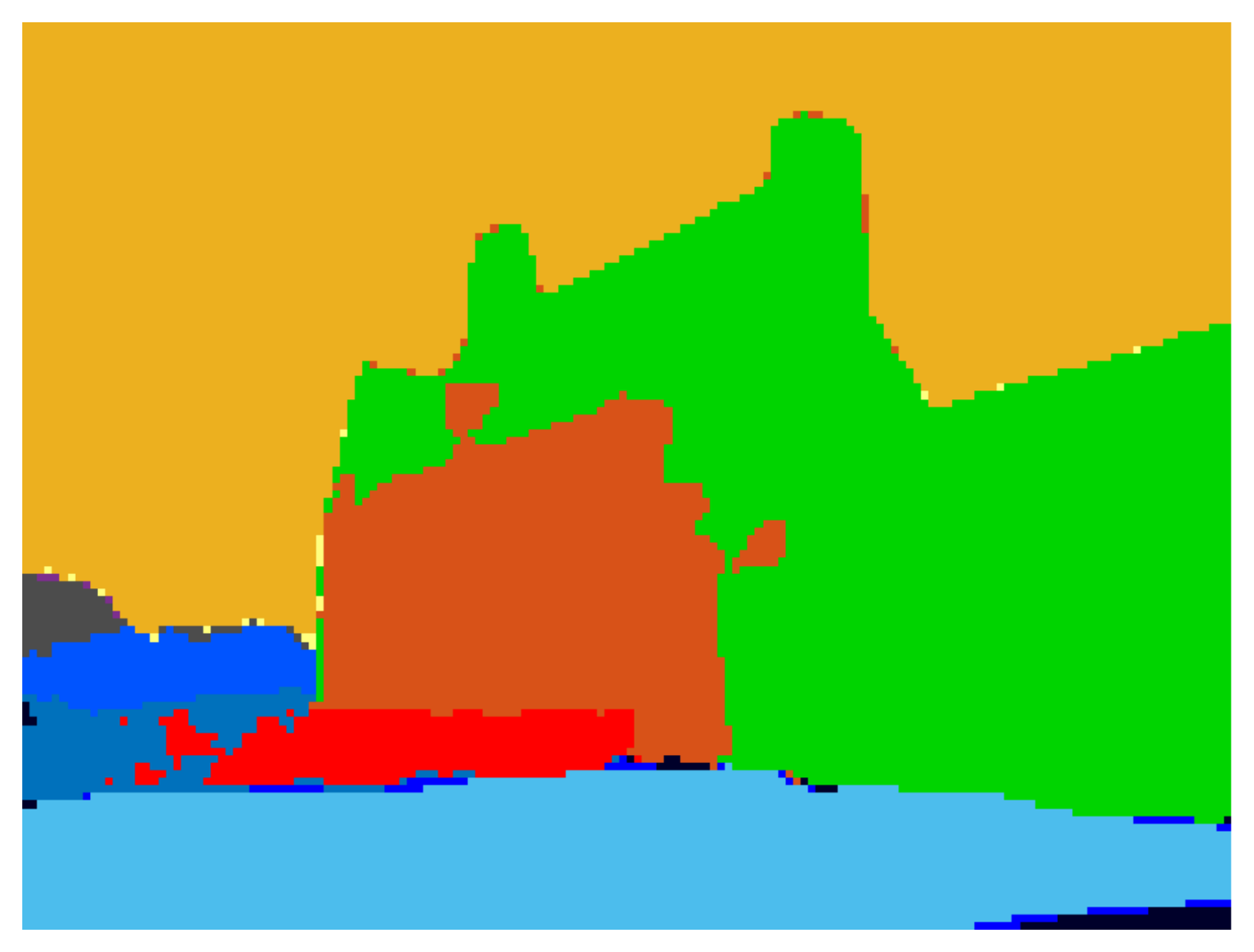}&%
\multirow{2}[2]{*}[1mm]{\includegraphics[width=0.051\linewidth]{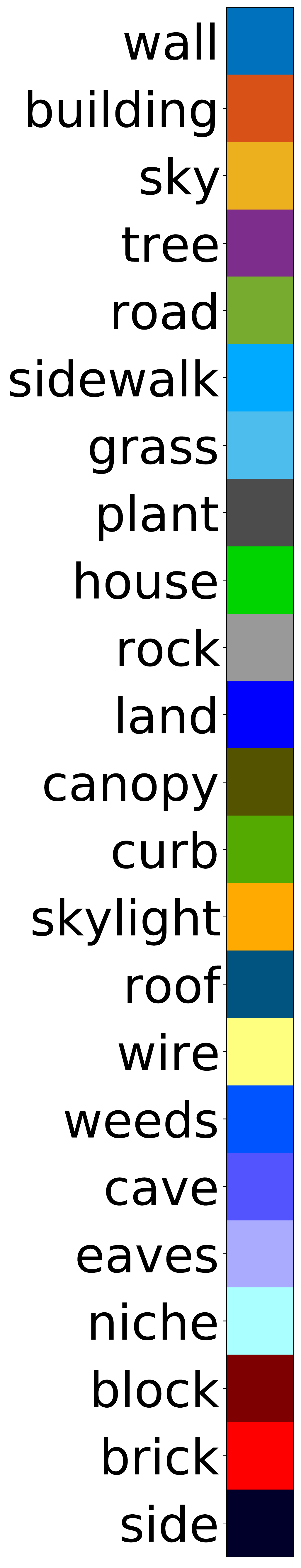}}\\
\rotatebox{90}{\bf{\makecell{Image Classes \\(Ground. mIoU)}}}&
\includegraphics[width=0.176\linewidth]{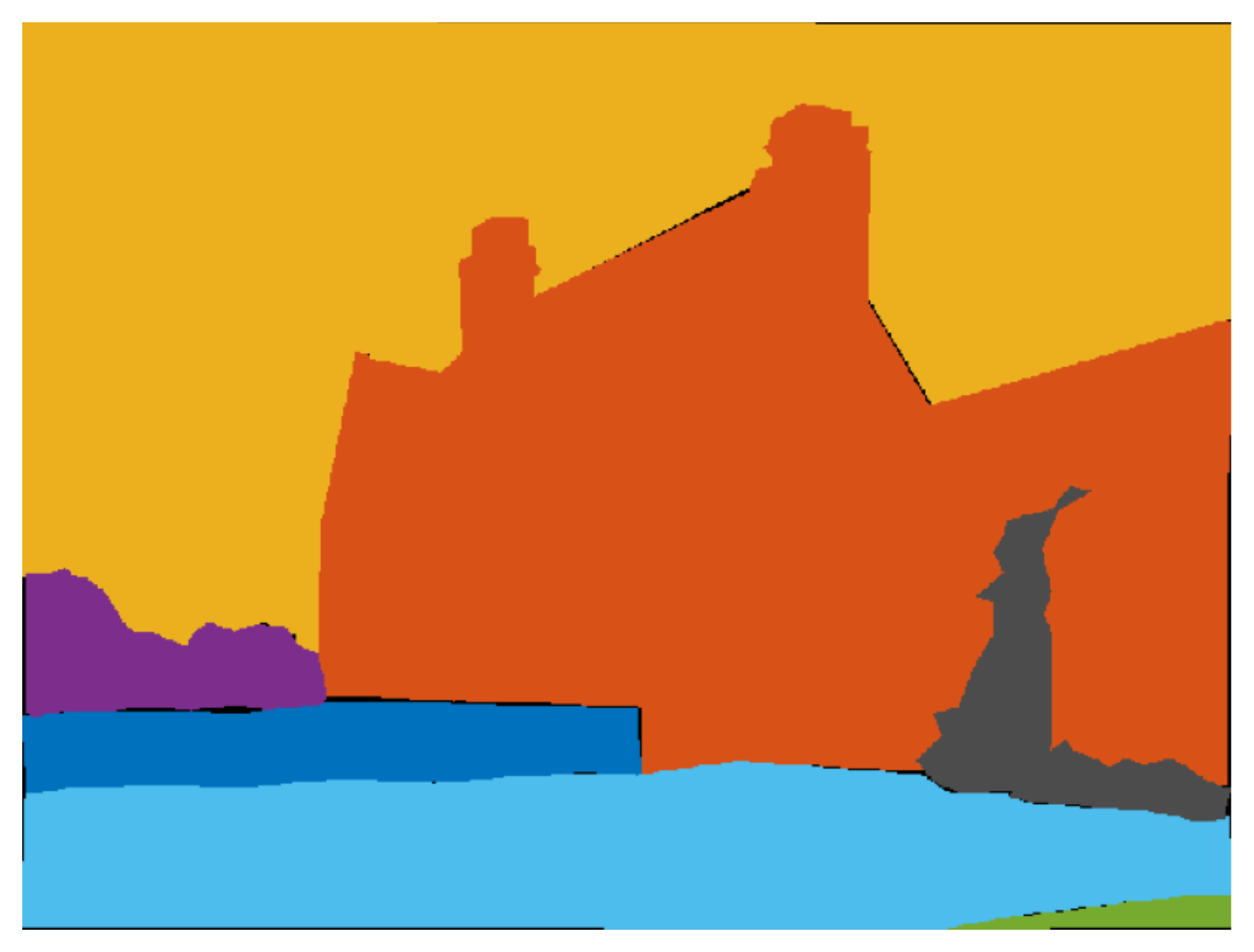}&
\includegraphics[width=0.176\linewidth]{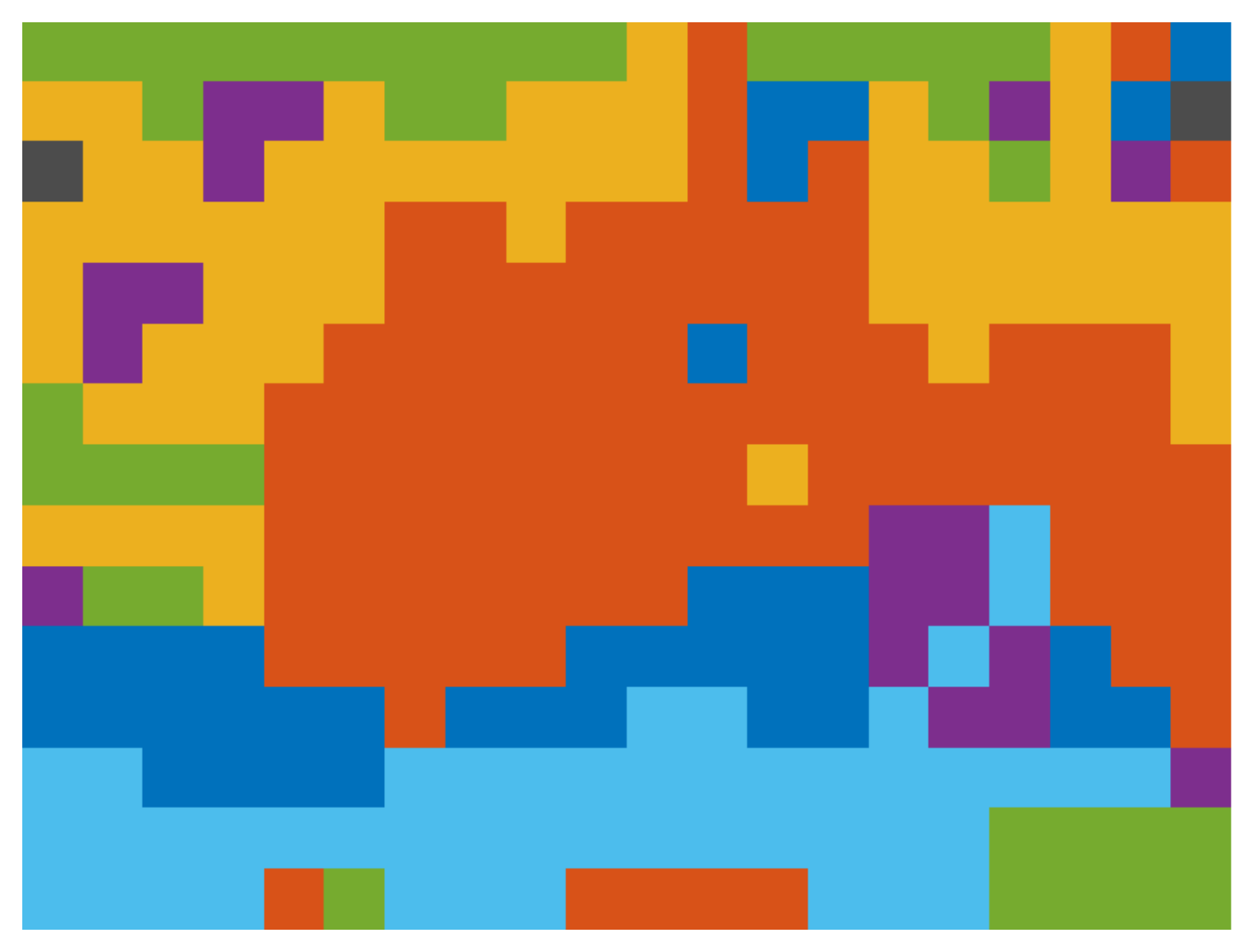}&
\includegraphics[width=0.176\linewidth]{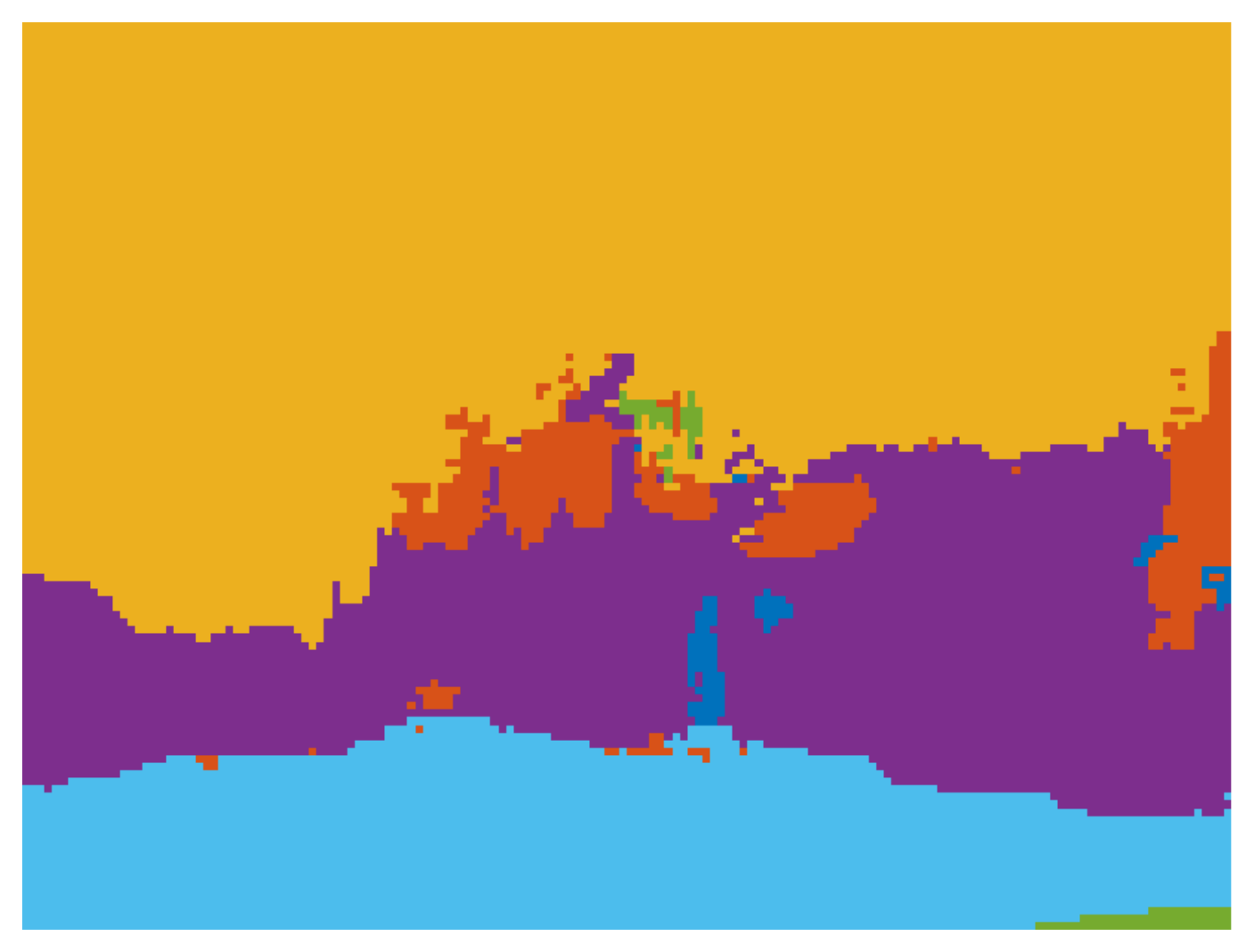}&%
\includegraphics[width=0.176\linewidth]{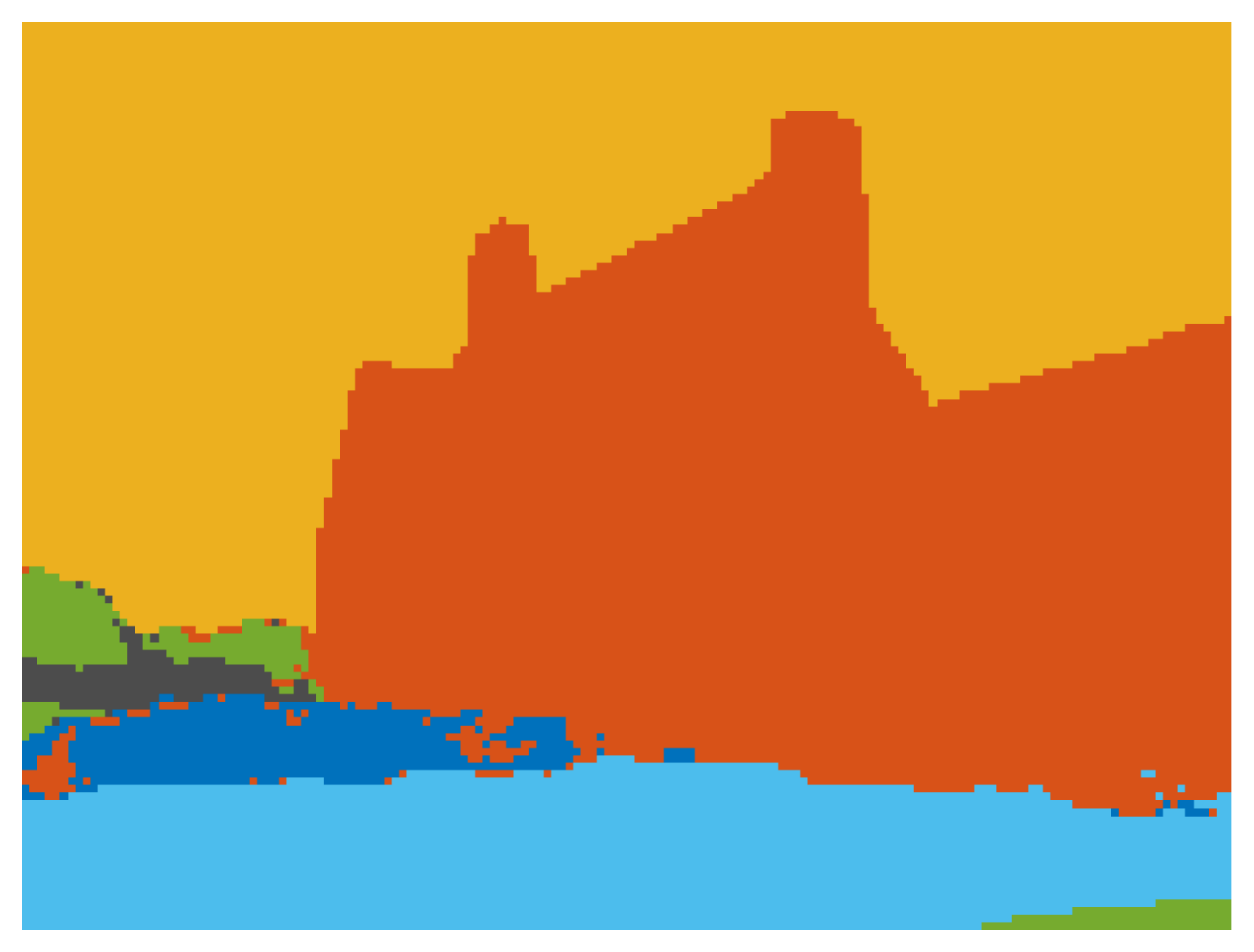}&%
\includegraphics[width=0.176\linewidth]{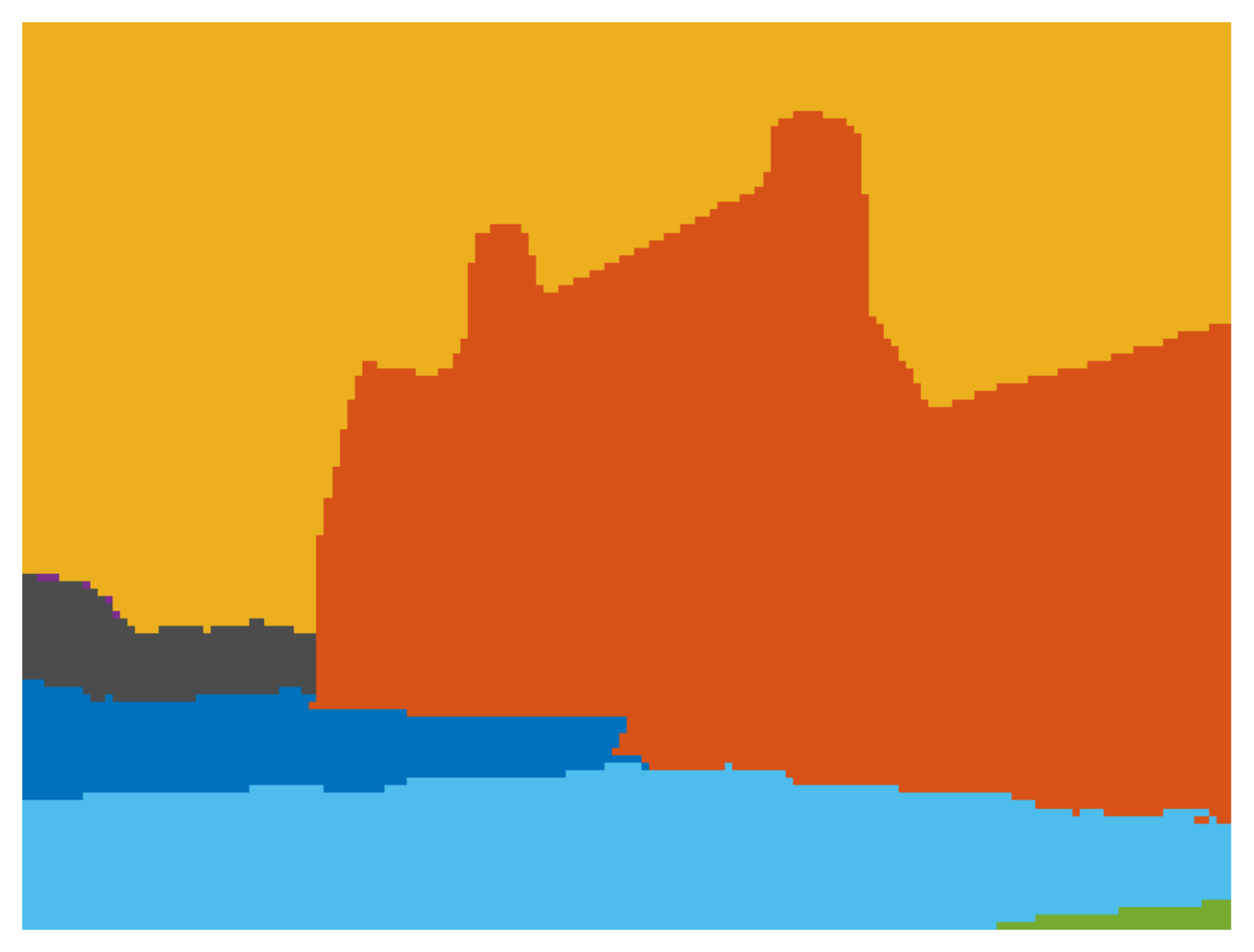}&%
\end{tabu}
\endgroup\\
\\
\begingroup
\scriptsize
\centering
\begin{tabular}{@{ }l|ccc|rrrrr|rrrrr@{ }}

& \multicolumn{3}{c|}{COCO Train}
  & \multicolumn{5}{c|}{mIoU}
  & \multicolumn{5}{c}{Grounding mIoU} \\
   & \moretiny{label} & \moretiny{mask} & \moretiny{cap.}
   & \moretiny{A-847} & \moretiny{PC-459}  & \moretiny{A-150} & \moretiny{PC-59} & \moretiny{\textcolor{gray}{COCO}}  
   & \moretiny{A-847} & \moretiny{PC-459}  & \moretiny{A-150} & \moretiny{PC-59} & \moretiny{\textcolor{gray}{COCO}}  \\
  \hline
   ALIGN                
   & \xmark & \xmark & \xmark 
   & 4.8 & 3.6 & 9.7 & 18.5  & \textcolor{gray}{15.6}
   & 17.8 & 21.8 & 25.7 & 34.2 & \textcolor{gray}{28.2}\\
   ALIGN w/proposal 
   & \xmark & \cmark & \xmark
   & 5.8 & 4.8 & 12.9 & 22.4  & \textcolor{gray}{17.9}
   & 17.3 & 19.7 & 25.3 &  32.0 & \textcolor{gray}{23.6}\\
   LSeg+
   & \cmark & \cmark & \xmark 
   & 3.8 & 7.8 & 18.0 & \textbf{46.5} & \textcolor{gray}{55.1}
   & 10.5 & 17.1 & 30.8  & 56.7 & \textcolor{gray}{60.8} \\
   \hline
   \ours~  
   & \xmark & \cmark & \cmark 
   & 6.3 & 9.0 & 21.1 & 42.1 & \textcolor{gray}{36.1}
   & 21.8 & 32.1 & 41.0 & 57.2 & \textcolor{gray}{48.2}\\
   \ours~w/L.\ Narr.
   & \xmark & \cmark & \cmark 
   & \textbf{6.8} & \textbf{11.2}  & \textbf{24.8} & 45.9 & \textcolor{gray}{38.1}
   & \textbf{25.4} & \textbf{39.0} & \textbf{45.5} & \textbf{61.5} & \textcolor{gray}{48.2}  \\
\end{tabular}
\endgroup\\
\end{tabular}

\caption{ \textbf{(Bottom) The mIoU and Grounding mIoU results of ALIGN, ALIGN w/proposal, LSeg+, and \ours.}
(Top) Segmentation predictions on an image from the ADE20k (847 categories). (First row) Predictions with all 847 classes as text queries. (Second row) Predictions with only classes in the ground-truth segmentation as text queries.
} 
\label{fig:main_result}  
\vspace{-0.2cm}
\end{figure}

\paragraph{Zero-shot transfer to ADE20k/PASCAL:}
We evaluate the performance of \ours~and the baselines on holdout image segmentation datasets whose train sets are not used for training. In Figure~\ref{fig:main_result} (bottom), we compare ALIGN, ALIGN w/proposal, LSeg+ and \ours~on the challenging \adefull~and \pascfull~datasets with large vocabularies and also on the widely used \ade~and \pasc.
In the following sections we discuss our findings based on these results. 

\paragraph{\ours~significantly outperforms pre-trained ALIGN~\cite{jia2021scaling}:}
\ours~trained on COCO outperforms ALIGN baseline on all of the benchmarks significantly. While adding proposals to ALIGN improves mIoU results. \ours~still performs significantly better. For example, on \pascfull~ \ours~outperforms ALIGN and ALIGN w/proposals by +5.4 and +4.2 mIoU, respectively.

\paragraph{Training on limited categories hurts generalization:}
LSeg+, which is trained with pixel-wise segmentation in COCO, outperforms ALIGN by a large margin on COCO (+39.5 mIOU) and \pasc~(+28.0 mIOU). Note COCO categories contain most of \pasc~categories.
However, when we evaluate LSeg+ on \adefull~which includes a larger set of vocabularies, the performance of LSeg+ is worse than ALIGN by 1.0 mIoU and 7.3 \imiou. These results demonstrate that training on the limited categories of COCO hurts the generalization of the model.

\paragraph{\ours~improves generalization:}.
While \ours~trained on COCO has worse mIoU on COCO and \pasc~in comparison to LSeg+, it generalizes better on all other benchmarks. \ours~outperforms LSeg+ by +2.5 mIoU and +11.3 \imiou~on \adefull~and also by +1.2 mIoU and +15.0 \imiou~on \pascfull. The \ours~uses class-agnostic masks and image-level caption supervision, while LSeg+ uses 134 per-pixel class name supervision. Although \ours~is trained with a weaker supervision, it has a better generalization to classes outside of COCO. These results reveal that we need open-vocabulary supervision such as captions for training a \textit{generalist} model.

\paragraph{Scaling training data with captions improves performance:}
To scale up training data we utilize the Localized Narrative dataset, which includes detailed narratives about the objects and stuff in each image. We train a segmentation teacher model on the COCO dataset and use it to generate segmentation pseudo labels on the Loc.\ Narr. dataset.
By scaling training data from 118k images to 652k images, the performance of \ours~improves on average by 2.5 mIoU and 4.8 \imiou~across 4 benchmarks (see Figure~\ref{fig:main_result}). In Appendix~\ref{sec:mask_pseudo_label},
we study the importance of using pseudo segmentation labels during training.

\paragraph{Ensembling of text queries and prompt engineering:}
To further improve the performance of \ours~we use ensembling where we include synonyms or subcategories of classes.
For example, we use `person', `child', `girl', `boy', \etc.
for the class of `person'.
We ensemble the multiple text queries by taking the max score as described in the Section~\ref{sec:inference}.
Also, since some of the class names of the segmentation datasets are not descriptive, we add a short context to the names.
\eg we change `glass' to `drinking glass'. These improvements give us on average 2.6 mIoU gain across 4 datasets (see Table~\ref{tab:sota}).
See Appendix~\ref{sec:text_ens} for more details.

\paragraph{Compare with existing methods:}
We compare \ours~with previous open-vocabulary and zero-shot segmentation methods in Table~\ref{tab:sota}.
We initialize ResNet101 backbone of \ours~and LSeg+ with ImageNet pretrained weights similar to the baselines. LSeg+ significantly outperforms LSeg (and also SPNet~\cite{xian2019semantic} and ZS3Net~\cite{bucher2019zeroshotseg}) as it is trained on the larger dataset of COCO instead of PASCAL-20. In contrast to LSeg and LSeg+ which are trained on COCO class labels, \ours~is trained on COCO captions and as a result has a better generalization. \ours~outperforms LSeg+ by +1.3 mIoU on \pascfull. Compared with GroupVit, \ours~learns visual grouping with class-agnostic segmentation, and has a superior performance. 
Also, by scaling up the training data from COCO to COCO+Loc. Narr. it achieves further gain of +1.4 on \pascfull.

For the strongest \ours~(last two rows), we initialize EfficientNet-b7 backbone with ALIGN pre-trained image encoder~\cite{jia2021scaling}. Also we train this model with sync loss (see Appendix~\ref{sec:batch_size_ablation} for more details). This model significantly outperforms the strongest LSeg model with ViT-L backbone (+19.9 mIoU on PASCAL-20).

\begin{table}[t]
\caption{\textbf{The mIoU results of our model and previous open-vocabulary and zero-shot segmentation methods.} Results for SPNet and ZS3Net on PASCAL-20 are reported from~\cite{li2022language}.
}
\vspace{-0.2cm}
\scriptsize
\centering
\begin{tabular}{l|l|c|c|rrrrr}
  & backbone
  & external dataset
  & target dataset
  & A-847 & PC-459 & A-150 & PC-59 & PAS-20 \\
  \hline

  LSeg~\cite{li2022language}
  & ViT-L/16
  & \xmark
  & \cmark (seen classes)
  & - & -& - & - & 52.3 \\
  \hline
  
  SPNet~\cite{xian2019semantic}
  & ResNet101
  & \xmark
  & \cmark  (seen classes)
  & - & -& - & 24.3 & 18.3 \\

  ZS3Net~\cite{bucher2019zeroshotseg}
  & ResNet101
  & \xmark
  & \cmark (seen classes)
  & - & -& - & 19.4 & 38.3 \\
 
  LSeg~\cite{li2022language}
  & ResNet101
  & \xmark
  & \cmark (seen classes)
  & - & -& - & - & 47.4 \\
  \hline
  
  LSeg+
  & ResNet101
  & COCO
  & \xmark
  & 2.5 & 5.2 & 13.0 & 36.0 & 59.0 \\
  
  \ours(ours)
  & ResNet101
  & COCO
  & \xmark
  & 4.0 & 6.5 & 15.3 & 36.9 & 60.0\\

  \ours(ours)
  & ResNet101
  & COCO+Loc. Narr.
  & \xmark
  & 4.4 & 7.9 & 17.5 & 40.1 & 63.8 \\  
  \hline
  
  GroupVit~\cite{xu2022groupvit}
  & VIT-S
  & CC12M+YFCC
  & \xmark
  & - & -& - & 22.4 & 52.3 \\
  \hline

  \ours(ours)
  & eff-b7
  & COCO+Loc. Narr.
  & \xmark
  & 8.1 & 11.5 & 26.4 & 44.8 & 70.2 \\
  
  +prompt eng.
  & eff-b7
  & COCO+Loc. Narr.
  & \xmark
  & \textbf{8.8} & \textbf{12.2} & \textbf{28.6} & \textbf{48.2} & \textbf{72.2} \\
  
\end{tabular}
\label{tab:sota}
\end{table}

\subsection{Ablation Experiments}

\paragraph{Importance of backbone initialization:}
In order to save the computation, we initialize \ours~from the state-of-the-art ALIGN checkpoint trained on 1.8 billion examples for image-text alignments.
In this section, we study the importance of initialization of the vision backbone from this checkpoint.
In Table~\ref{tab:backbone_init}, we compare the performance of training \ours~from scratch, initializing from the NoisyStudent checkpoint~\cite{xie2019self} and initializing from the ALIGN checkpoint. For training these models, we use the same hyper-parameters, and only tune the learning rate (0.32 for scratch, 0.08 for NoisyStudent init. and 0.005 for ALIGN init.) and number of steps (180k steps for scratch and 60k for NoisyStudent and ALIGN init.).
Table~\ref{tab:backbone_init} shows that using the NoisyStudent checkpoint to initialize the backbone achieves slightly worse results (less than 0.5 mIoU on all benchmarks) compared to using the ALIGN checkpoint. This shows initializing from the ALIGN model is not necessary for good word-region alignments. However, training from scratch is still trailing behind. We may be able to reduce the gap by increasing the batch size and training with more data. 

\begin{table}[h]
\caption{\textbf{Backbone initialization with an ALIGN pre-trained image encoder is not critical.}
The models use the pre-trained ALIGN text encoder and are trained on COCO and Loc. Narr. datasets.}
\scriptsize
\centering
\begin{tabular}{l|rrrr}
    & A-847 & PC-459 & A-150 & PC-59 \\
   \hline
   Random init.   & 4.5 & 7.6 & 18.6 & 40.6 \\ 
   NoisyStudent init.   & 6.6 & 10.7 & 24.4 & \textbf{46.9} \\
   ALIGN init.  & \textbf{6.8} & \textbf{11.2} & \textbf{24.8} & 45.9 \\
\end{tabular}
\label{tab:backbone_init}
\end{table}

\paragraph{Incorporating proposals at inference time improves accuracy:}
\label{sec:proposals_at_inference}
We are curious about the importance of mask proposals in \ours~during inference. To study this problem, we take the feature map $\mathcal{F}_z$ in \ours~and perform per-pixel segmentation by taking the dot product of $\mathcal{F}_z$ with word embeddings $\mathbf{w}$. This method performs inference without mask proposals. 
In Table~\ref{tab:proposals_at_inference}, we compare the performance of \ours~and its counterparts that do not use mask proposals (the above method) or using ground-truth as mask proposals.
The performance of \ours~is much worse if not using proposals: mIoU on \pasc~drops from 42.1 to 32.1 and from 21.1 to 16.4 on \ade. Using ground-truth as proposals can be seen as an upper bound when we have perfect class-agnostic localization. The results show the room for improving localization. It also demonstrates even with perfect localization, the semantic alignment is still challenging.

\begin{table}[t]
\scriptsize
\centering
\caption{\textbf{Incorporating predicted masks at inference improves mIoU accuracy.} Using the ground-truth masks can be seen as the performance upper bound when segmentation masks are perfectly predicted. The model is trained on COCO.}
\begin{tabular}{l|rrrr}
    & A-847 & PC-459 & A-150 & PC-59 \\
   \hline
   \ours~            & 6.3 & 9.0  &  21.1 & 42.1 \\ 
    - pred. masks         & \minus{1.7}4.6  & \minus{3.1}5.9   &  \minus{4.7}16.4  & \minus{10.0}32.1  \\
    + gt. masks       & \plus{2.8}9.1   & \plus{3.3}12.3 &  \plus{6.4}27.5 & \plus{7.2}49.3 \\
\end{tabular}
\label{tab:proposals_at_inference}  
\end{table}

\begin{table}[t]
\caption{\textbf{Using all words in training captions hurts performance.}
Using nouns+adj for training achieves the best results. The model is trained on COCO.
}
\scriptsize
\centering
\begin{tabular}{l|rrrr}
    caption filter  & A-847 & PC-459 & A-150 & PC-59 \\
   \hline
   all words                  & 5.3 & 8.8 &   20.0 & 41.3 \\
   noun + adj. + verb   & 6.0 & 8.8 &  20.9 & 41.7 \\
   noun + adj                 & \textbf{6.3} & \textbf{9.0}  &  \textbf{21.1} & \textbf{42.1} \\

\end{tabular}
\label{tab:text_filtering}
\end{table}

\paragraph{Importance of text filtering:}
We train \ours~with image captions which may include words that do not represent any regions in an image. These noises make training more challenging. We perform a simple pre-processing on the captions and extract the list of nouns and adjectives. This procedure removes conjunctions, pronouns, adverbs, verbs, \etc. which reduces the noises.
In Table~\ref{tab:text_filtering}, we study the performance of \ours~when using different types of filtering on the captions. Keeping only nouns and adjectives yields the best results. The worst results are from using all words, which show 0.2-1.1 worse mIoU. The small performance differences across different ways of text filtering show \ours~is robust to the noise in the input words to some degree.

\section{Conclusion}
We propose \ours, an open-vocabulary image segmentation model, to organize an image into regions described with arbitrary text queries. This is in stark contrast to previous works in semantic segmentation learned to predict categories in closed vocabulary. We propose to represent an image with a set of mask regions followed by visual-semantic alignments. Such representations support weakly-supervised learning for grounding words in a caption to predicted mask proposals, and thus make the training data scalable. We are the first work to directly evaluate on holdout image segmentation datasets, attaining significant performance gains against strong baselines initialized by a pre-trained ALIGN model. We hope to encourage future works to learn a \textit{generalist} segmentation model that can transfer across datasets using language as the interface.

\clearpage
%
%
\bibliographystyle{splncs04}
\bibliography{main}

\clearpage

\begin{appendix}

\section{Potential societal impacts}
\ours~provides a language interface for recognition and localization of visual concepts. Such an interface may facilitate downstream applications such as interactive intelligent home assistants, interactive content creation, and instructing robot actions with language~\cite{shridhar2021cliport}. However, \ours~models are trained on large-scale datasets, which may contain bias towards certain image and text distributions. As a result, we share the similar concern as recent studies for evaluating image-text models like Clip~\cite{agarwal2021evaluat_clip}. 
Thus \ours~is not suitable for deployment in the real world without properly studying the model biases and calibrating the predictions.

\section{Limitations of our approach}
The limitations come with the strength of our method. The mask representations can organize an image into a small number of regions, and thus enable scalable visual-semantic alignments. 
However,
our method will not be able to segment visual concepts that do not have associated segmentation proposals.
That being said, we still believe the generalization to unseen concepts with class-agnostic mask representations is easier than pixel-wise representations. How to improve generalization or adaptation of mask predictions will be an interesting research topic.

\section{Architecture of the cross-attention module}
\label{sec:cross_attention_arch}
\begin{figure}[b]
    \centering
    \includegraphics[width=1.0\linewidth]{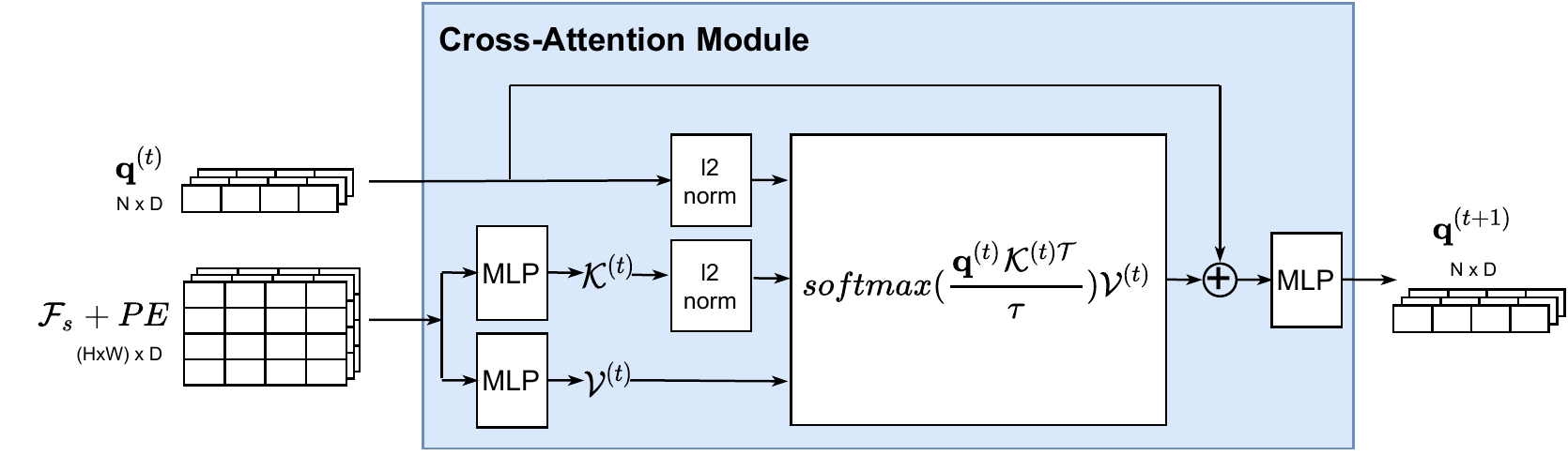}
\caption{\textbf{Model architecture of the region-image cross-attention module.} The module is stacked repeatedly to generate mask queries for region segmentation.}
\label{fig:cross_attention}
\end{figure}

Figure~\ref{fig:cross_attention} shows the model architecture of the cross-attention module. The mask queries $\mathbf{q}^{(t)}$ interact with the position-augmented image features $\mathcal{F_s} + PE$ to generate mask queries $\mathbf{q}^{(t+1)}$. The first mask queries $\mathbf{q}^{(0)}$ is randomly initialized at the beginning of training. The module is stacked three times ($T=3$) in our experiments. We try to include self-attention of queries followed by the query-image cross-attention as in~\cite{cheng2021maskformer}, but that does not improve performance.

\begin{table*}[tb]
\caption{\textbf{Our mask prediction can generalize across datasets}. We report the   recall at IoU 0.5.}
\small
\centering
\begin{tabular}{lrrrrrrr}
  Train / Test & ~COCO & ~ADE20K & ~Mapillary & ~IDD & ~BDD & ~Cityscapes & ~SUN \\
  \hline
  Cityscapes & 28.8 & 25.8 & 42.5 & 50.0 & 61.2 & 69.3 & 22.4 \\
  COCO       & 82.2 & 68.0 & 48.6 & 58.1 & 64.2 & 58.6 & 90.8 \\
  MSeg       & 82.9 & 80.3 & 55.8 & 68.1 & 73.4 & 67.2 & 93.6 \\
\end{tabular}
\label{tab:mseg}  
\end{table*}

\section{Mask generalization on MSeg dataset}
\label{sec:mseg_experiments}
In this experiments, we use the setup and the curated annotations in MSeg~\cite{lambert2020mseg}. Our goal is to verify if a model trained on a single dataset can generalize to multiple datasets. Table~\ref{tab:mseg} summarizes the results of recall at an IoU of 0.5. 
The model achieves the best results when trained on MSeg, which aggregates the training images and annotations of all datasets, and can be seen as the performance upper bound.
The results are slightly worse when trained on COCO, showing the model trained on COCO can generalize reasonably well. The model trained on Cityscapes generalizes poorly, indicating \ours~requires a large and diverse training dataset to provide class-agnostic segmentation proposals.

\section{Visualization of full mask proposals}
\label{sec:visualization_of_full_proposals}
\begin{figure*}[ht!]
    \centering
    \includegraphics[width=1.0\linewidth]{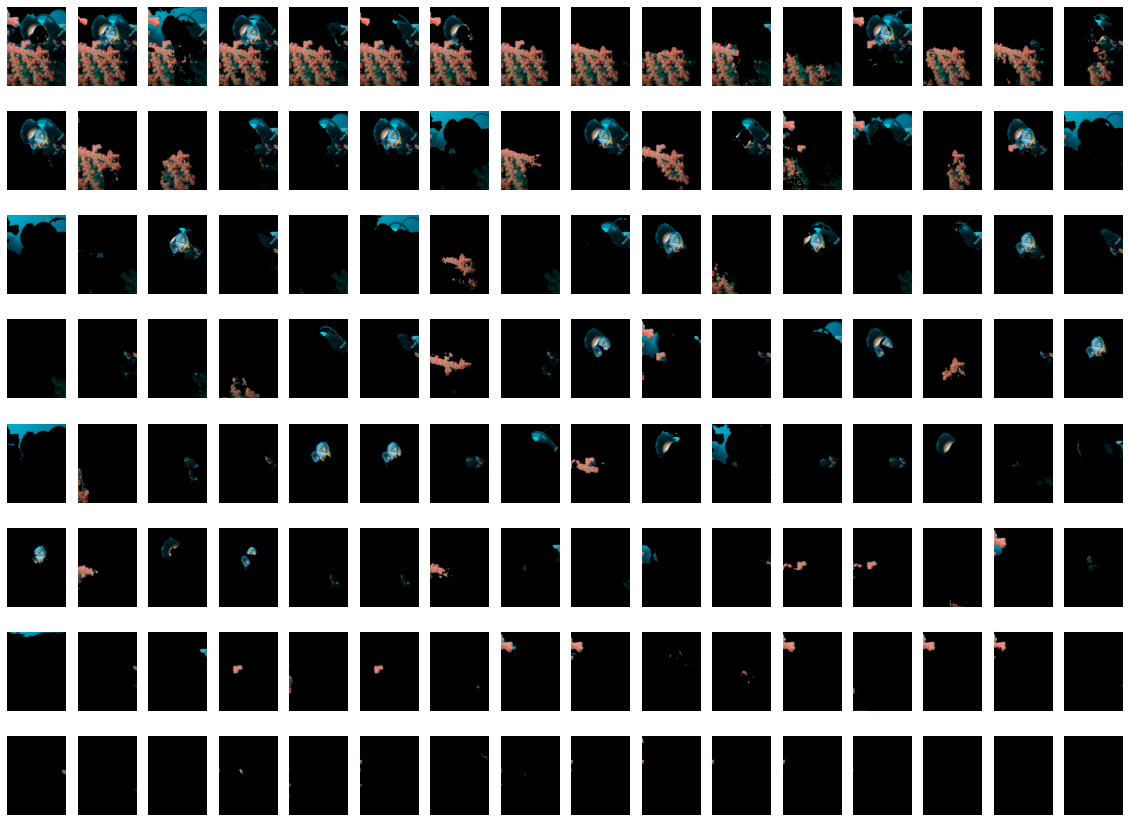}
\caption{\textbf{Full set of predicted segmentation masks.} This model is trained to predict 128 segmentation masks.}
\label{fig:full_regions}
\end{figure*}

In Figure~\ref{fig:proposals} of the main paper we present a subset of predicted segmentation masks in an unseen scene. Figure~\ref{fig:full_regions} shows all the 128 mask proposals on the same image.

\section{Ablation on batch size}
\label{sec:batch_size_ablation}
In all of the experiments unless otherwise stated, we use a global batch size of 1024 and our local batch size is 16 (we have 64 cores). Also, we compute the contrastive loss over the local examples of each core. Therefore, the effective batch size of loss is equal to local batch size. In this section, we compare \ours~with unsync and  sync contrastive loss. In the sync version, we compute the loss over all cores and effective batch size of loss is equal to global batch size.
As shown in Table~\ref{tab:batch_size}, sync loss improves the performance of \ours. We also train \ours~with different batch sizes (same epochs). We find that OpenSeg is robust to the batch size.

\begin{table}[tbh]
\caption{\textbf{\ours~is robust to the batch size.} We present performance of \ours~trained on COCO+Loc. Narr. and different batch sizes. Numbers inside the parentheses represent effective batch size for the contrastive loss. }
\footnotesize
\centering
\begin{tabular}{l|l|rrrr}
    batch size  & steps &A-847 & PC-459 & A-150 & PC-59   \\
    
   \hline 
   1024 (16) & 60k & 6.8 & 11.2 & 24.8 & 45.9 \\
   \hline
   256 (256)    & 240k & 8.1 & 11.2  &  26.2 & 45.4 \\
   512 (512)   & 120k & 8.2 & 11.2  &  25.7 & 45.0 \\
   1024 (1024)   & 60k  & 8.1 & 11.5  &  26.4 & 44.8 \\
   2048 (2048)   & 30k  & 8.4 & 11.4  &  26.9 & 45.3 \\
\end{tabular}
\label{tab:batch_size}  
\end{table}

\section{Importance of segmentation loss}
\label{sec:mask_pseudo_label}
Since grounding loss $\mathcal{L_G}$ back-propagates through the segmentation head, it is possible to train the complete \ours~model with only the grounding loss.
In this section we study the importance of having segmentation loss in addition to the grounding loss in training \ours. In Table~\ref{tab:importance_of_pseudo_labels}, we compare the performance of \ours~when it is trained with only grounding loss (second row) vs when it is trained with both losses (first row). The former model has significantly worse performance which illustrates the importance of segmentation loss in learning the visual grouping.

When training with COCO + Loc.\ Narr.,
we annotate Loc.\ Narr. dataset with mask pseudo labels (Section~\ref{sec:open_vocab_segm}), so that we can compute both grounding and segmentation losses on all training examples. 
We study the importance of the pseudo labels and the segmentation loss by setting the weight of segmentation loss to zero on examples from Loc.\ Narr.
In Table~\ref{tab:importance_of_pseudo_labels}, we compare the performance of these two approaches. The model trained with mask pseudo labels (third row) has worse mIoU on \pasc. However, it has better performance on \adefull{}, \pascfull{} and \ade{} datasets, which include categories outside the COCO dataset. These results indicate that adding mask pseudo labels and computing the segmentation loss on all of the training examples helps the generalization of \ours.

\begin{table*}[htb]
\caption{\textbf{Grounding loss is not sufficient for training \ours.} We experiment with setting segmentation loss to zero on COCO examples when training on COCO, or on Loc.\ Narr.\ examples when training on COCO + Loc.\ Narr. In both cases the performance of the model drops.}
\scriptsize
\centering
\begin{tabular}{l|rr|rr|rrrr}
    & \multicolumn{2}{c|}{Segmentation loss} \\
  
    & COCO & Loc.\ Narr.
    & A-847 & PC-459 & A-150 & PC-59 \\
   \hline

   \ours(COCO) &
   \cmark & -
   & 6.3 & 9.0  &  21.1 & 42.1 \\ 

  \ours(COCO) &
  \xmark & -
  & \minus{3.8}2.5 & \minus{5.7}3.3  &  \minus{14.6}6.5 & \minus{26.5}15.6 \\
   
   \hline
   
   \ours(COCO + Loc.\ Narr.) &
   \cmark & \cmark
   & 6.8 & 11.2 & 24.8 & 45.9 \\ 
   
   \ours(COCO + Loc.\ Narr.) &
   \cmark & \xmark 
   & \minus{0.5}6.3  & \minus{0.8}10.4   &  \minus{2.4}22.4 & \plus{1.4}47.3 \\
   
\end{tabular}

\label{tab:importance_of_pseudo_labels}  
\end{table*}

\section{Ablation on randomly dropping words}
\label{sec:dropping_words}
In our experiments, we extract the list of nouns from the captions and then we keep each word by the probability of 0.75. In Table~\ref{tab:keepprob} we compare the performance of \ours{} with different keeping probabilities. We obtain the best results when the keep prob is 0.5 or 0.75, which shows that randomly dropping words within a certain probability range prevents overfitting and improves the performance.

\begin{table}[t!]
\caption{\textbf{Randomly dropping words improves performance of \ours.} We extract the list of nouns from captions and keep each noun by a probability of $kp$. We get the best results with $kp=0.5/0.75$.}
\small
\centering
\begin{tabular}{l|rrrr}
    & A-847 & PC-459 & A-150 & PC-59 \\
    \hline
   $kp = 1.0$  & 5.8          & 10.4          & 22.7          & 44.9 \\
   $kp = 0.75$ &\textbf{6.8}  & \textbf{11.2} & \textbf{24.8} & \textbf{45.9} \\
   $kp = 0.5$  &\textbf{7.0}  & 10.7          & \textbf{25.0} & \textbf{46.0} \\
   $kp = 0.25$ &\textbf{6.8}  & 9.8           & 22.4          & 43.5 \\
\end{tabular}
\label{tab:keepprob}
\end{table}

\section{Ensembling and prompt engineering}
\label{sec:text_ens}
In this section, we study how we can further improve the performance of \ours{} by prompt engineering and ensembling with the class names provided by testing segmentation datasets.

\paragraph{Ensembling:} 
An object can often be referred with more than one possible description. Some of them exist in the testing dataset, and some do not. For example, image captions usually include one of the descriptions of `man, woman, boy, girl, \etc' when referring to the `person' category in the testing segmentation datasets. Thus to further improve the performance, we manually assemble a list of synonyms, subcategories and plurals for some of the categories. Here are a few examples:
\begin{itemize}
\itemsep0em 
    \item person $\rightarrow$ person, child, girl, boy, woman, man, people, children, girls, boys, women, men
    \item dog $\rightarrow$ dog, puppy, dogs, puppies
    \item cat $\rightarrow$ cat, kitty, cats, kitties
    \item grass $\rightarrow$ grass, grasses, lawn, turf 
    \item bottle $\rightarrow$ bottle, bottles, water bottle, water bottles
\end{itemize}

\paragraph{Polysemy:} Some class names of the segmentation datasets are polysemous. As a result, a model may make predictions for different meanings of a concept, while the dataset only includes one of the meanings. For example, the class `fan' in ADE20k refers to a cooling machine, but \ours{} sometimes labels a crowd of fans (people) on a stage watching a game as `fan'.  To resolve this issue, we add a short context to some of the labels. \eg, we change `fan' to `ceiling fan, floor fan'.

\paragraph{Categories have overlap:} Another challenge is that some of the classes of a dataset may have overlap. Although the annotators may follow some rules that prevent the overlap. An example of this can be seen in the Figure~\ref{fig:main_result}. \adefull~has both `roof' and `building' categories, and both of them are correct labels for the `roof' region in this figure. Another example is the `clothes' and `person' categories in the \ade~and \pasc~datasets, where a model is penalized for predicting the `clothes' category on a person. This issue happens more frequently as the vocabulary size gets larger. We don't have a good solution for this issue. However, in addition to the mIoU metric, we calculate the \imiou~metric that has less of this issue.

Table~\ref{tab:text_prompt} provides the gains that we attain by applying ensembling and prompt engineering. Overall, the improvement is less significant when we scale up training data from COCO to COCO+Loc.\ Narr.\ (+2.4 \vs +1.8 on average across 4 benchmarks). Moreover, since there is less ambiguity in terms of class names for the Grounding mIoU, the improvement is smaller for this metric in comparison to mIoU (+1.2 vs +1.8 on average across 4 benchmarks when we train on COCO+Loc.\ Narr.).
We will open source our modified list of class names used for this experiment.

\newcommand{\plust}[1]{
{\begin{tiny}\textcolor{blue}{+#1}\end{tiny}}
}
\newcommand{\minust}[1]{
{\begin{tiny}\textcolor{red}{-#1}\end{tiny}}
}

\begin{table}
\caption{\textbf{Ensembling and prompt engineering improves performance of \ours.}}
\centering
\scriptsize
\begin{tabular}{l|c|rrrr|rrrr}
 \multicolumn{1}{c|}{ } &
 \multicolumn{1}{c|}{ } &
 \multicolumn{4}{c|}{mIoU} &
 \multicolumn{4}{c}{\imiou{}} \\
  Training data &  eng. &
  A-847 & PC-459 & A-150  & PC-59 &
  A-847 & PC-459 & A-150  & PC-59   \\
  \hline
  
  COCO & \xmark
  & 6.3 & 9.0 & 21.1 & 42.1 
  & 21.8 & 32.1 & 41.0 & 57.2 \\
  
  COCO & \cmark
  & \plust{0.5}6.8  & \plust{1.0}10.0 & \plust{3.7}24.8 & \plust{4.3}46.4
  & \plust{0.6}22.4 & \plust{0.6}32.7 & \plust{2.5}43.5 & \plust{3.7}60.9 \\
  
  \hline
  
  COCO+L.Narr. & \xmark
  & 6.8 & 11.2 & 24.8 & 45.9
  & 25.4 & 39.0 & 45.5 & 61.5 \\
  
  COCO+L.Narr. & \cmark
  & \plust{0.8}7.6 & \plust{0.6}11.8 & \plust{2.9}27.7 & \plust{3.1}49.0 
  & \plust{0.1}25.5 & \plust{0.7}39.7 & \plust{1.3}46.8 & \plust{2.7}64.2 \\
  \hline
\end{tabular}
\label{tab:text_prompt}  
\end{table}

\section{Visualization of segmentation predictions}
\begin{figure*}%
\centering%
\setlength\tabcolsep{0.5pt}
\renewcommand{\arraystretch}{0.2}
\begin{tabular}{rlrl}%
\centering%
&%
\includegraphics[height=0.255\linewidth]{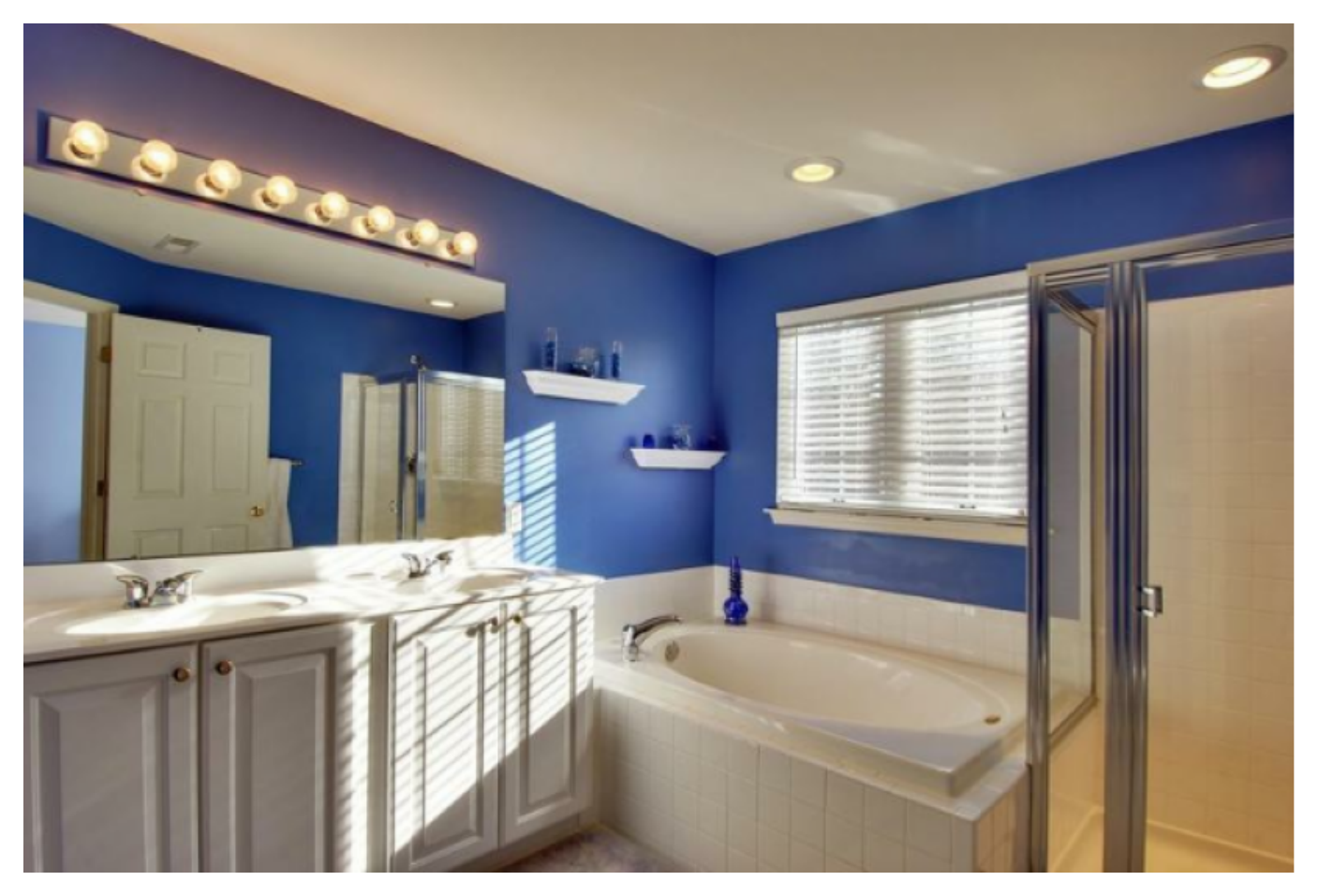}&%
\includegraphics[height=0.245\linewidth]{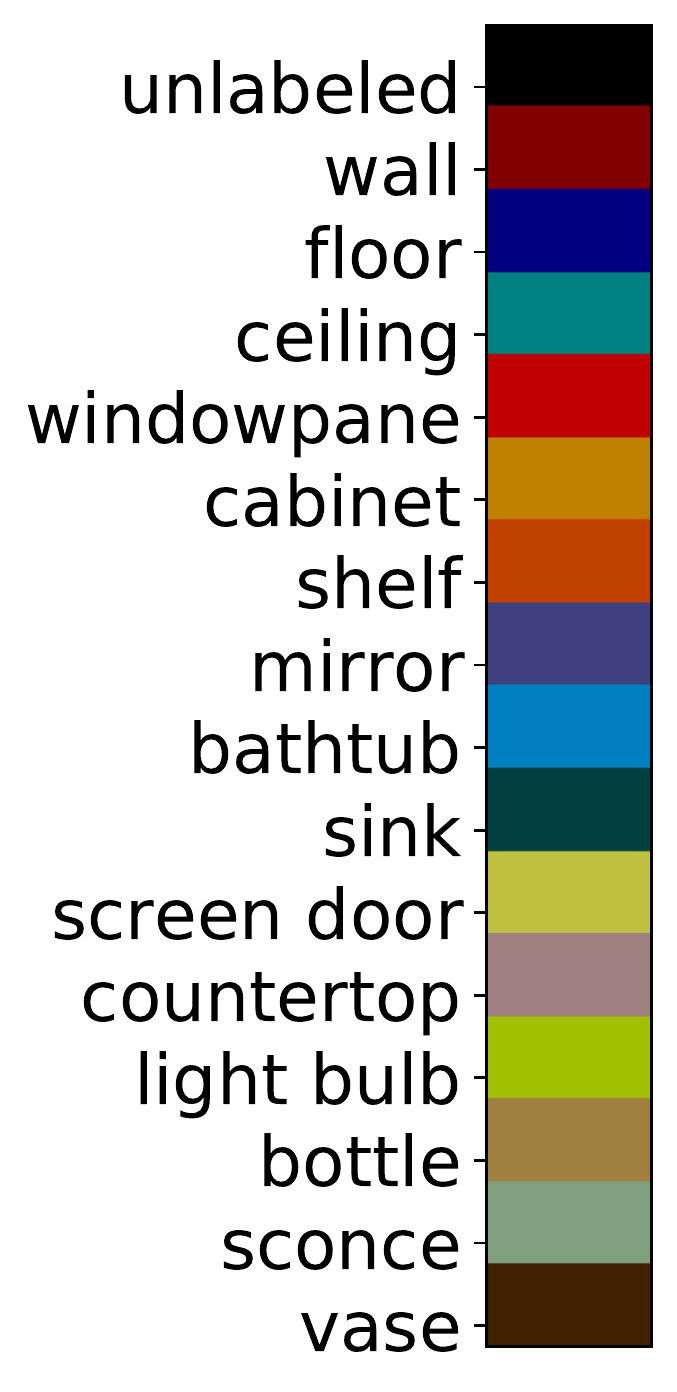}&%
\includegraphics[height=0.255\linewidth]{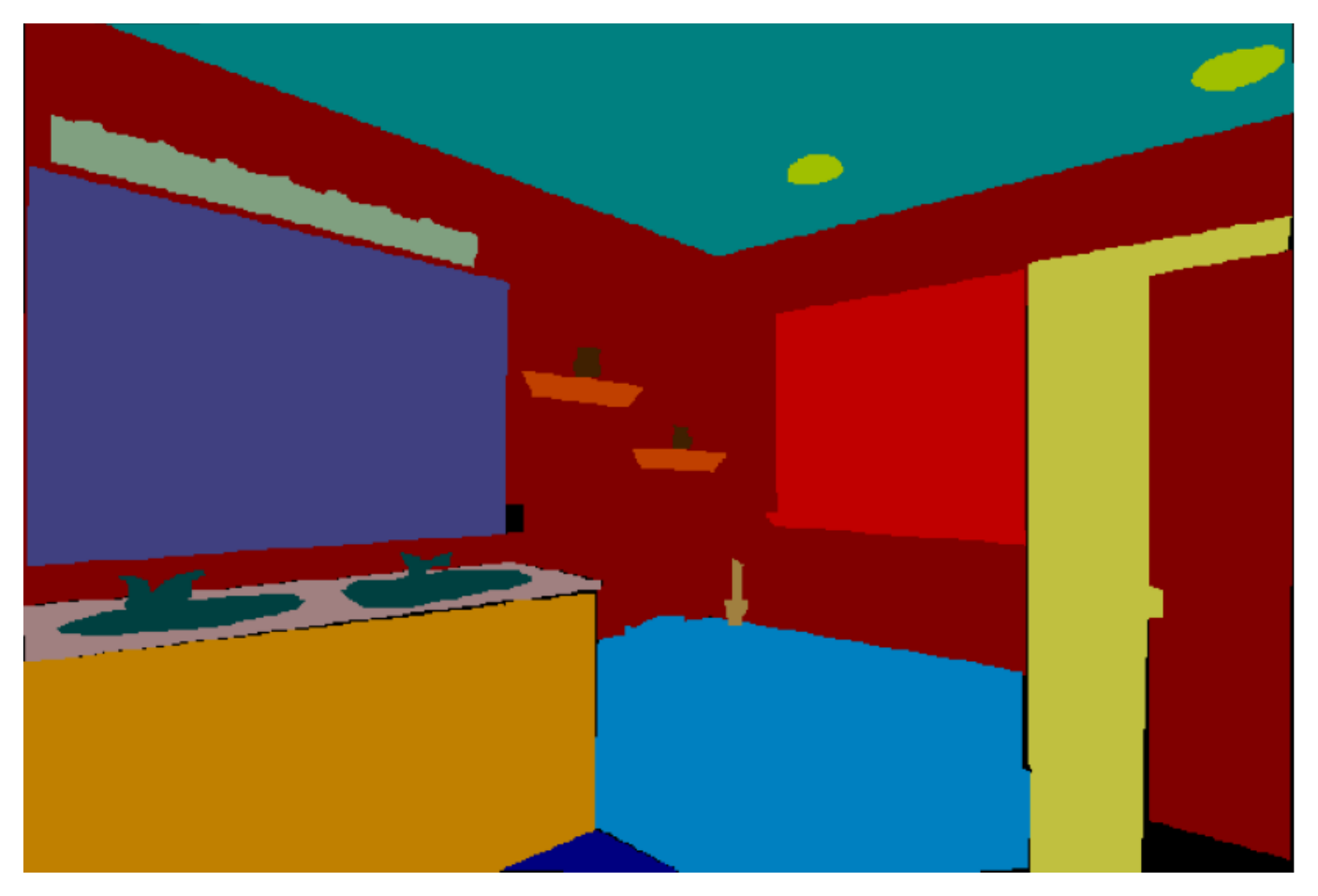}\\%
\includegraphics[height=0.255\linewidth]{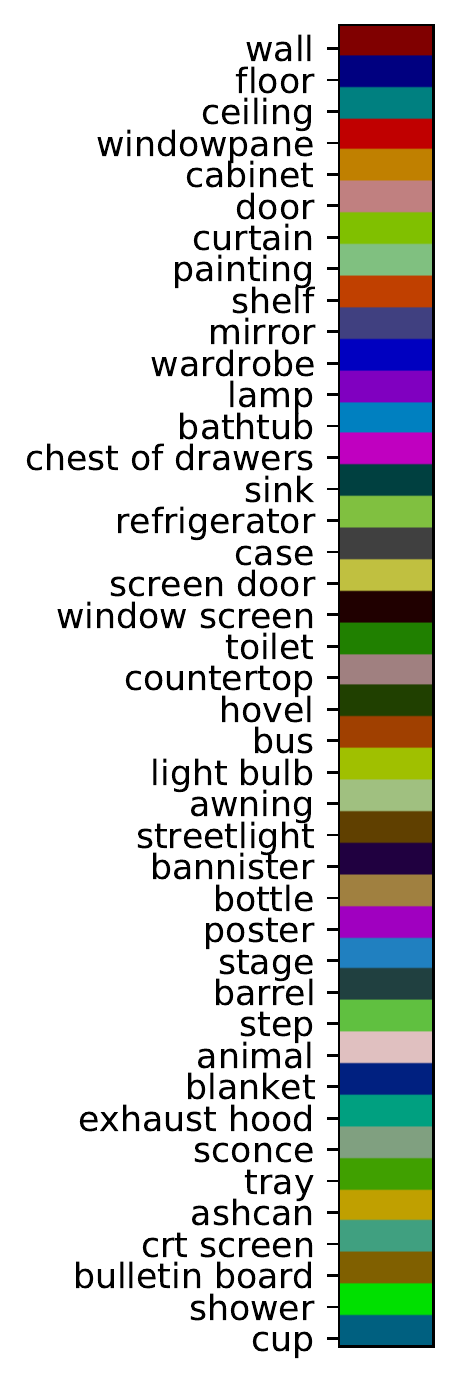}&%
\includegraphics[height=0.255\linewidth]{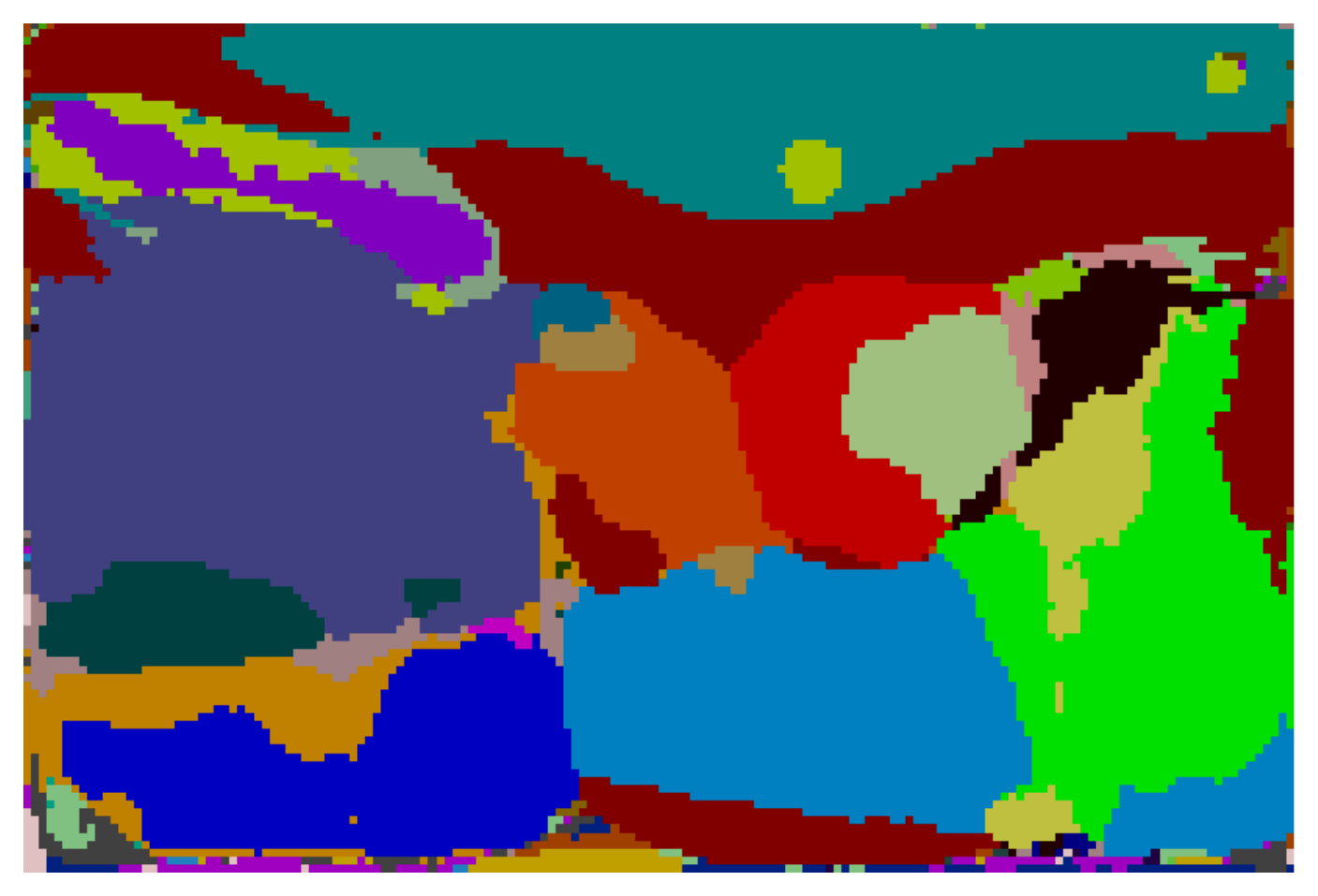}&
\includegraphics[height=0.255\linewidth]{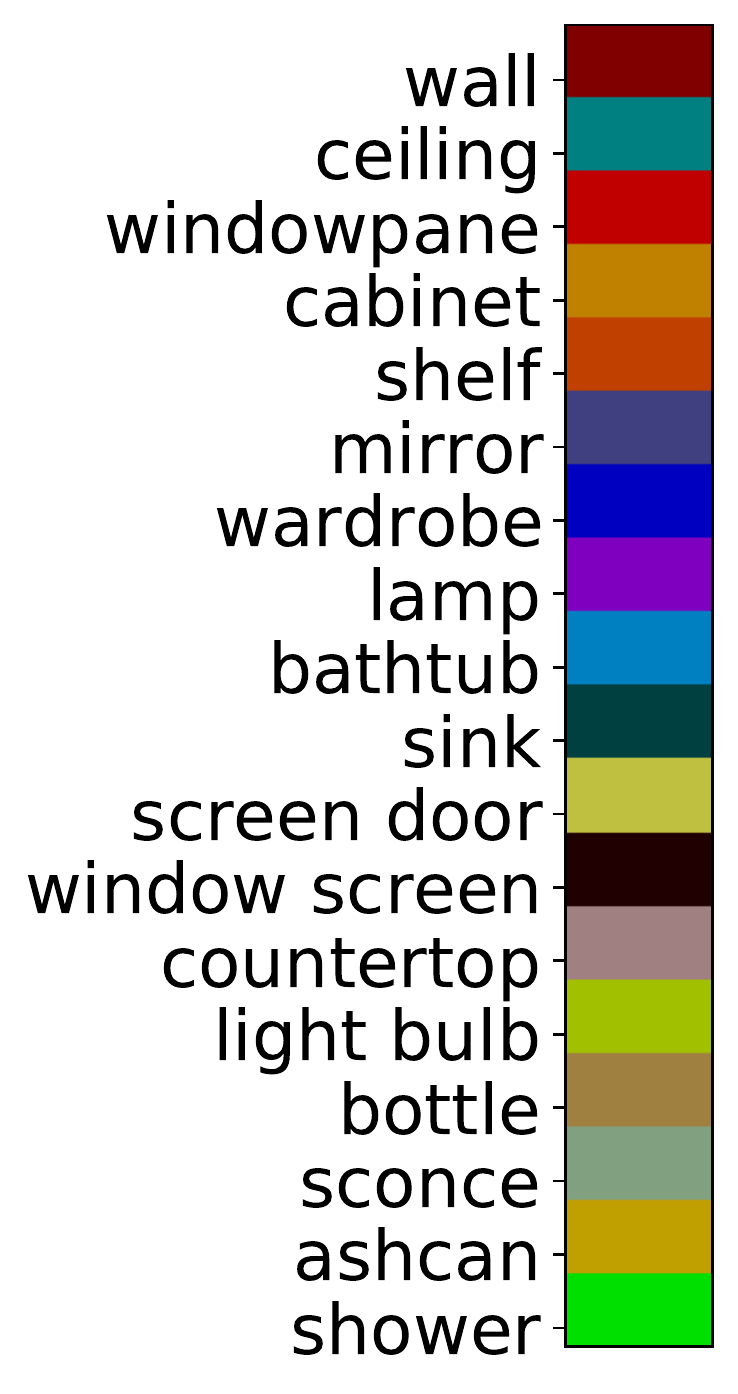}&%
\includegraphics[height=0.255\linewidth]{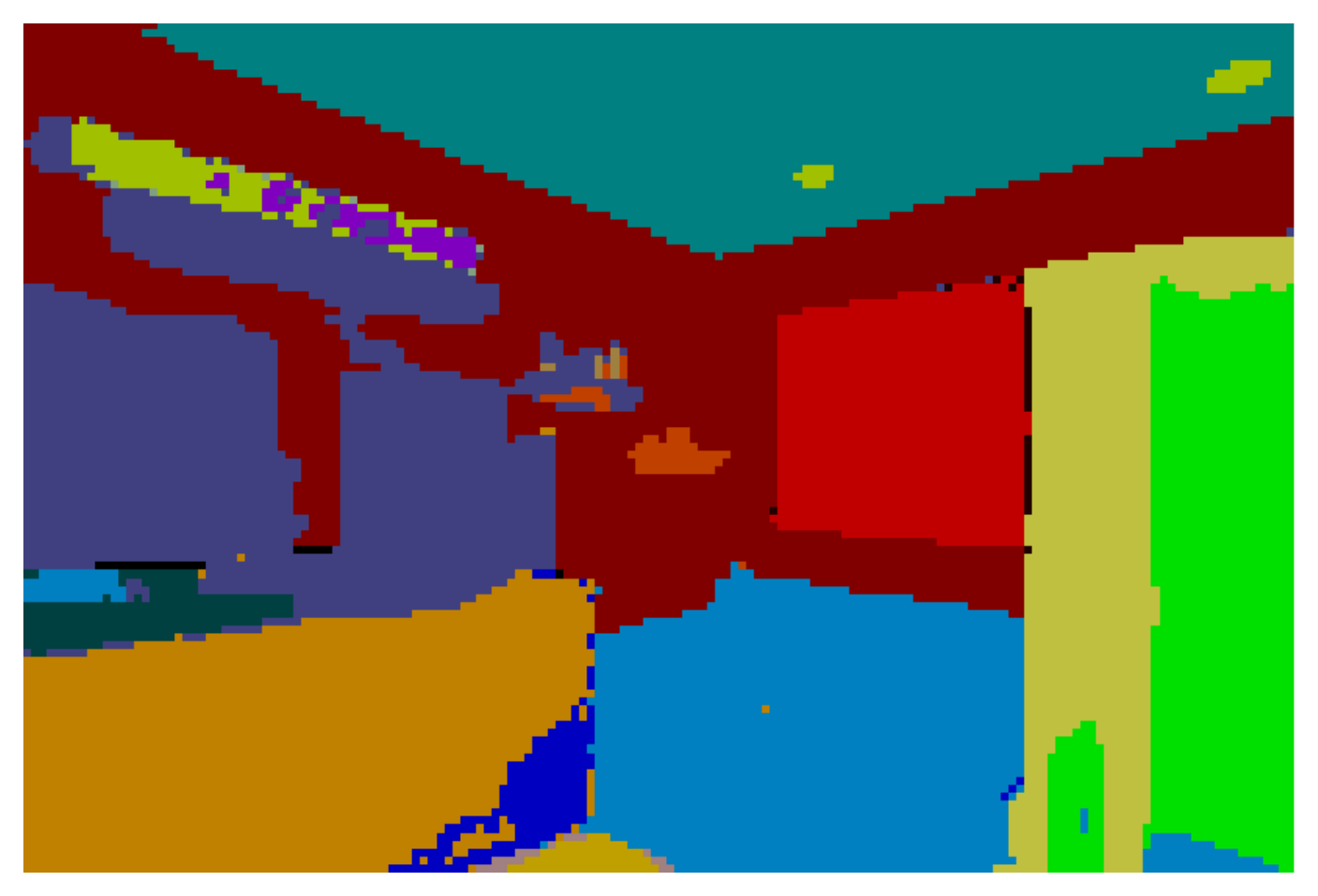}\\%
\vspace{1cm}\\
&%
\includegraphics[height=0.265\linewidth]{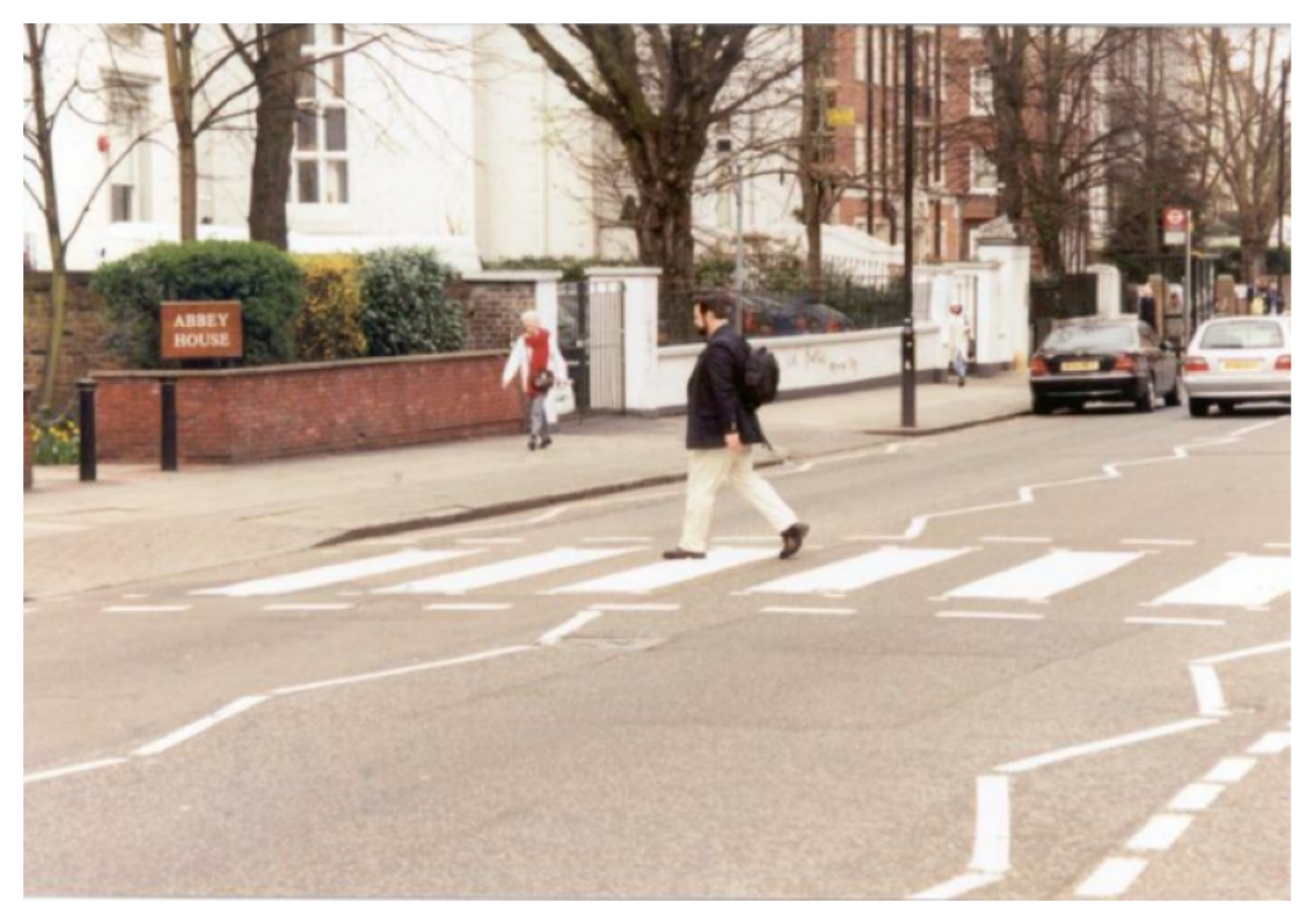}&%
\includegraphics[height=0.19\linewidth]{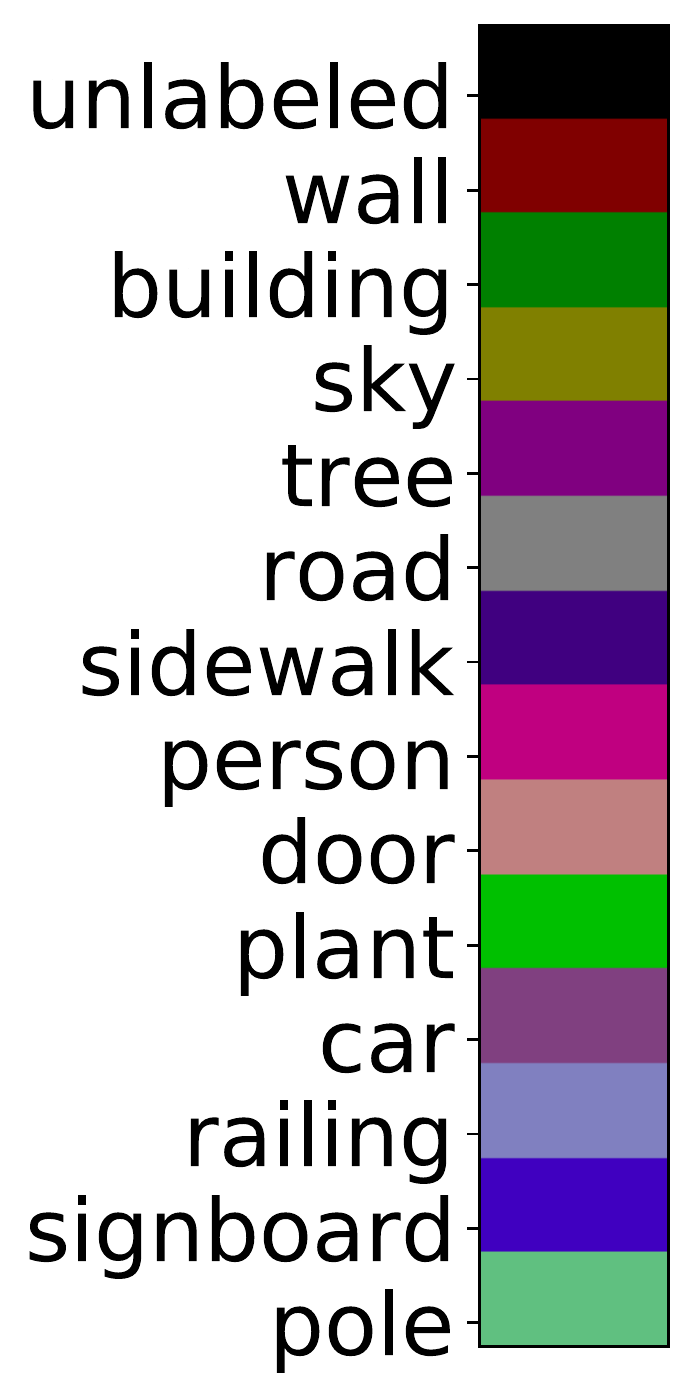}&%
\includegraphics[height=0.265\linewidth]{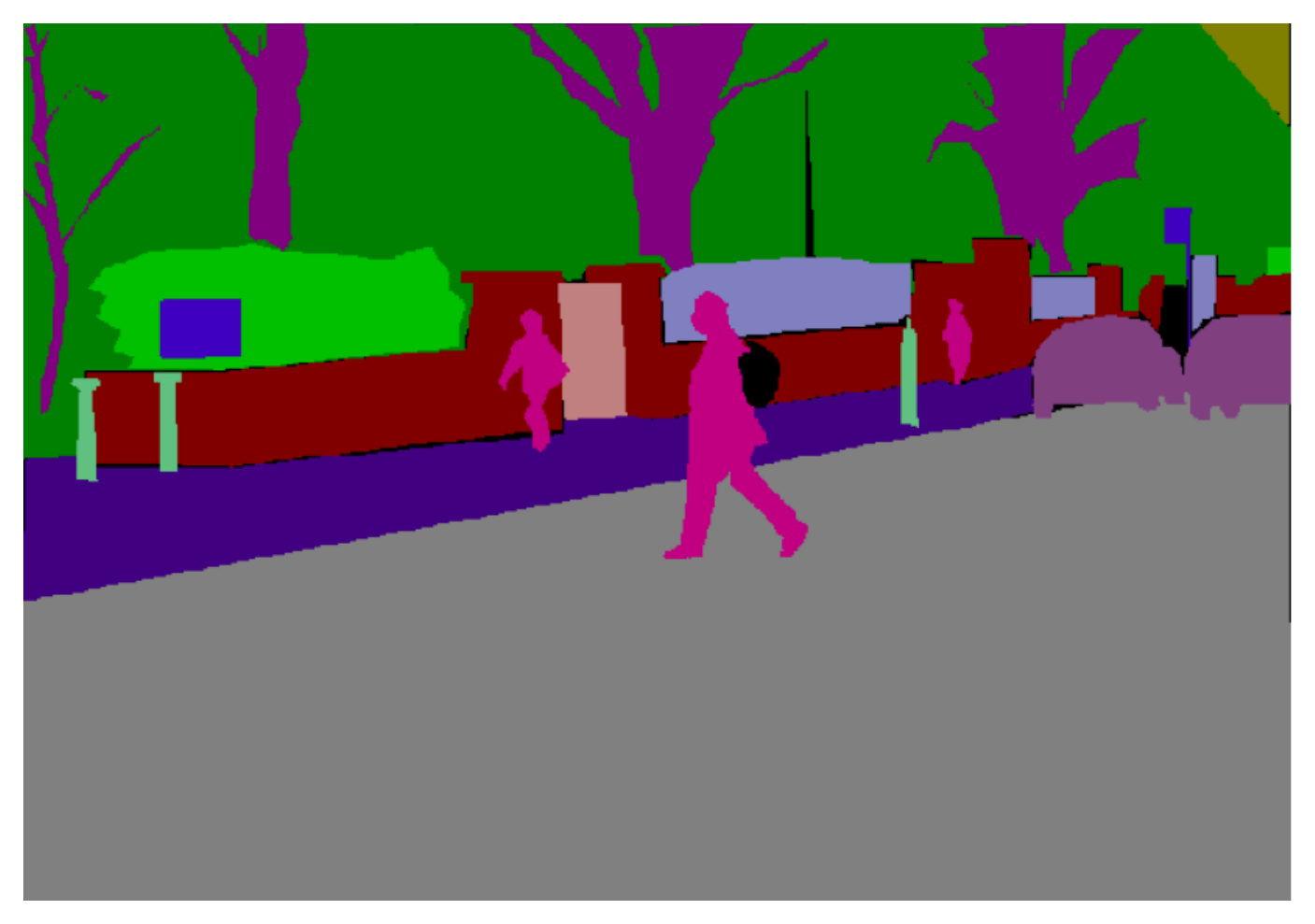}\\%
\includegraphics[height=0.265\linewidth]{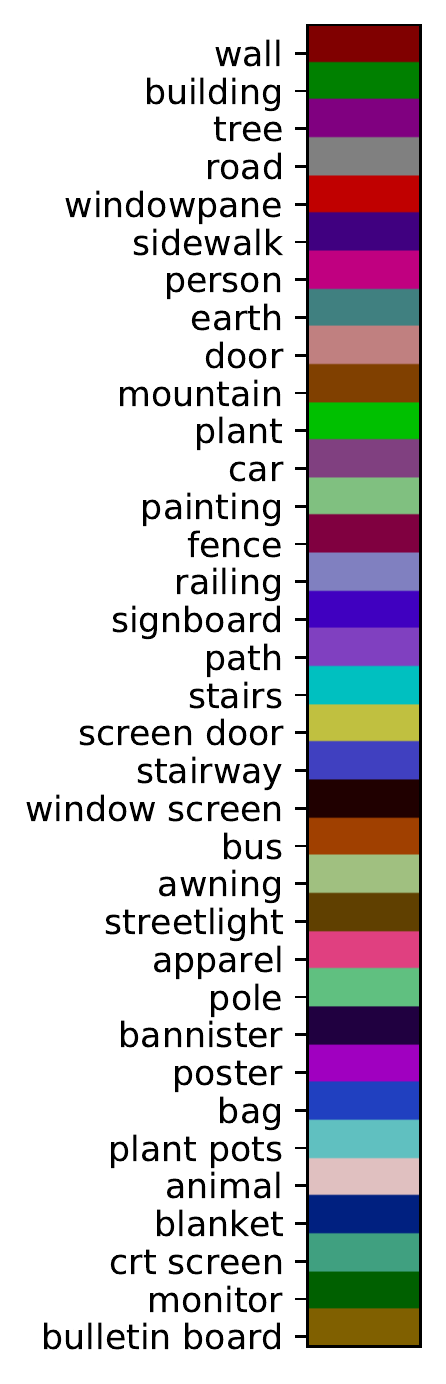}&%
\includegraphics[height=0.265\linewidth]{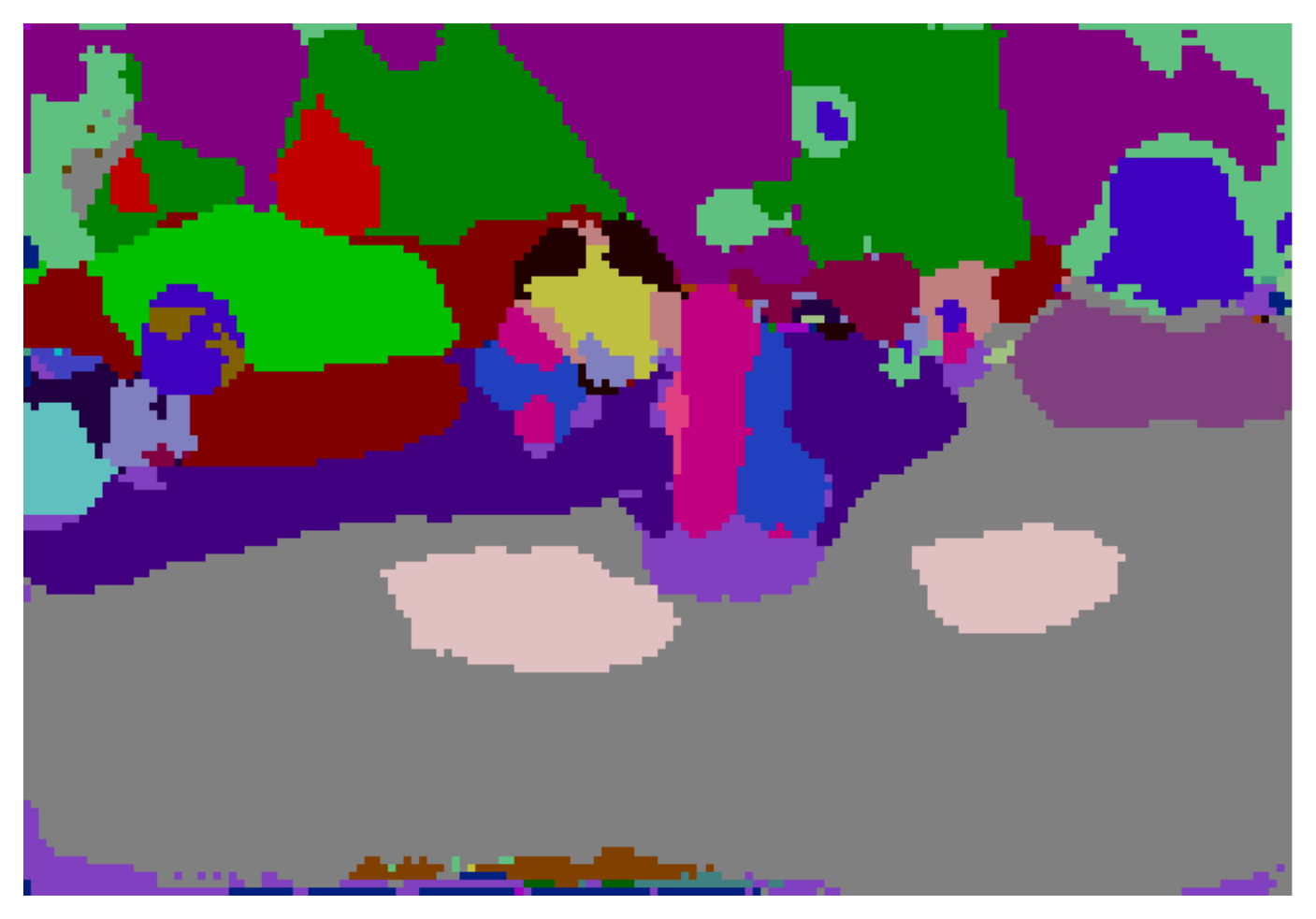}&
\includegraphics[height=0.265\linewidth]{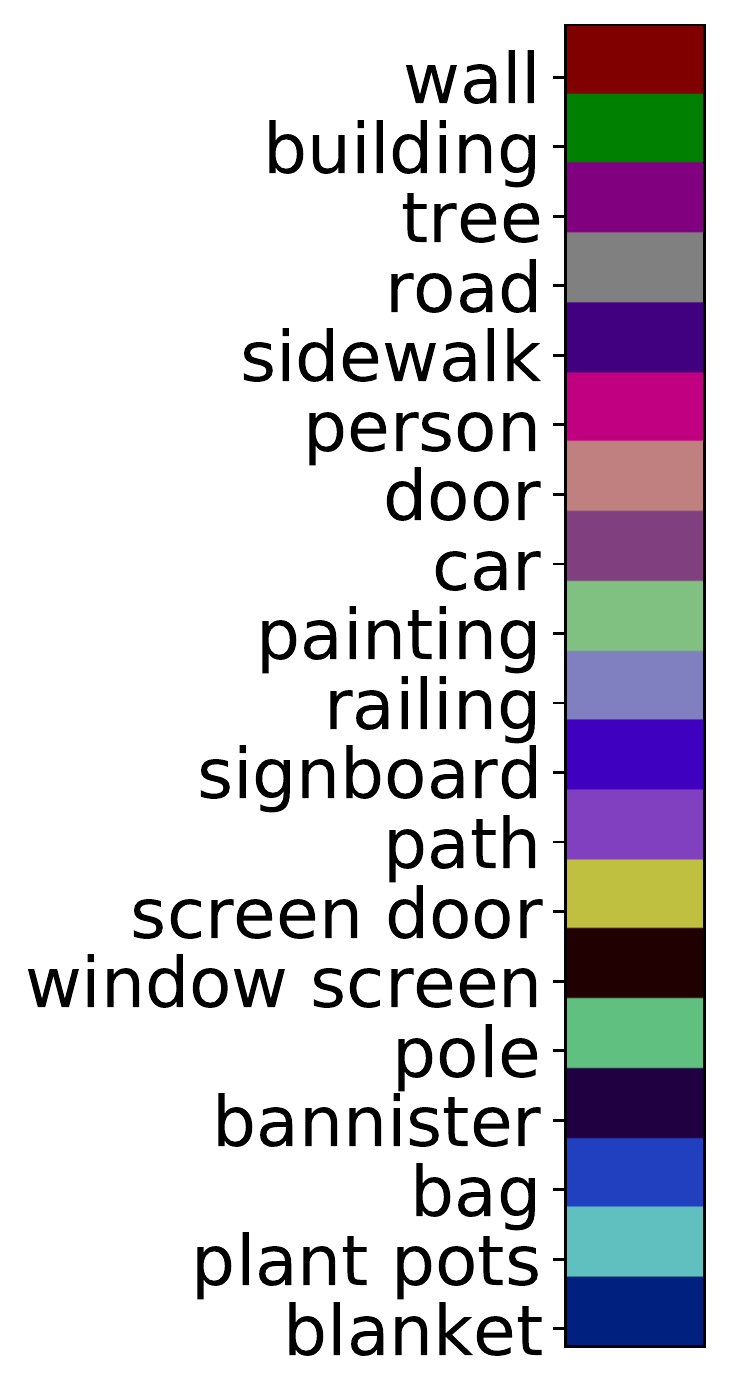}&%
\includegraphics[height=0.265\linewidth]{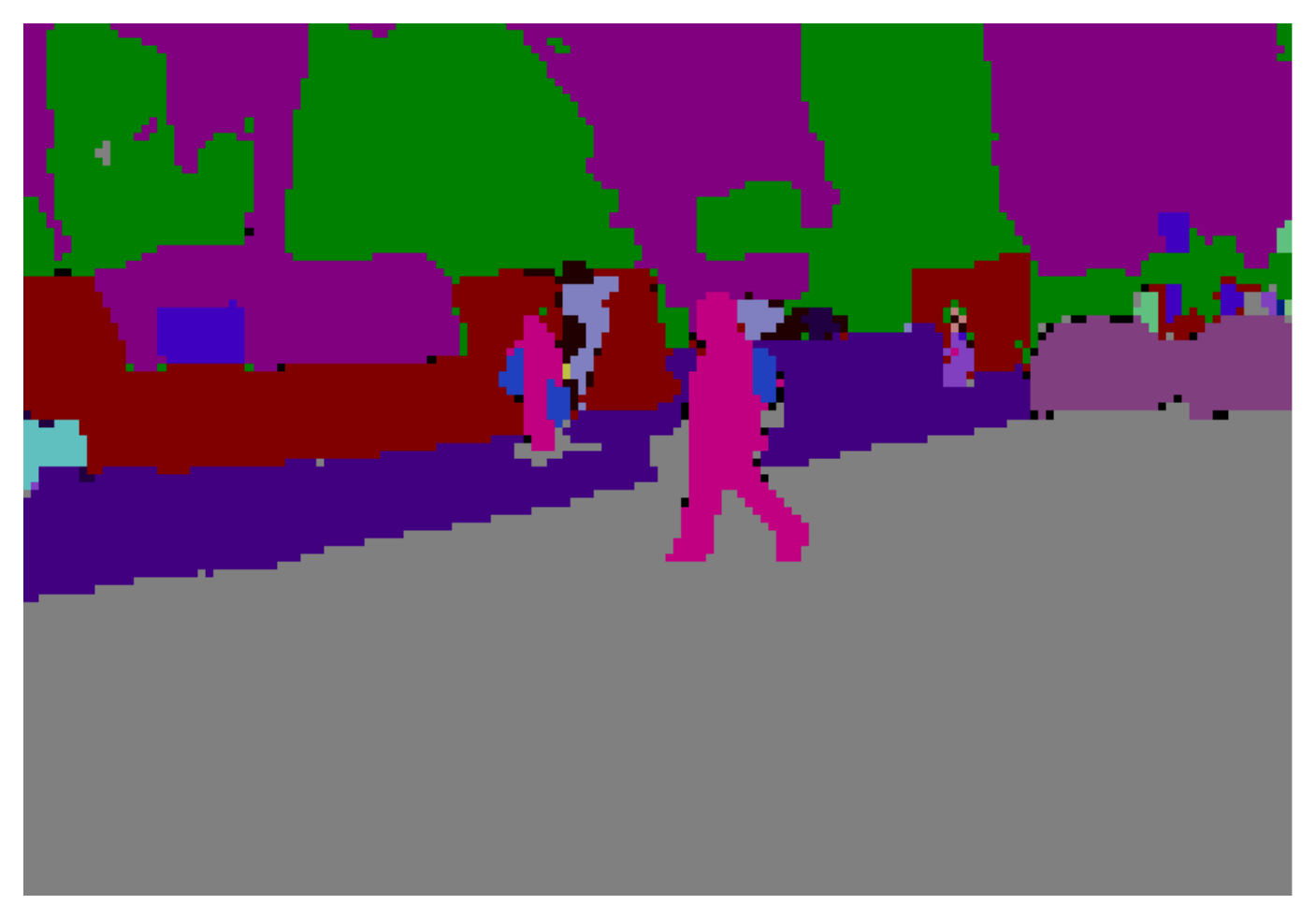}\\%
\end{tabular}
\caption{\textbf{Predictions of \ours~on random examples in the \ade~dataset (Part1).} For each example, top left is the input image, top right is the ground-truth mask, and bottom right is the output of \ours. Bottom left shows per-pixel prediction of \ours~without incorporating segmentation proposals. Note we only display one name in the legend for each category, but each category may include a list of names. }
\label{fig:ade20k_examples_1}
\end{figure*}

\begin{figure*}%
\centering%
\setlength\tabcolsep{0.5pt}
\renewcommand{\arraystretch}{0.2}
\begin{tabular}{rlrl}%
\centering%
&%
\includegraphics[height=0.28\linewidth]{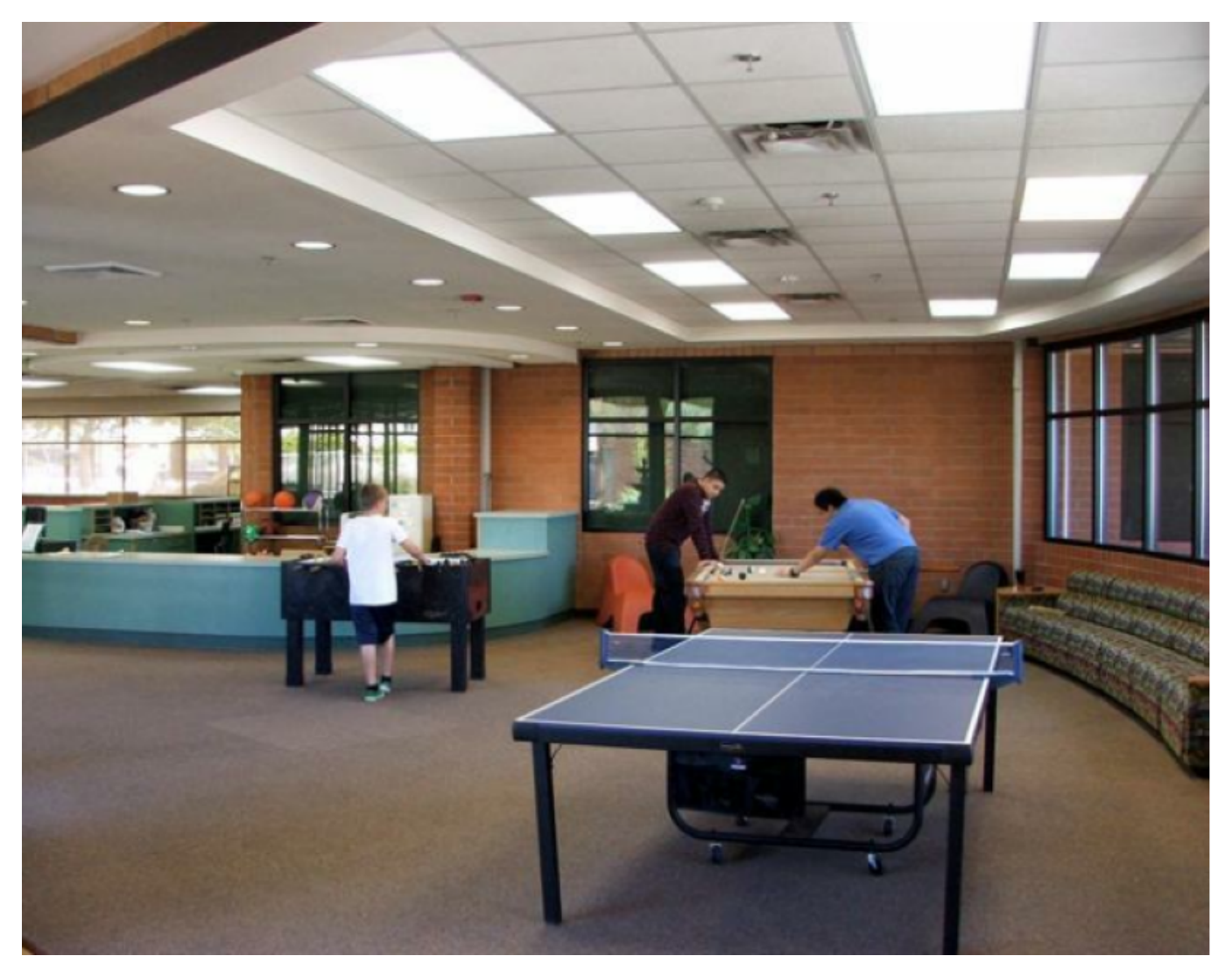}&%
\includegraphics[height=0.25\linewidth]{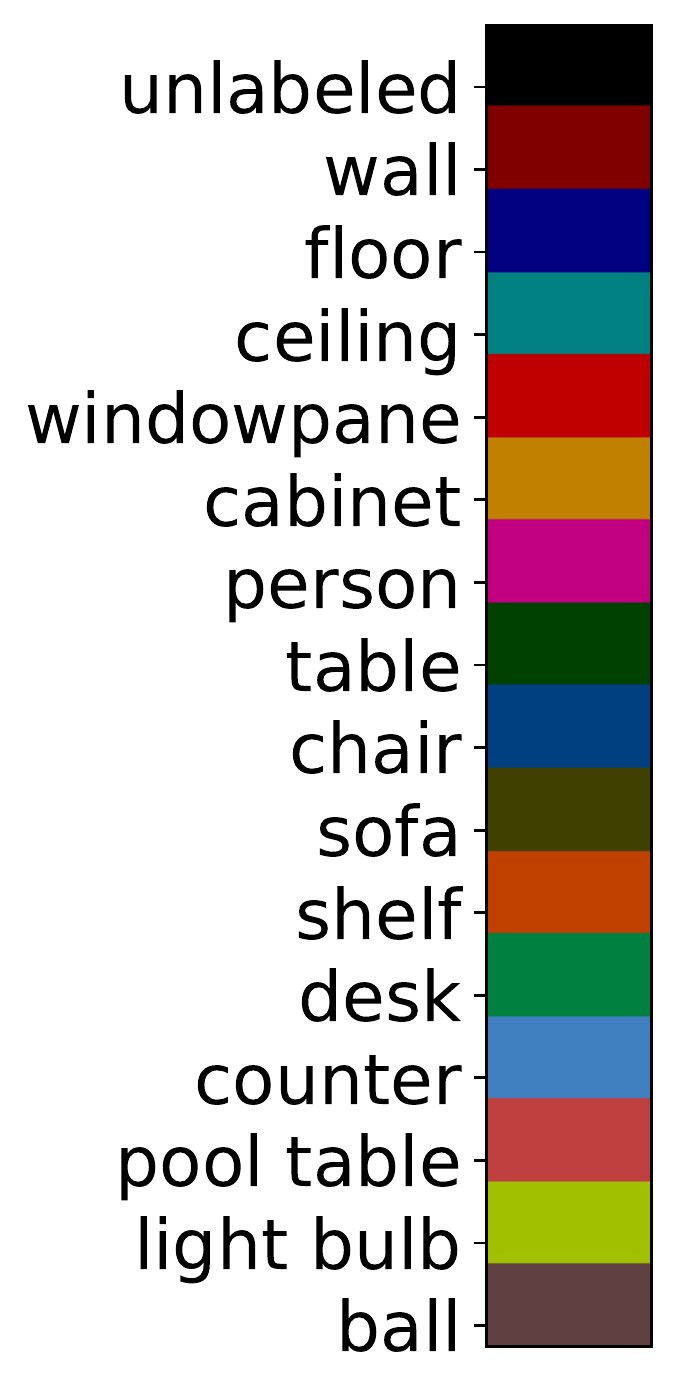}&%
\includegraphics[height=0.28\linewidth]{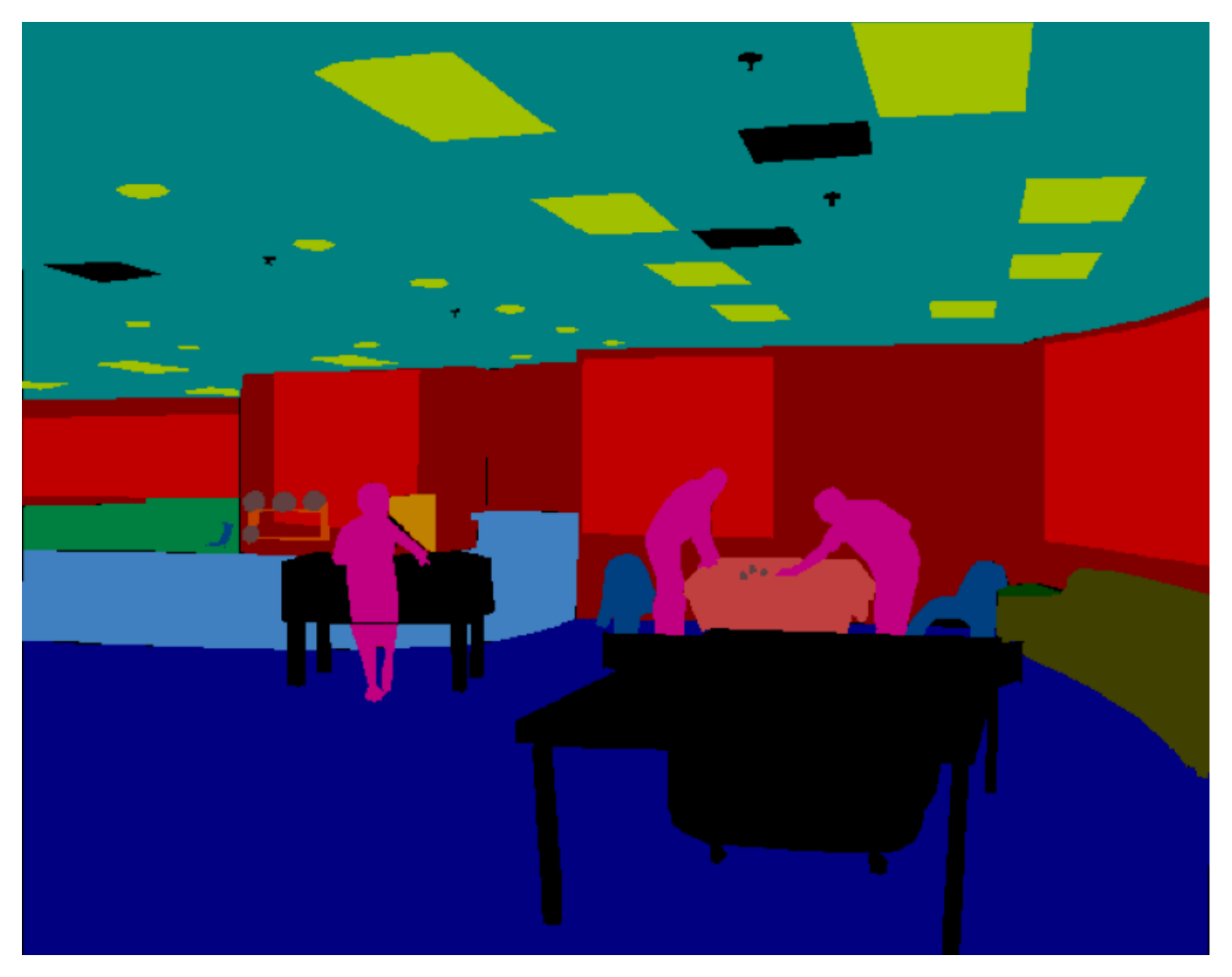}\\%
\includegraphics[height=0.28\linewidth]{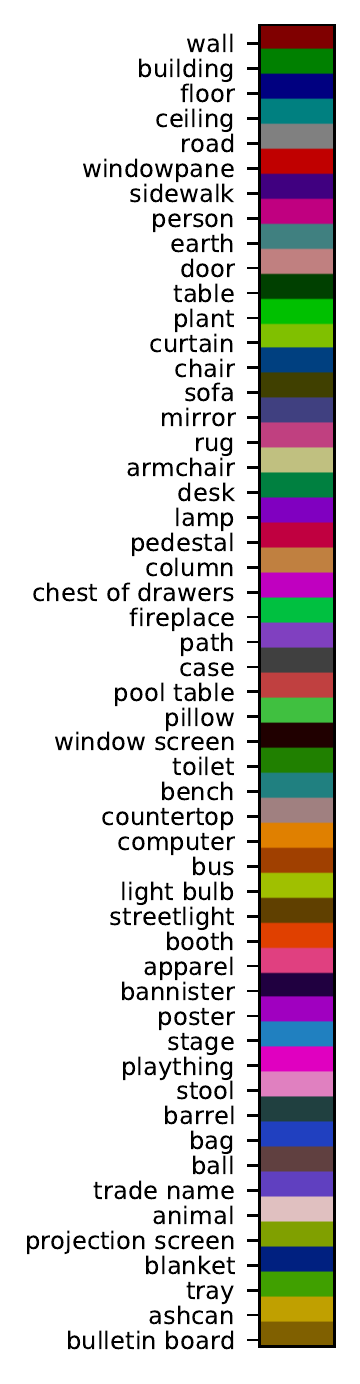}&%
\includegraphics[height=0.28\linewidth]{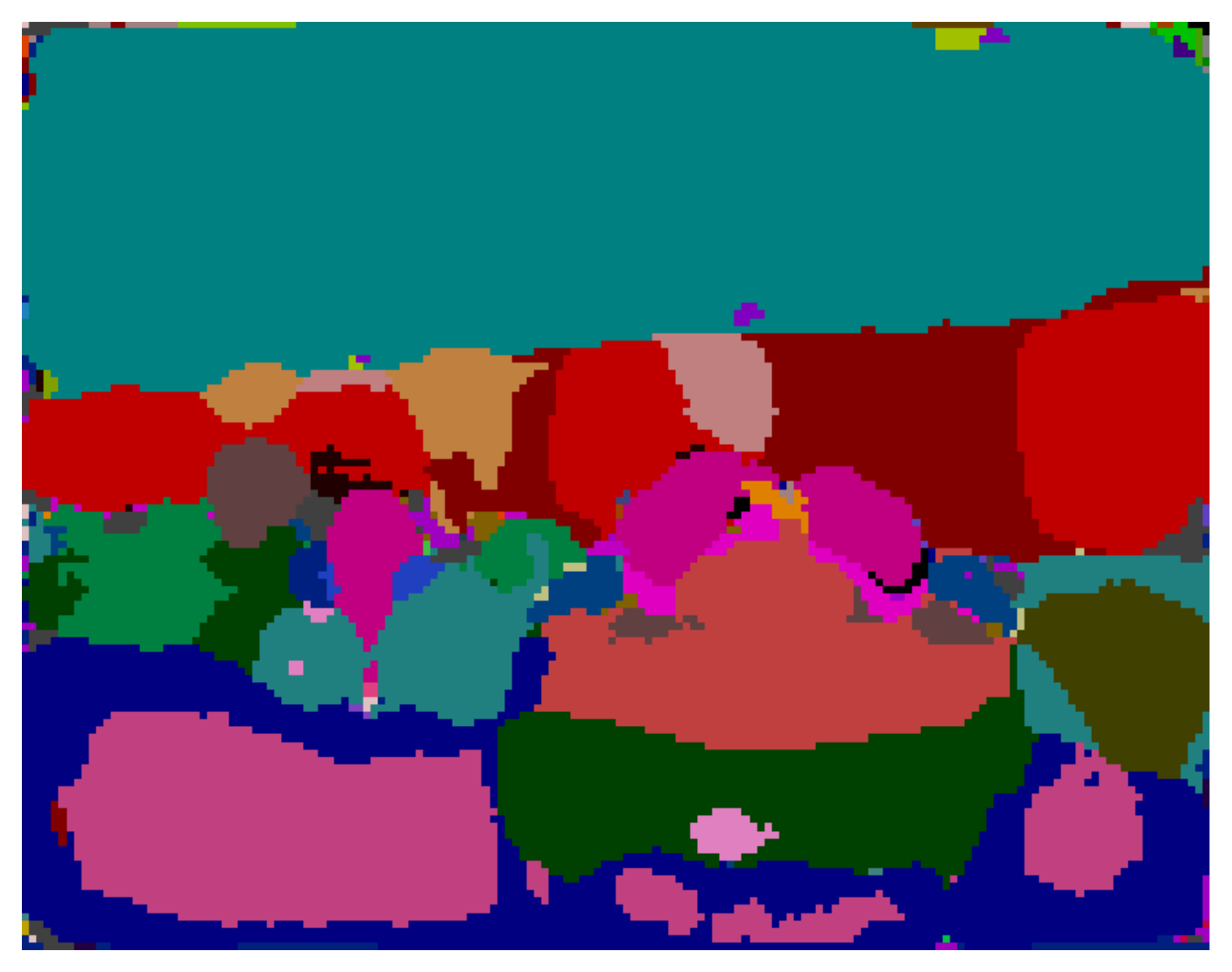}&
\includegraphics[height=0.27\linewidth]{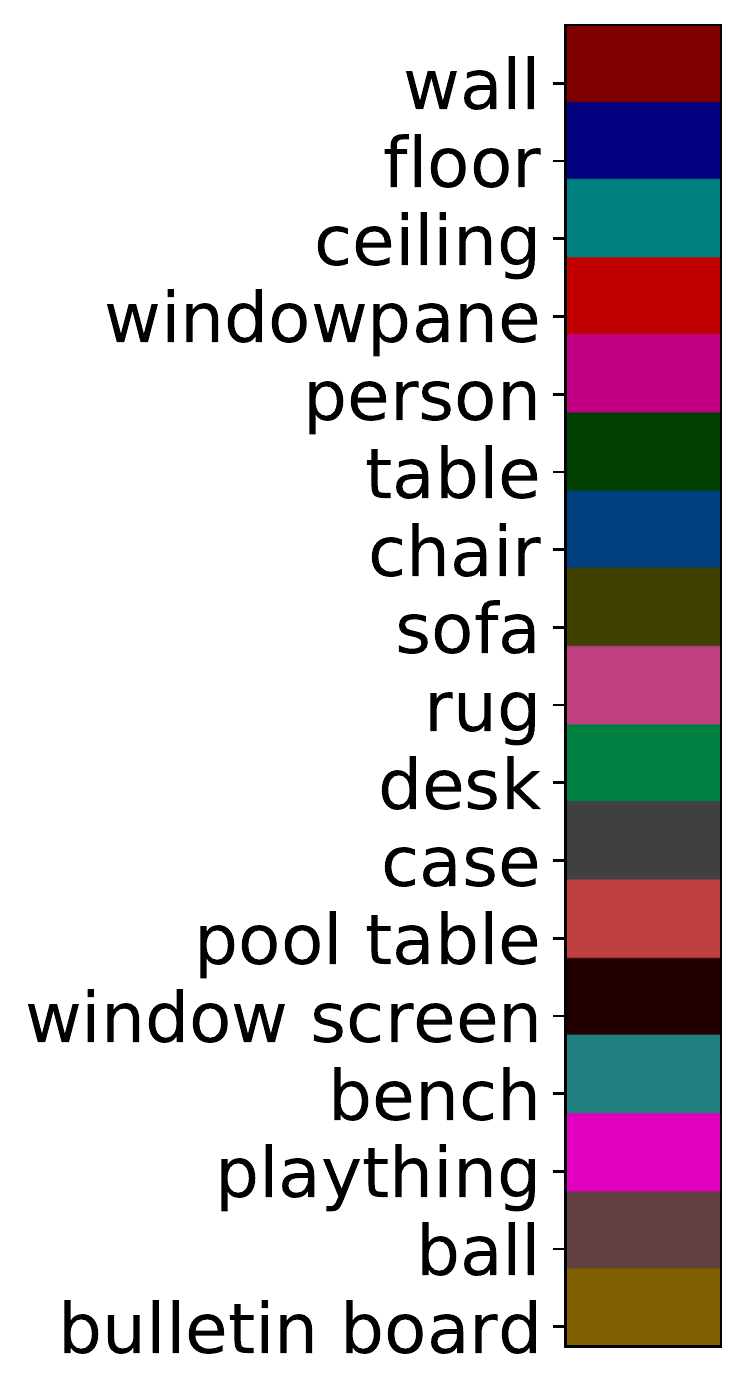}&%
\includegraphics[height=0.28\linewidth]{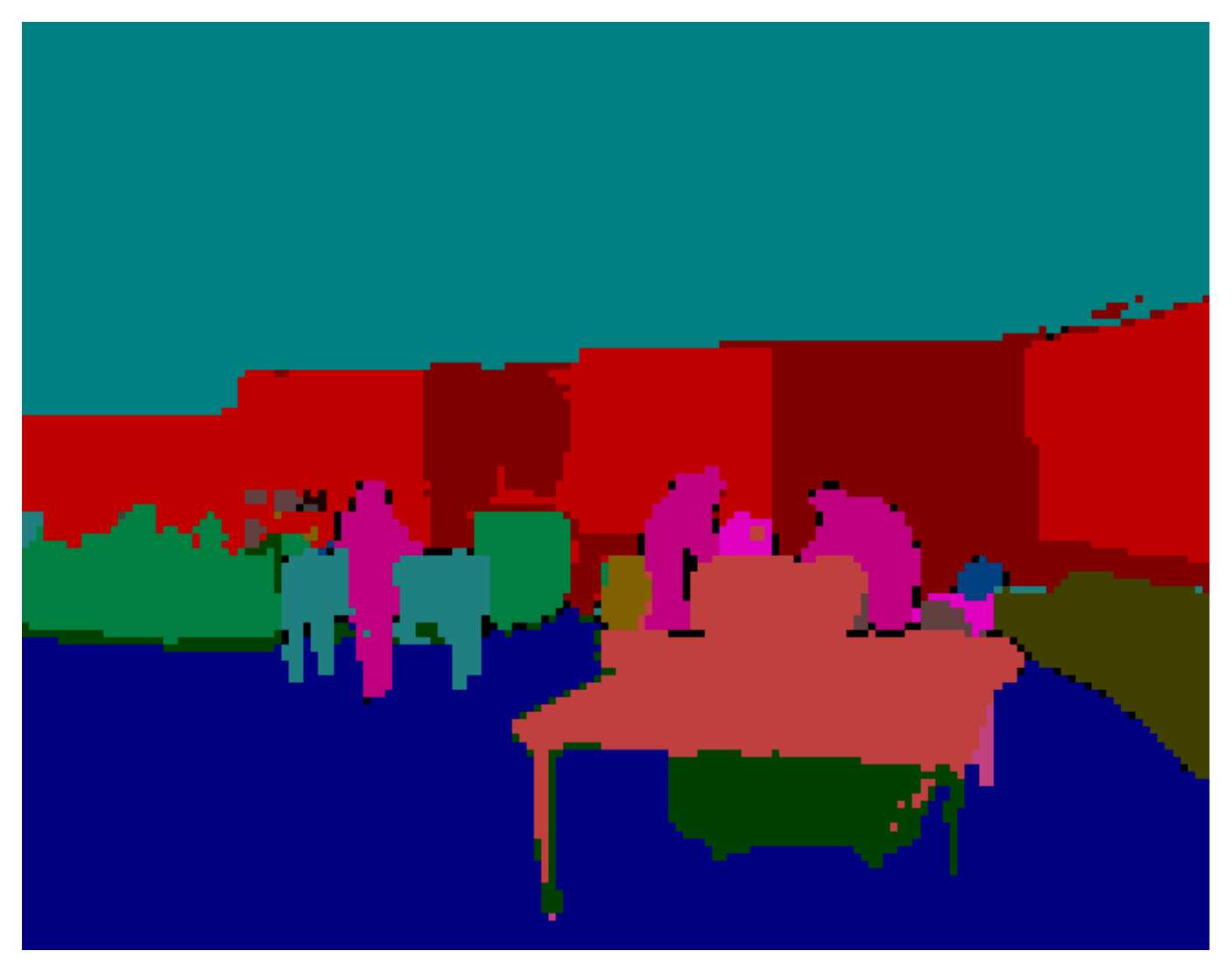}\\%
\vspace{1cm}\\
&%
\includegraphics[height=0.27\linewidth]{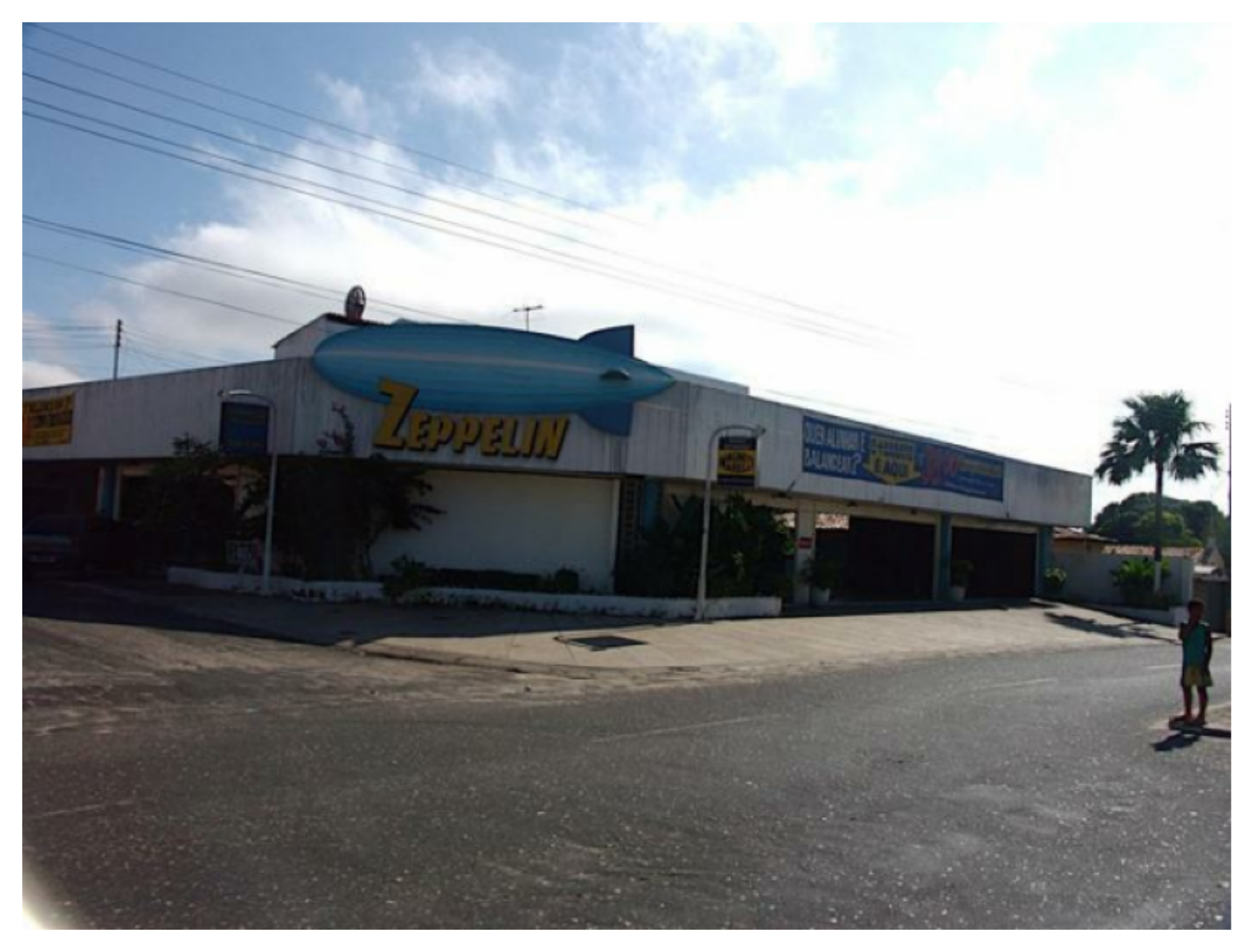}&%
\includegraphics[height=0.22\linewidth]{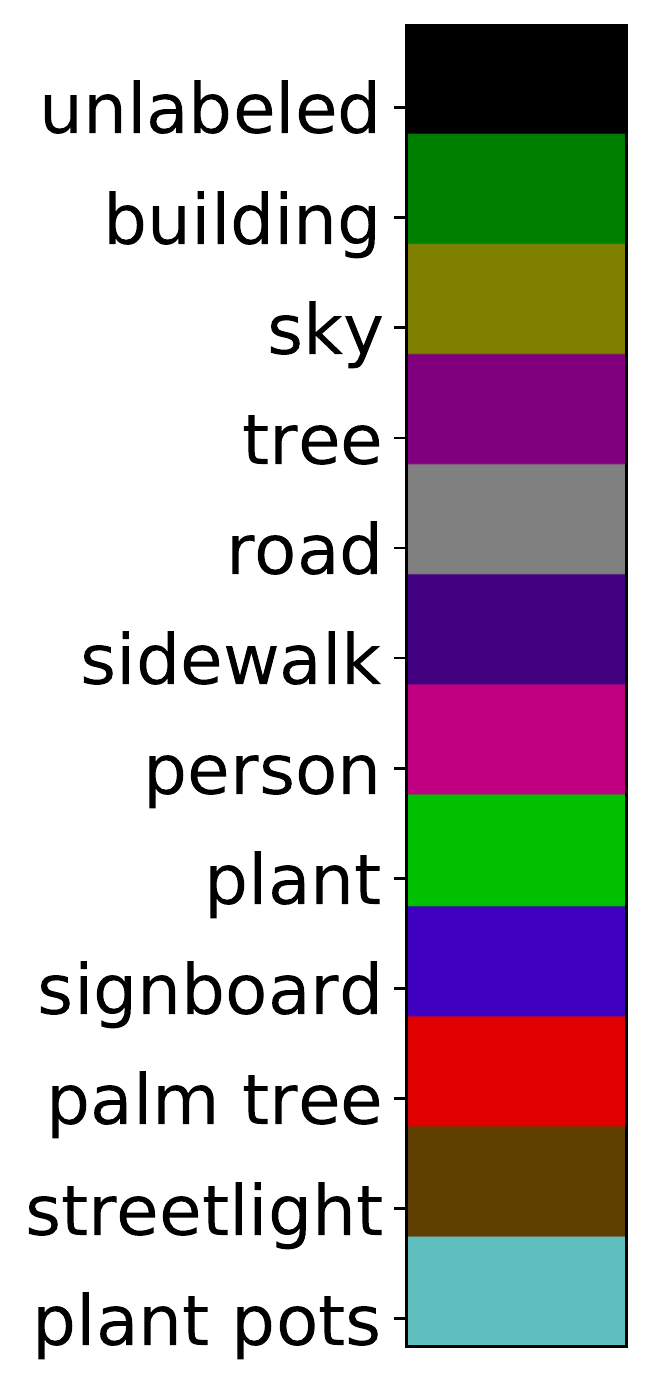}&%
\includegraphics[height=0.27\linewidth]{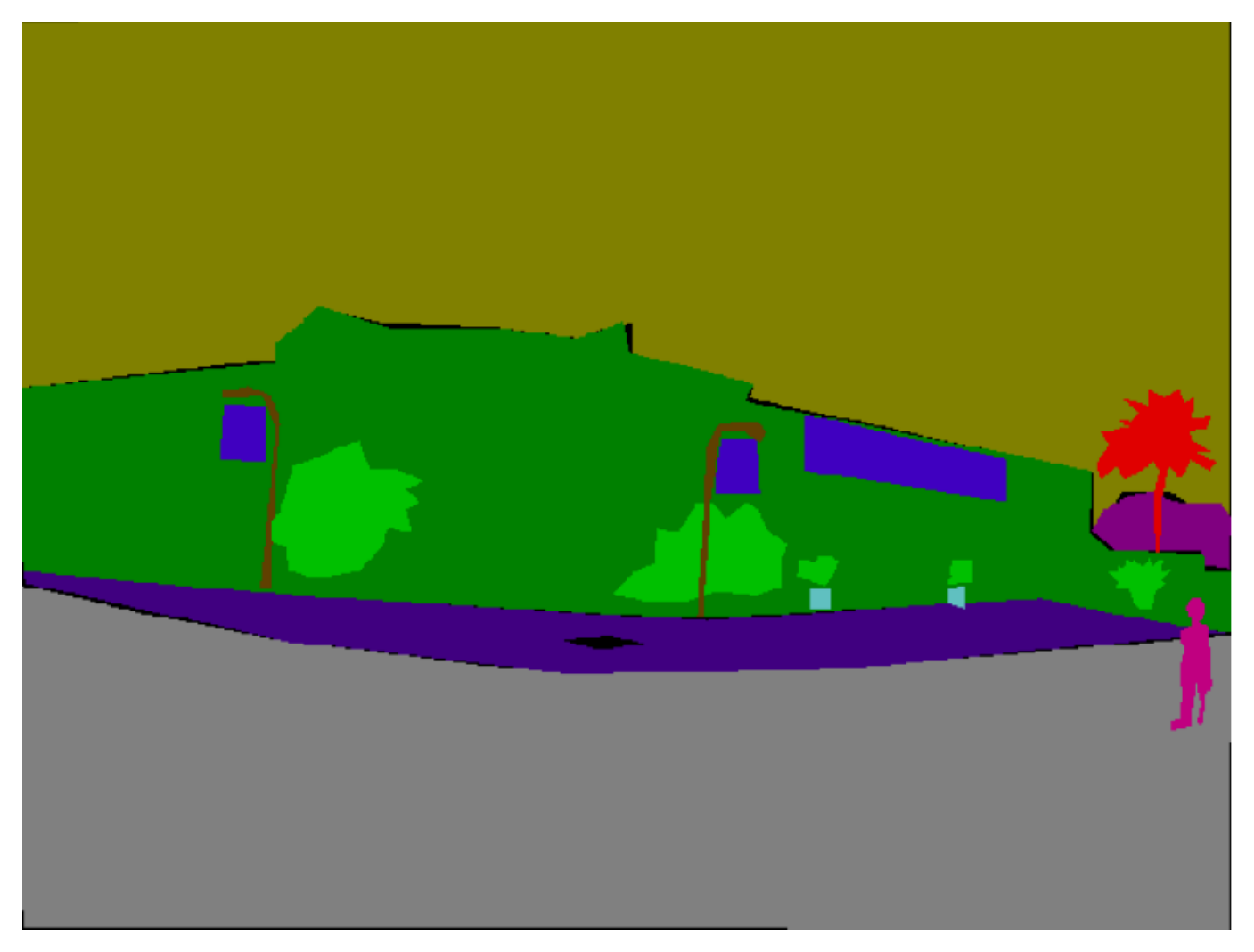}\\%
\includegraphics[height=0.27\linewidth]{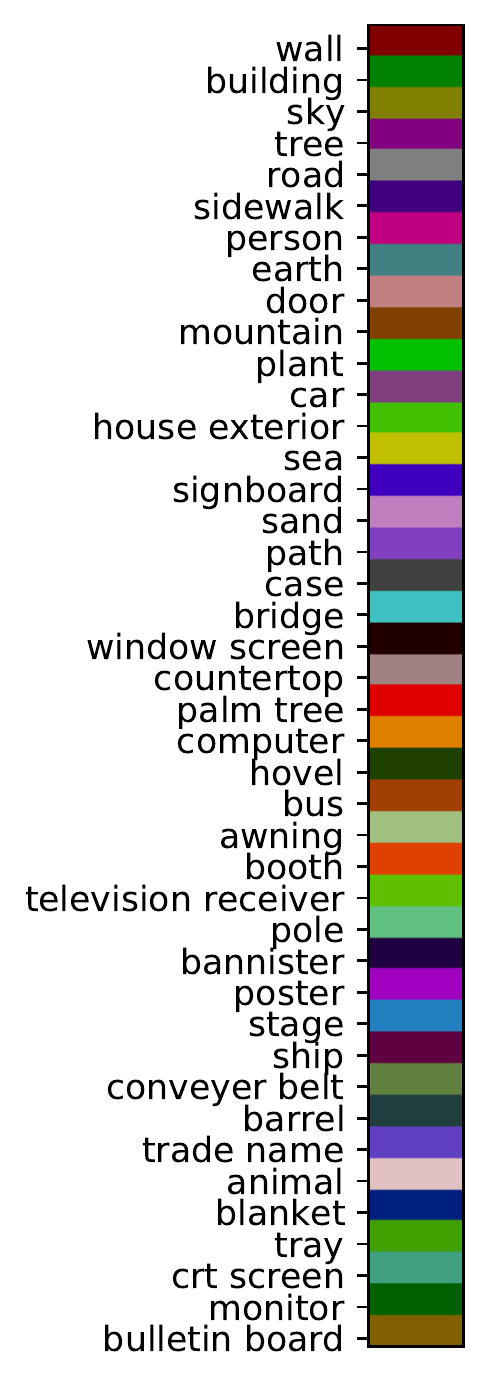}&%
\includegraphics[height=0.27\linewidth]{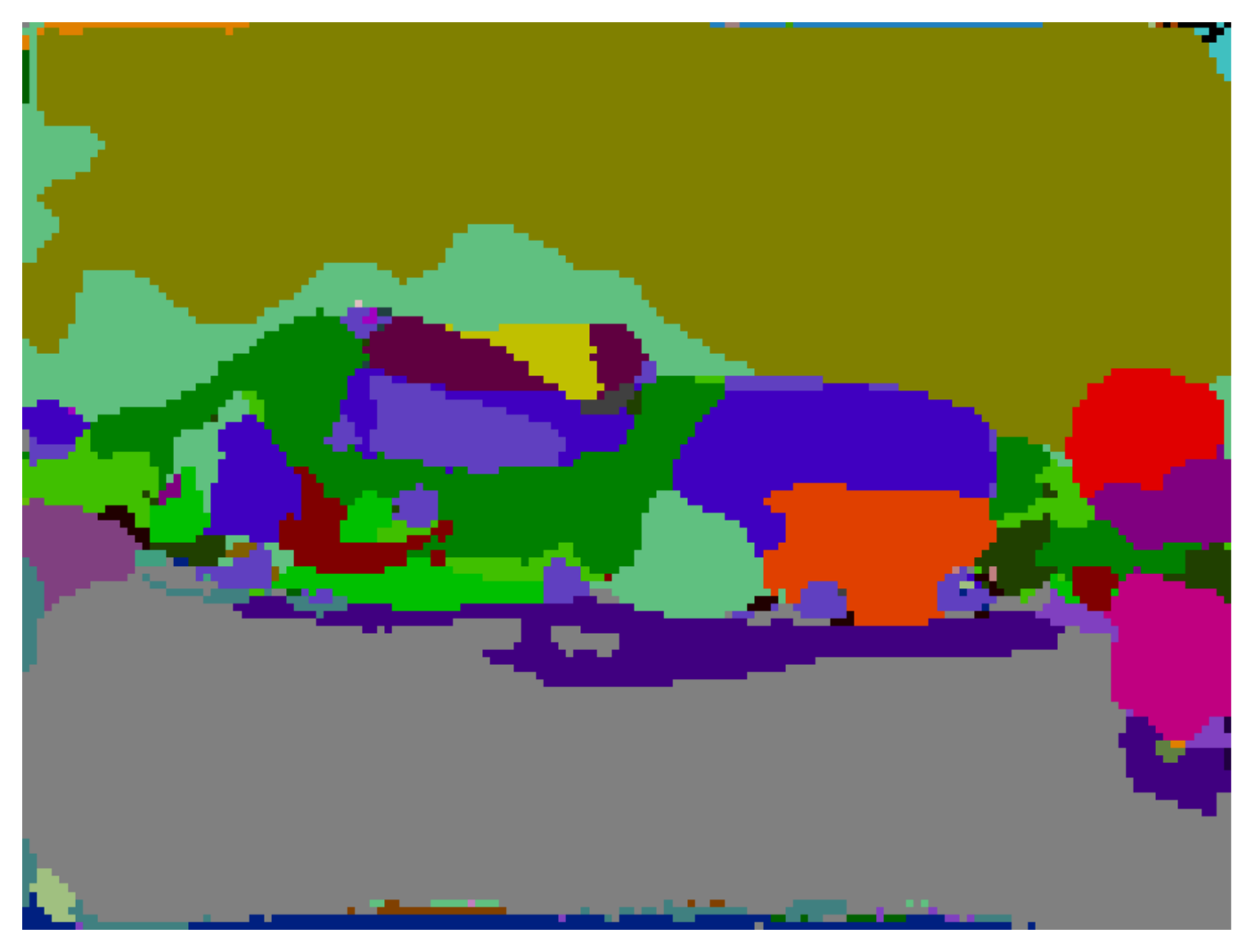}&
\includegraphics[height=0.27\linewidth]{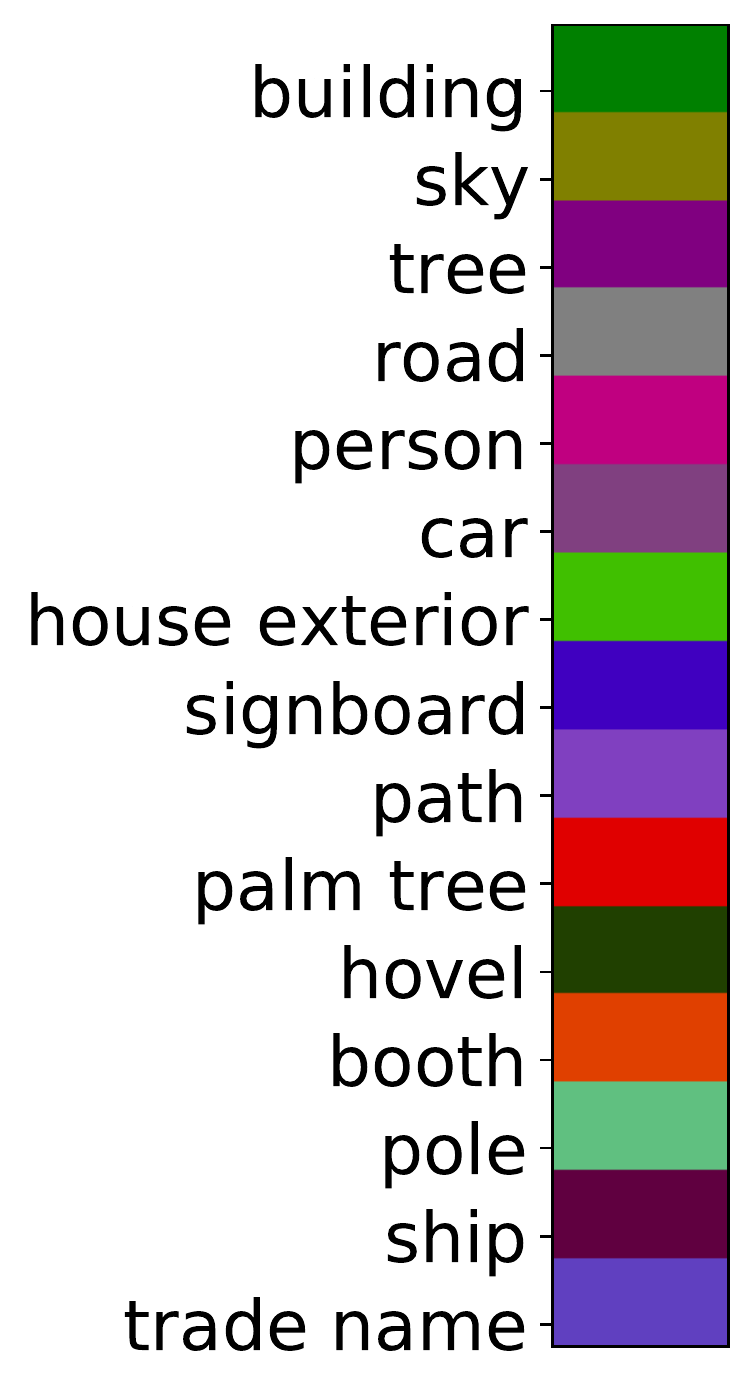}&%
\includegraphics[height=0.27\linewidth]{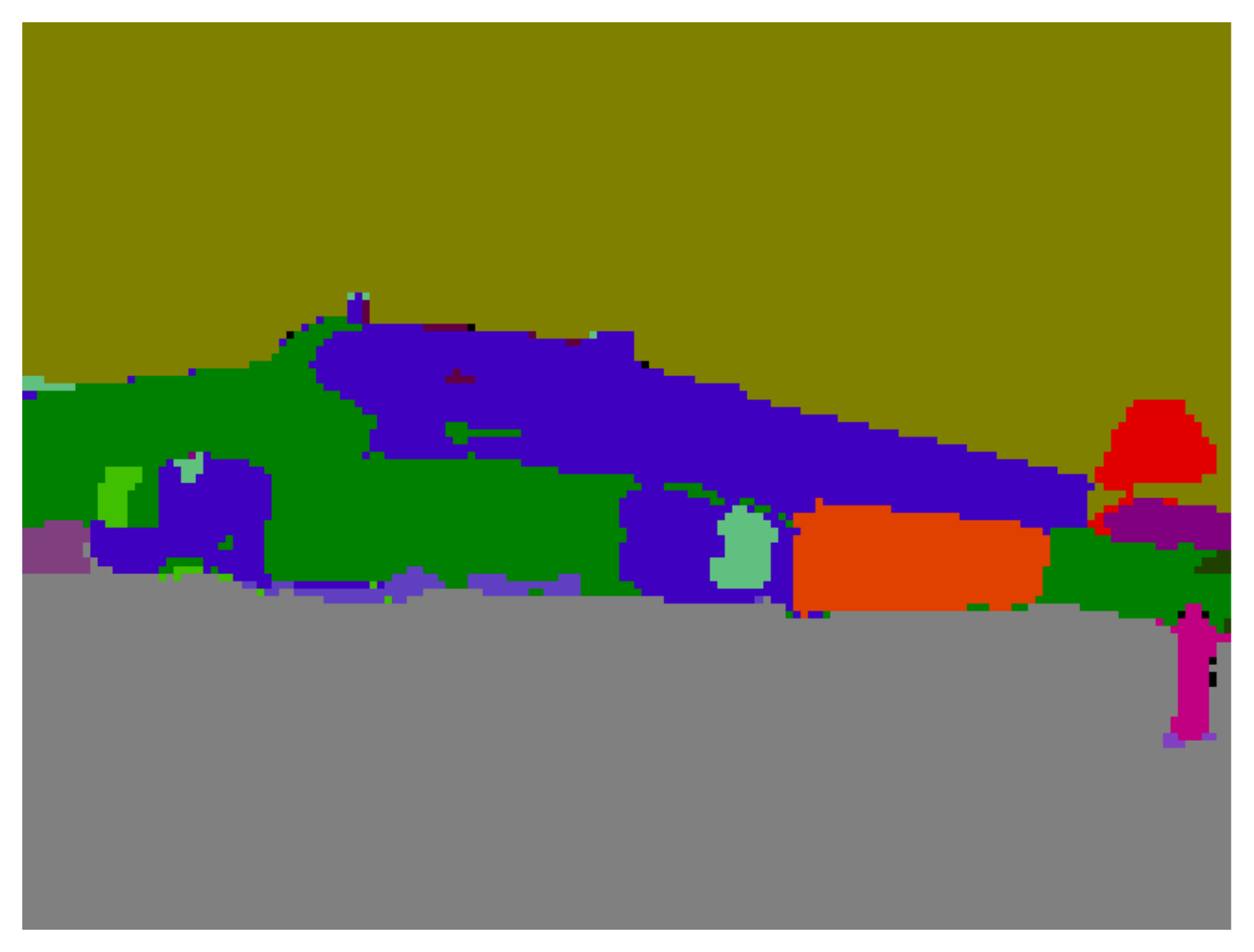}\\%
\end{tabular}
\caption{\textbf{Predictions of \ours~on random examples in the \ade~dataset (Part2).}
}
\label{fig:ade20k_examples_2}
\end{figure*}

\begin{figure*}%
\centering%
\setlength\tabcolsep{0.5pt}
\renewcommand{\arraystretch}{0.2}
\begin{tabular}{rlrl}%
\centering%
&%
\includegraphics[height=0.27\linewidth]{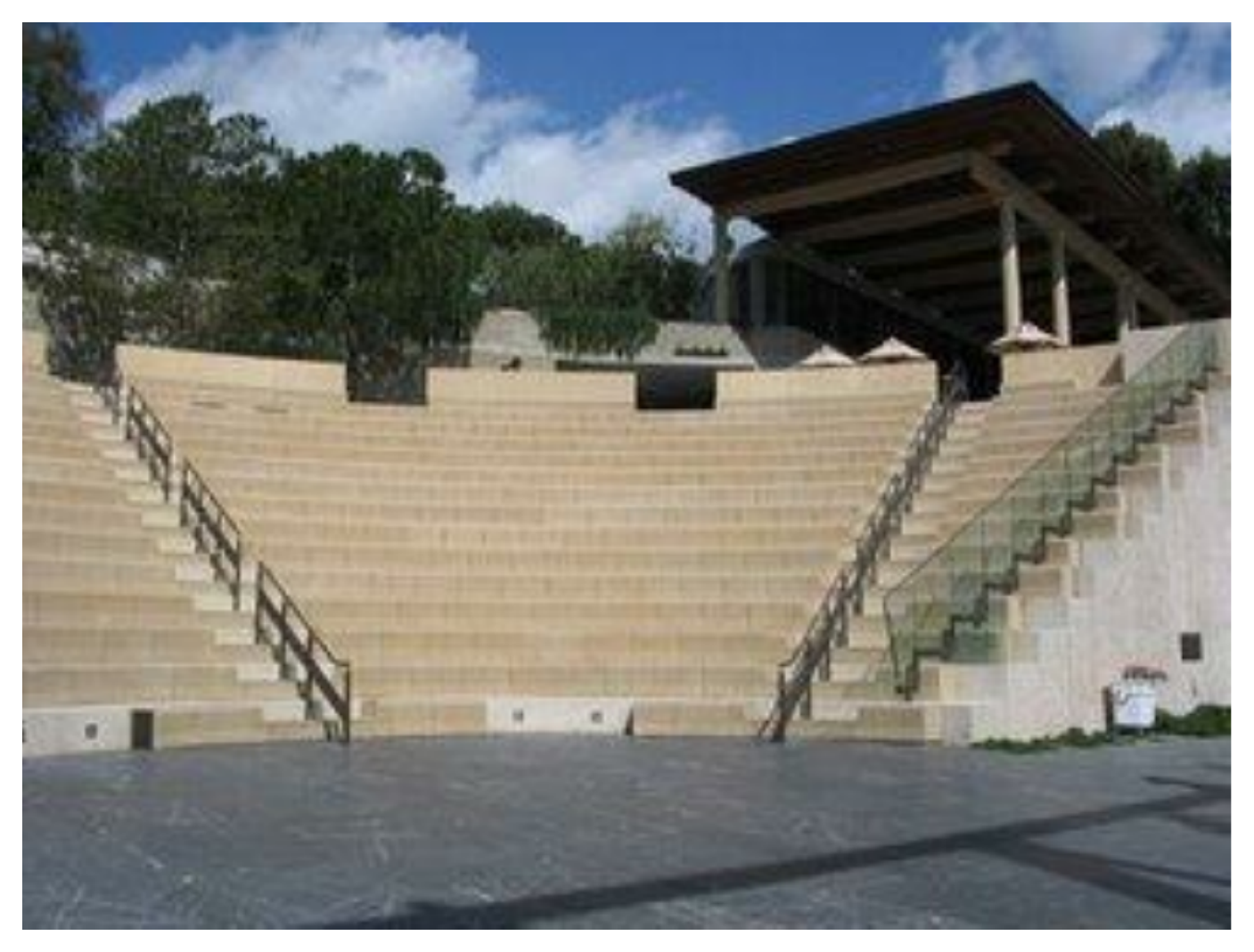}&%
\includegraphics[height=0.22\linewidth]{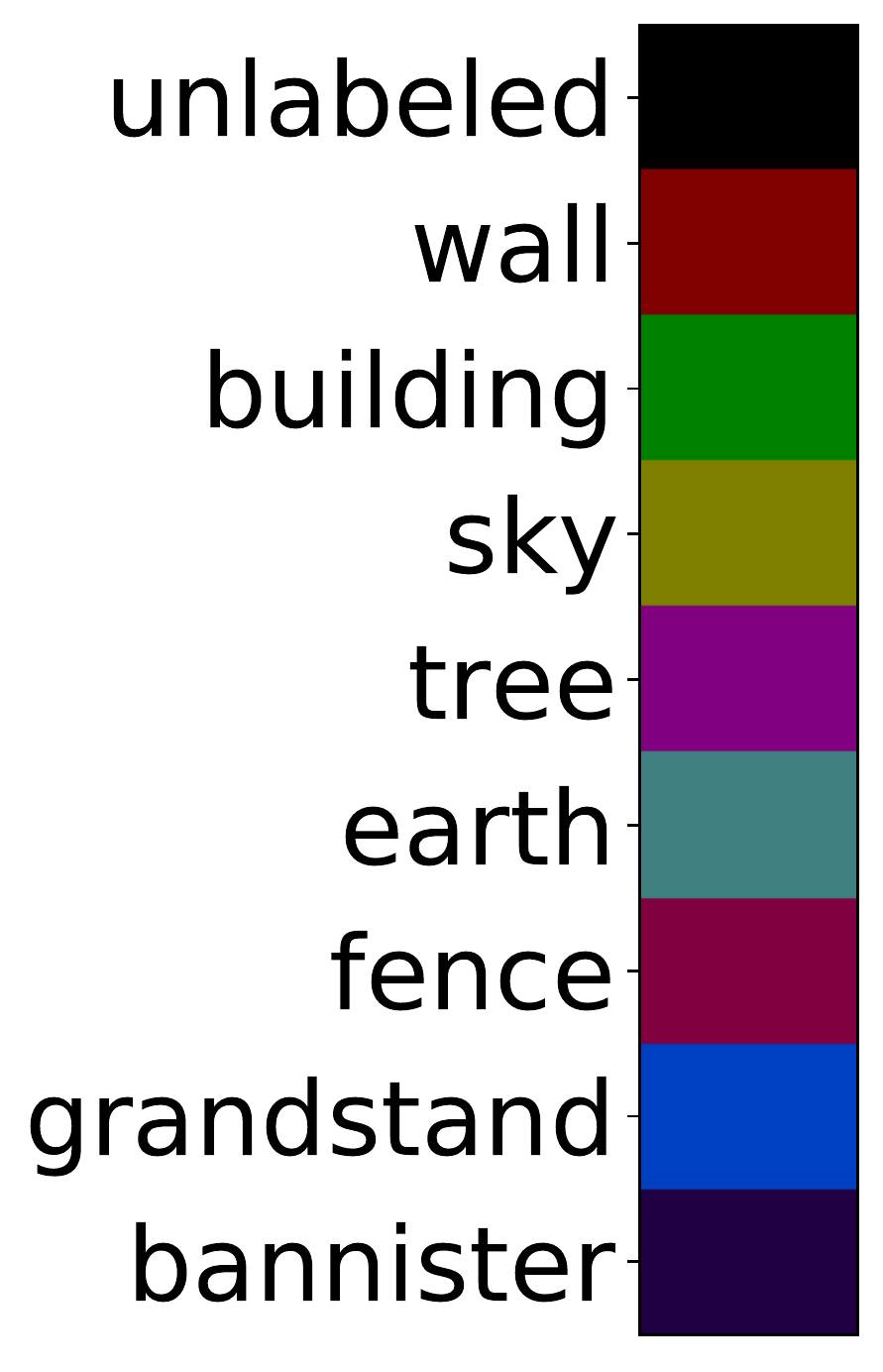}&%
\includegraphics[height=0.27\linewidth]{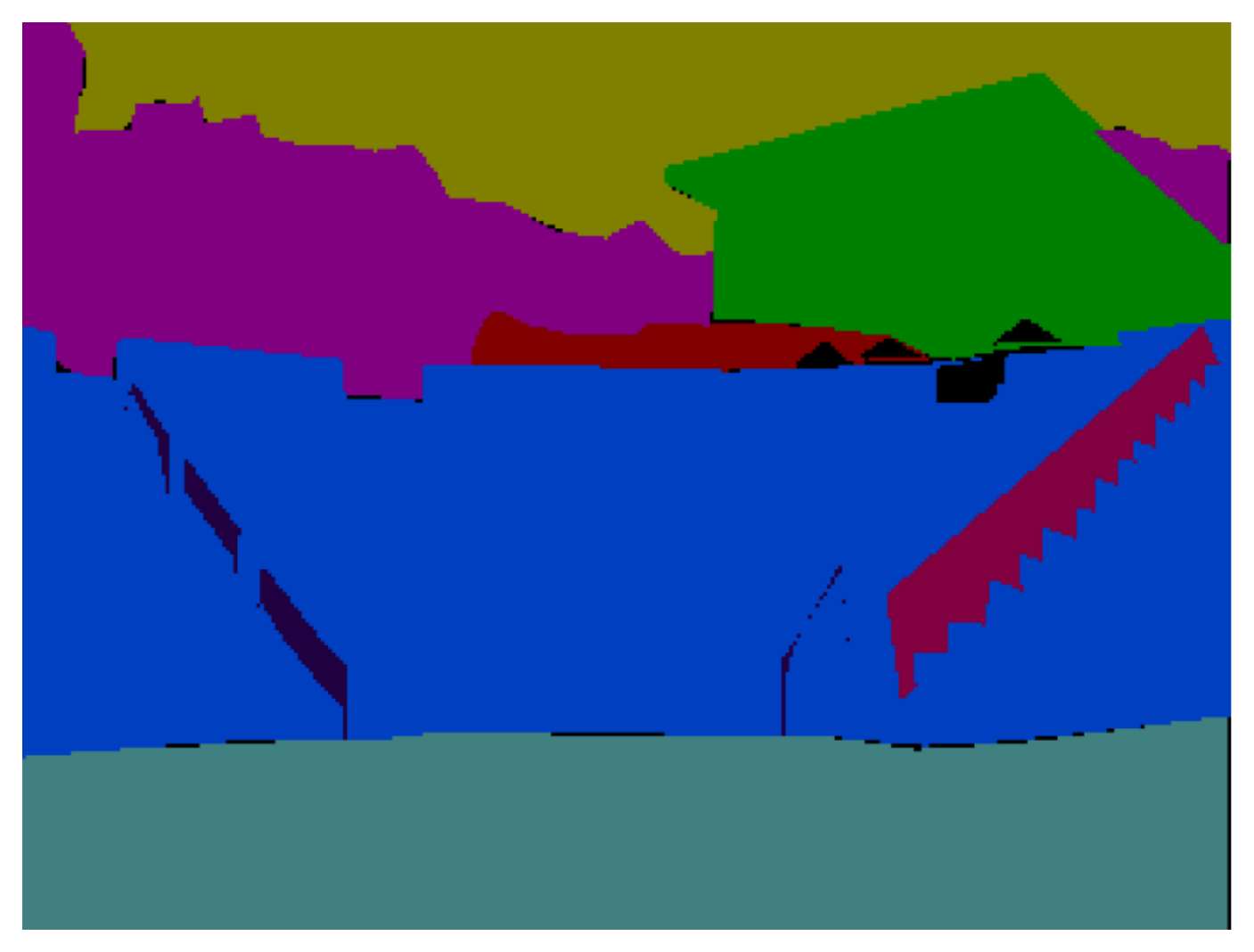}\\%
\includegraphics[height=0.27\linewidth]{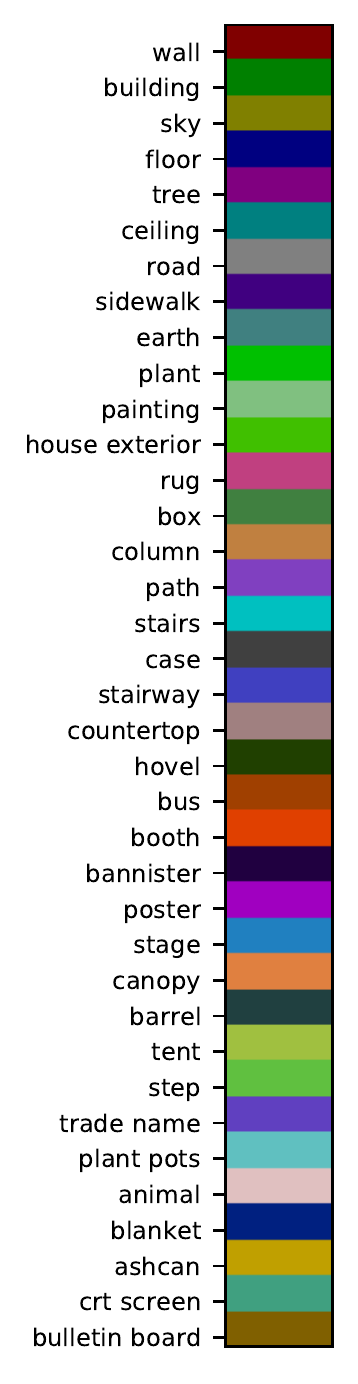}&%
\includegraphics[height=0.27\linewidth]{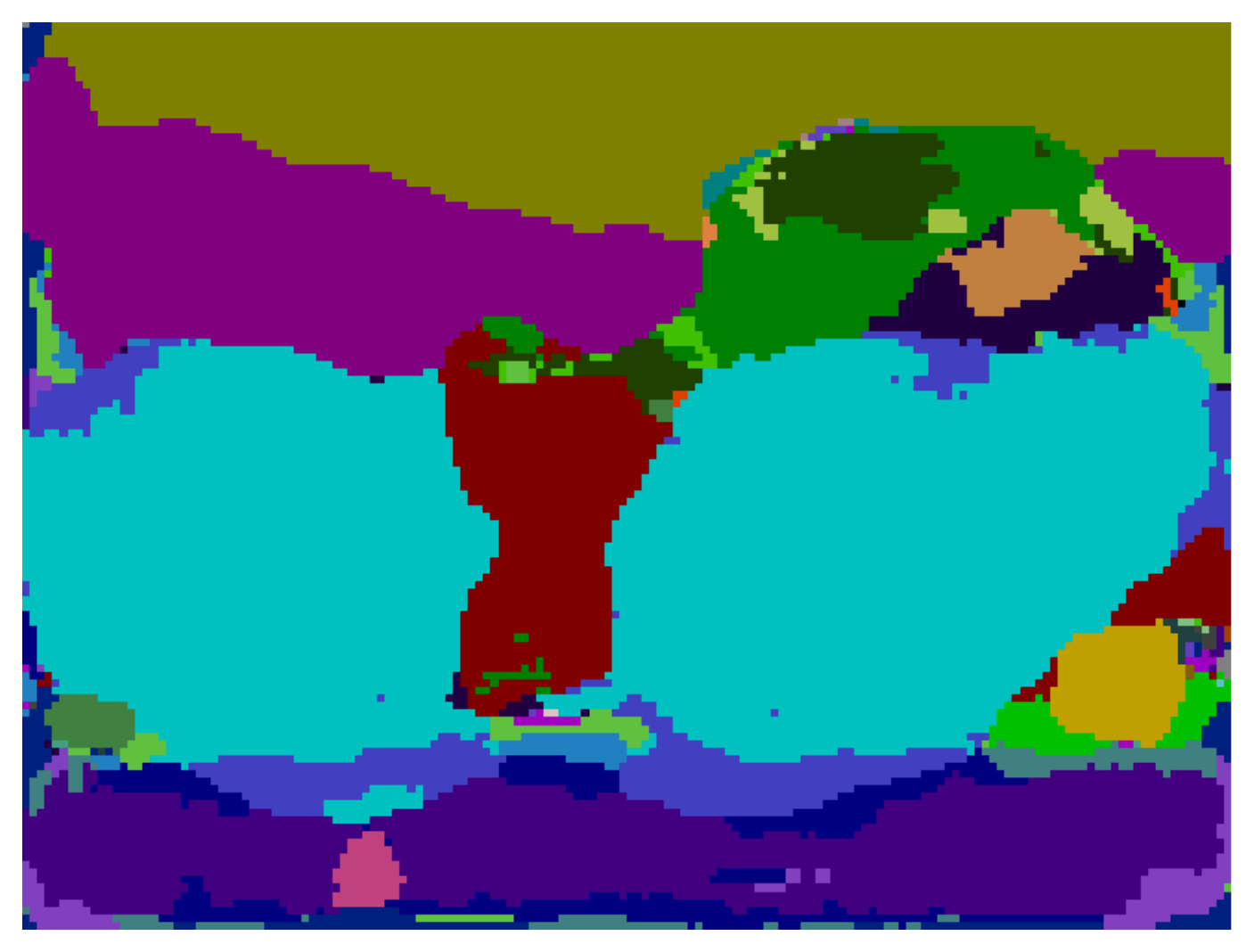}&
\includegraphics[height=0.26\linewidth]{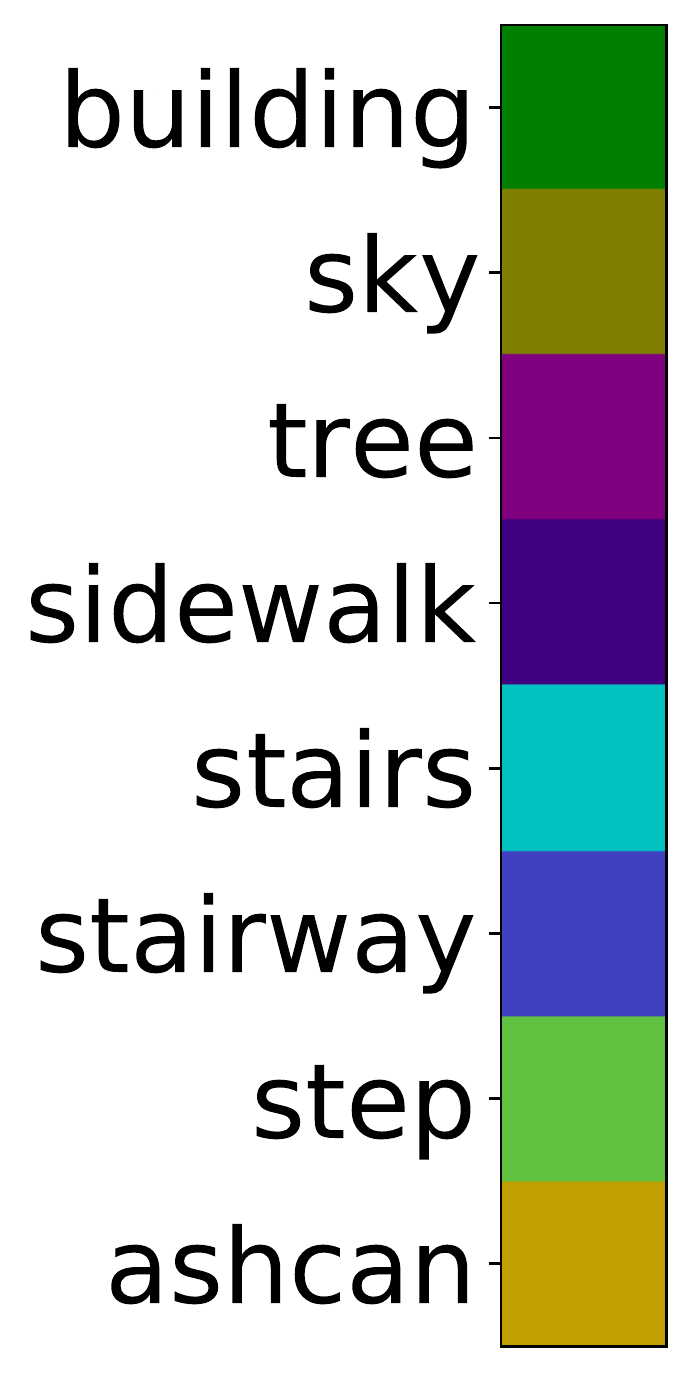}&%
\includegraphics[height=0.27\linewidth]{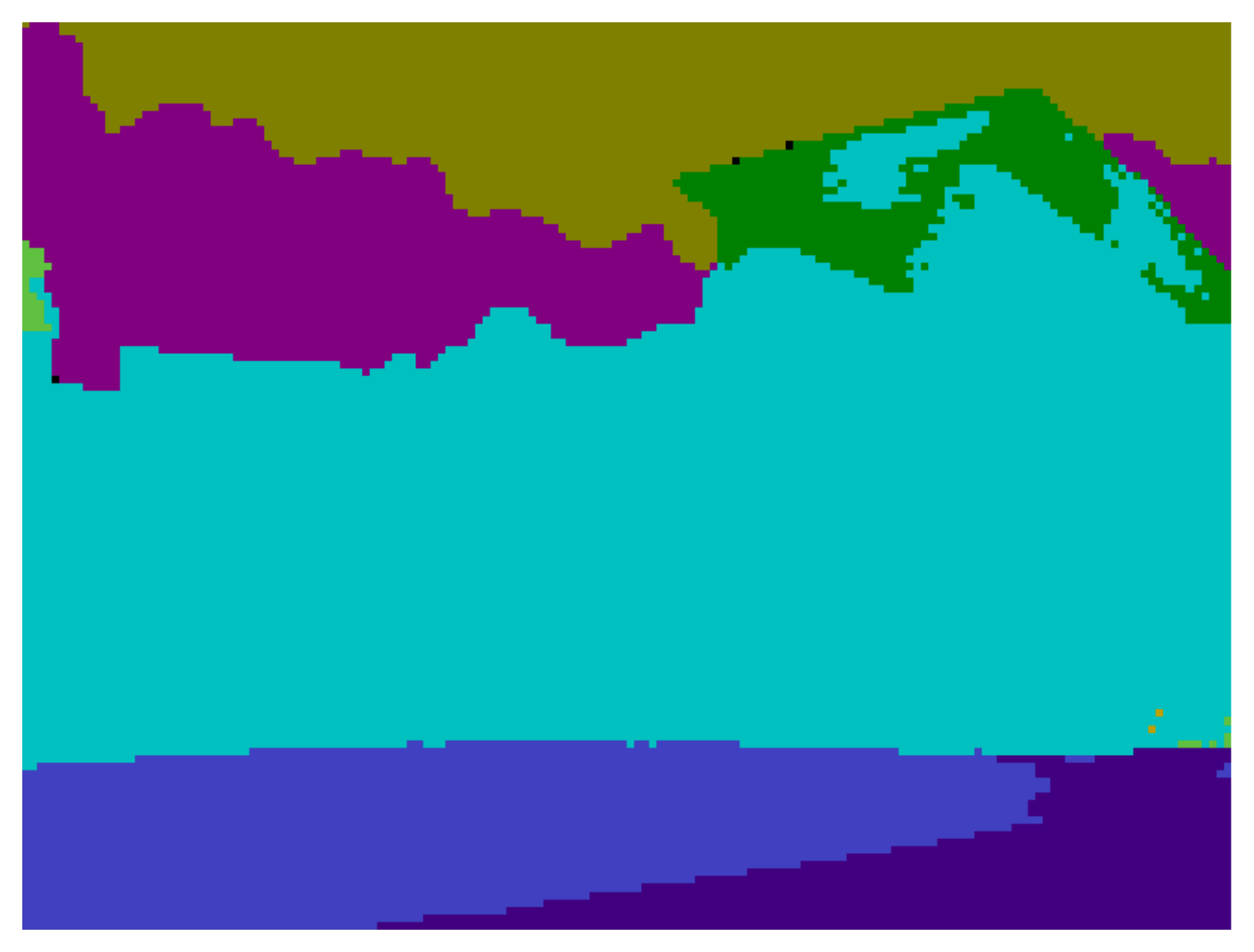}\\%
\vspace{1cm}\\
&%
\includegraphics[height=0.27\linewidth]{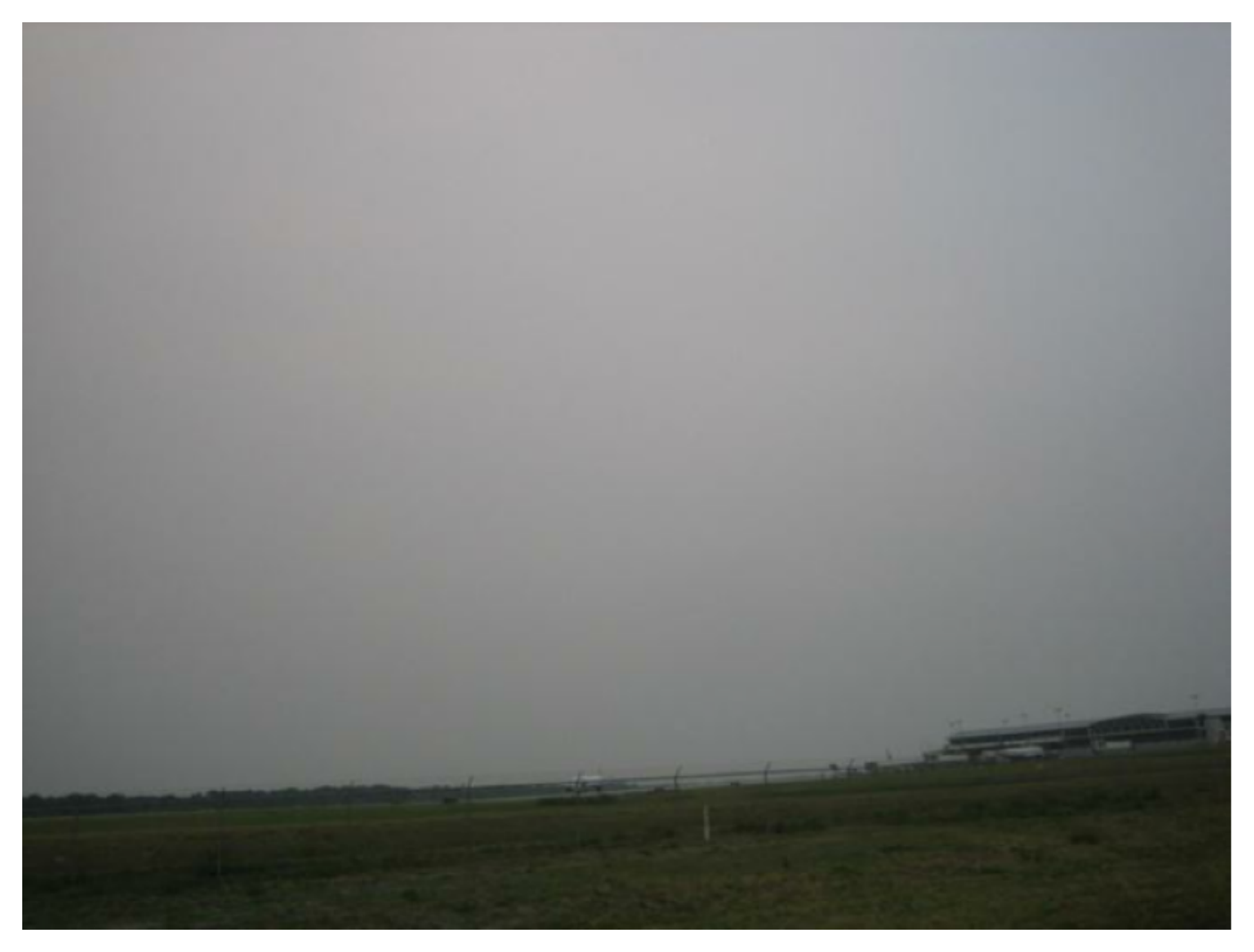}&%
\includegraphics[height=0.23\linewidth]{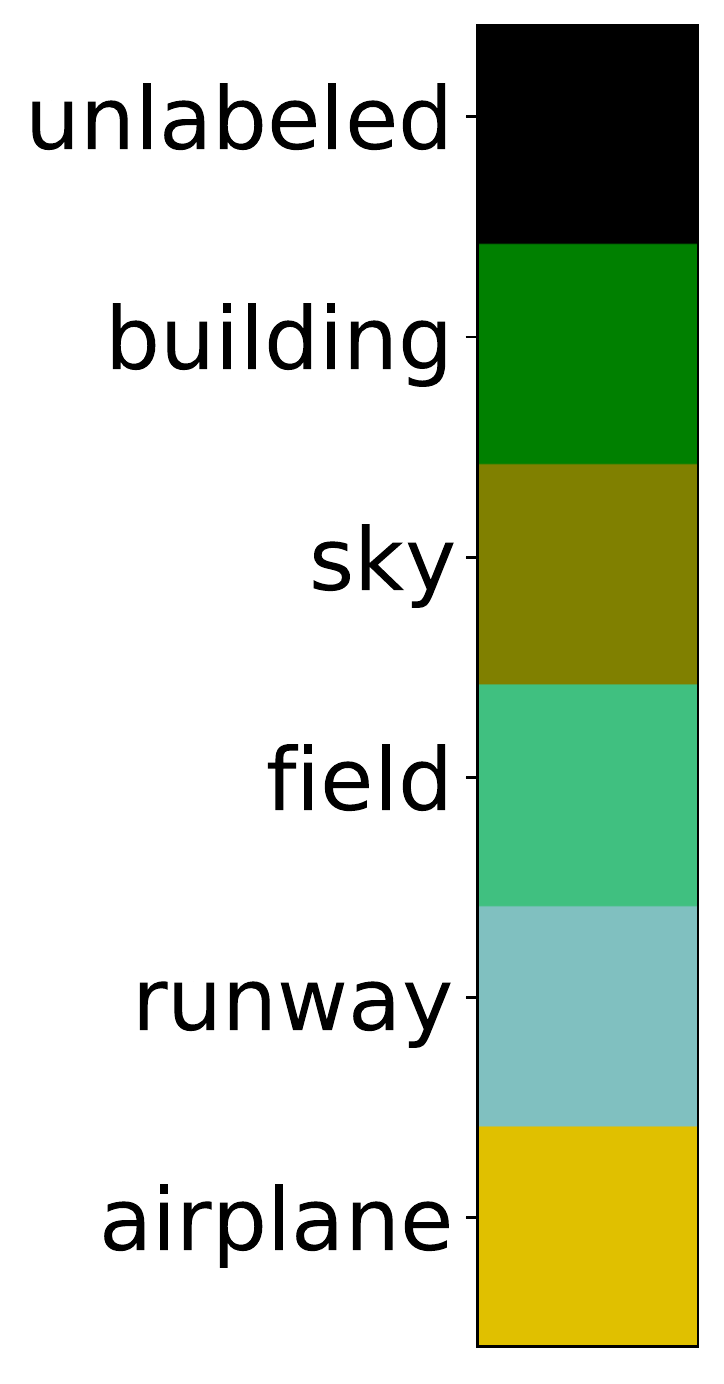}&%
\includegraphics[height=0.27\linewidth]{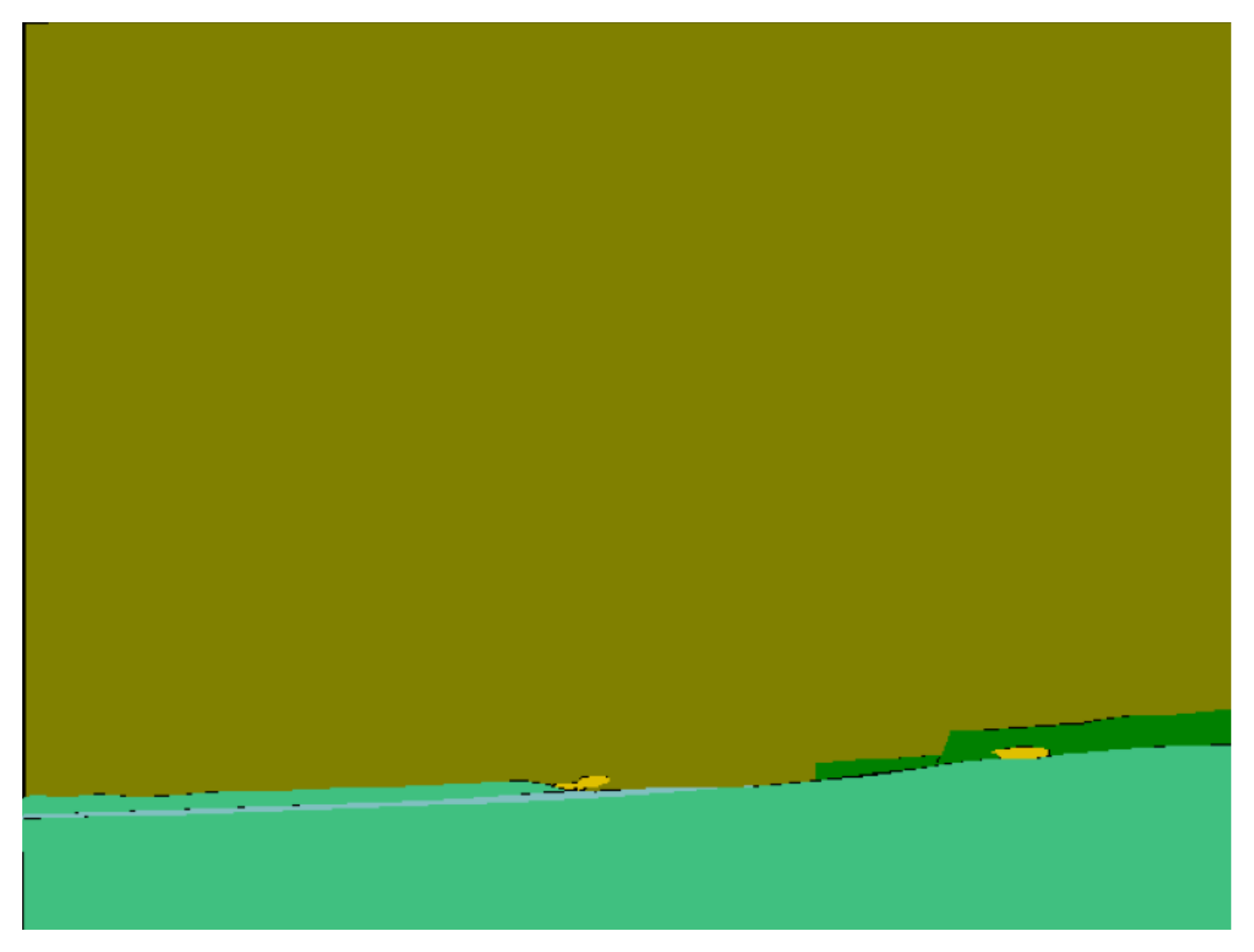}\\%
\includegraphics[height=0.27\linewidth]{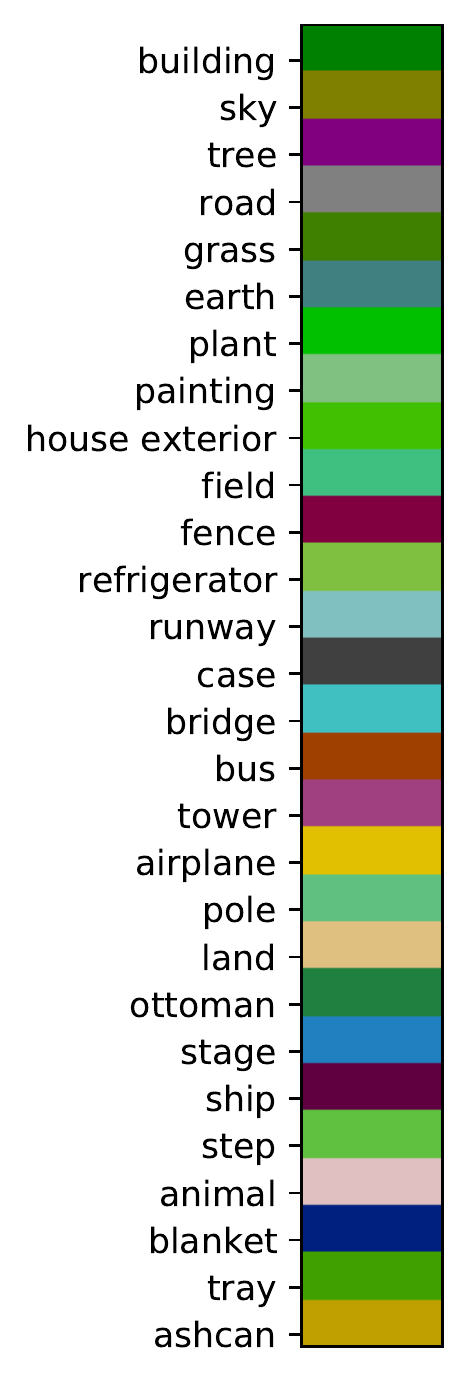}&%
\includegraphics[height=0.27\linewidth]{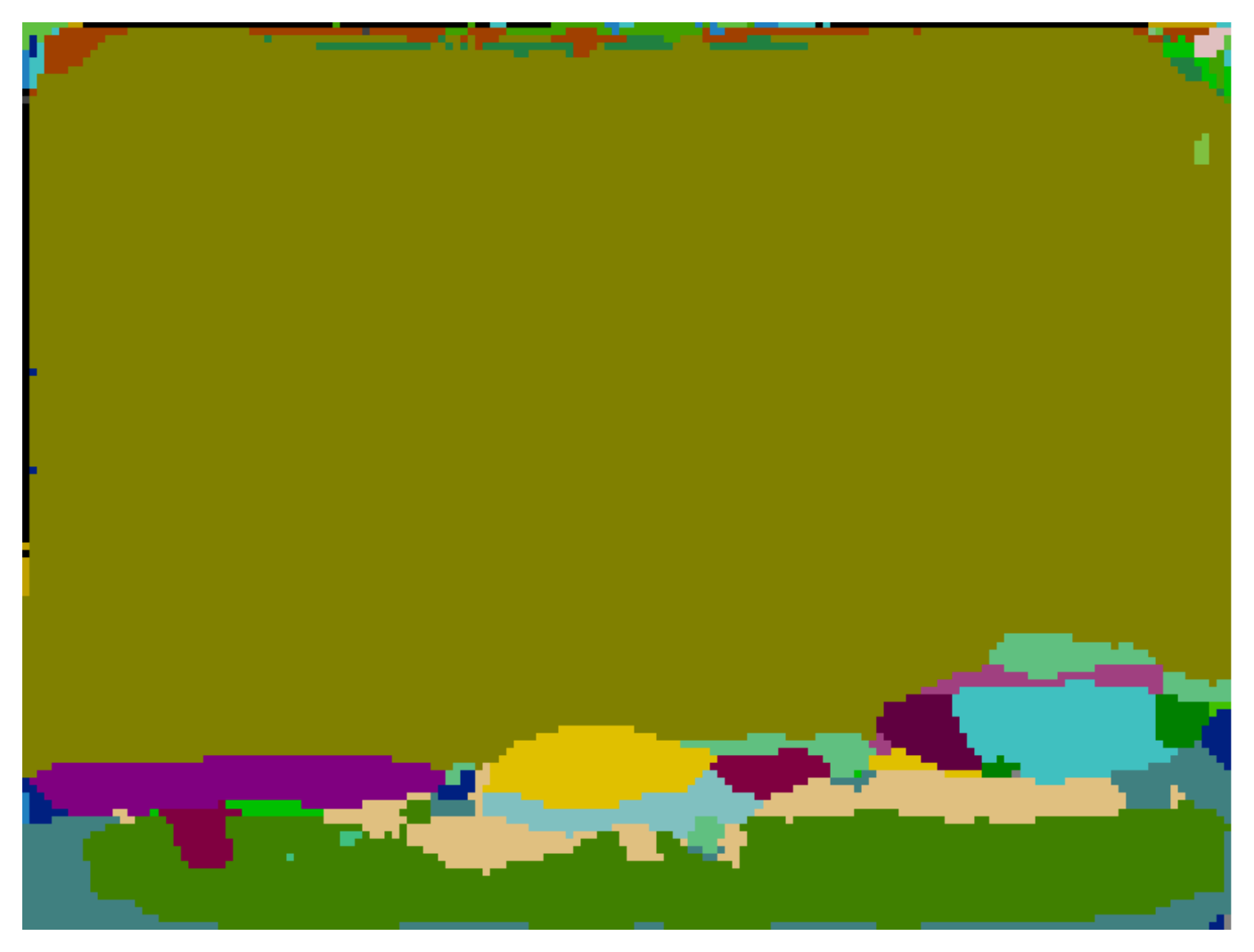}&
\includegraphics[height=0.26\linewidth]{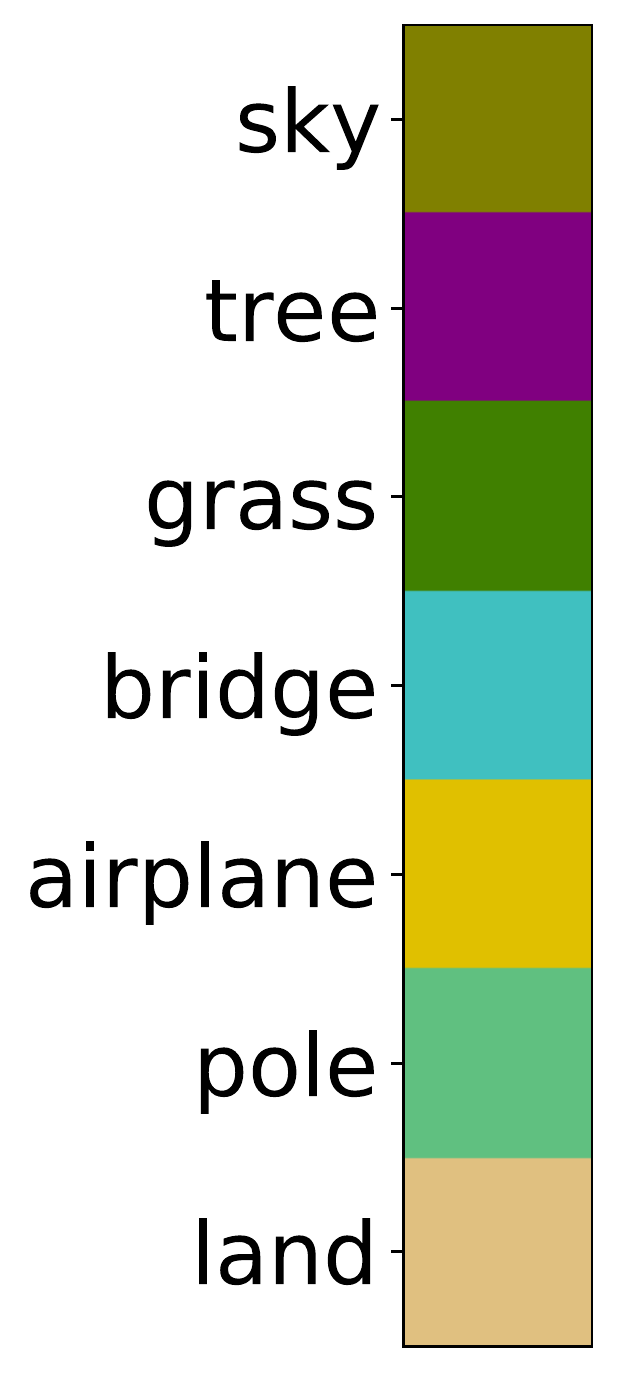}&%
\includegraphics[height=0.27\linewidth]{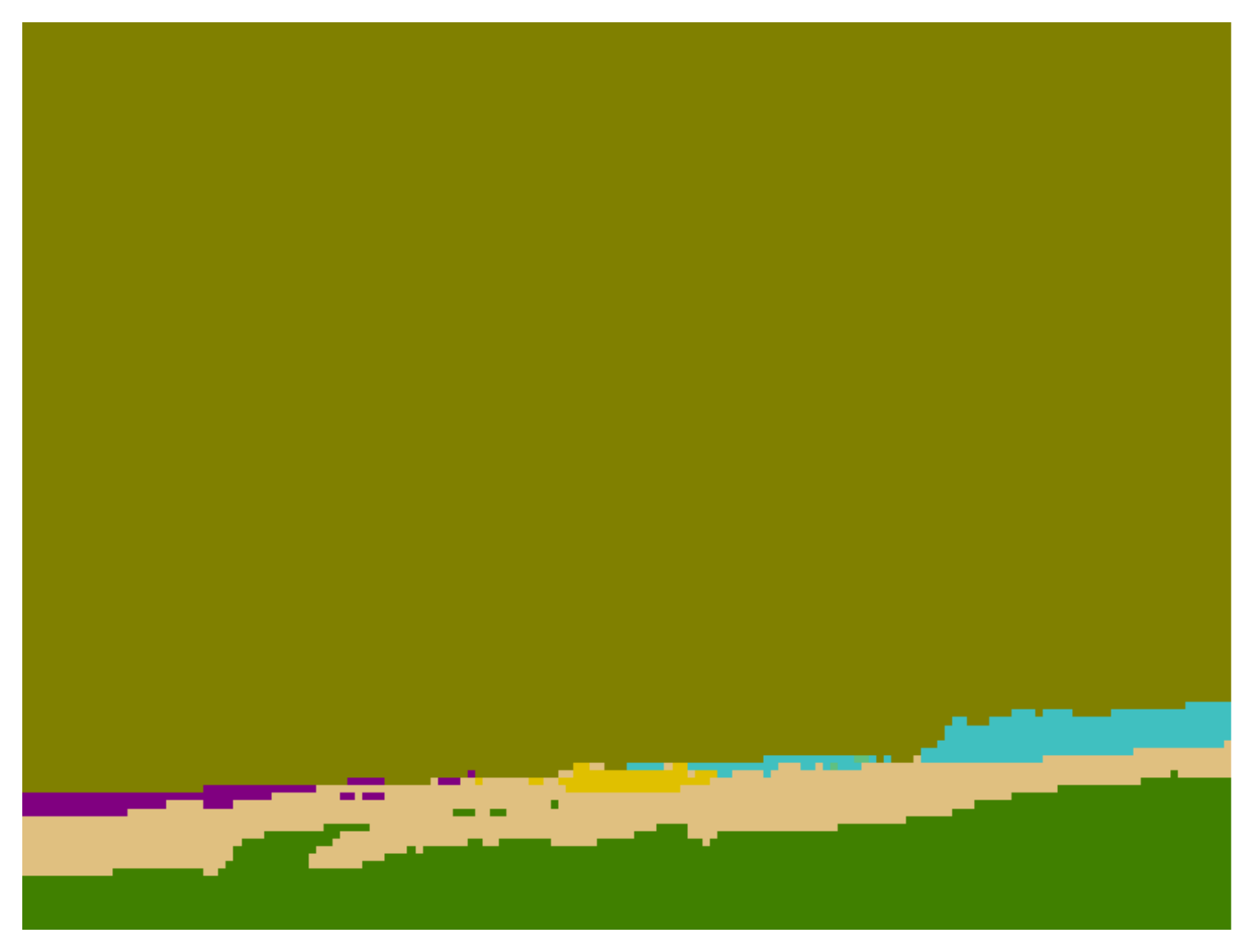}\\%
\end{tabular}
\caption{\textbf{Predictions of \ours~on random examples in the \ade~dataset (Part3).} 
}
\label{fig:ade20k_examples_3}
\end{figure*}

\begin{figure*}%
\setlength\tabcolsep{0.5pt}
\renewcommand{\arraystretch}{0.2}
\centering%
\begin{tabular}{rlrl}%
\centering%
&%
\includegraphics[height=0.27\linewidth]{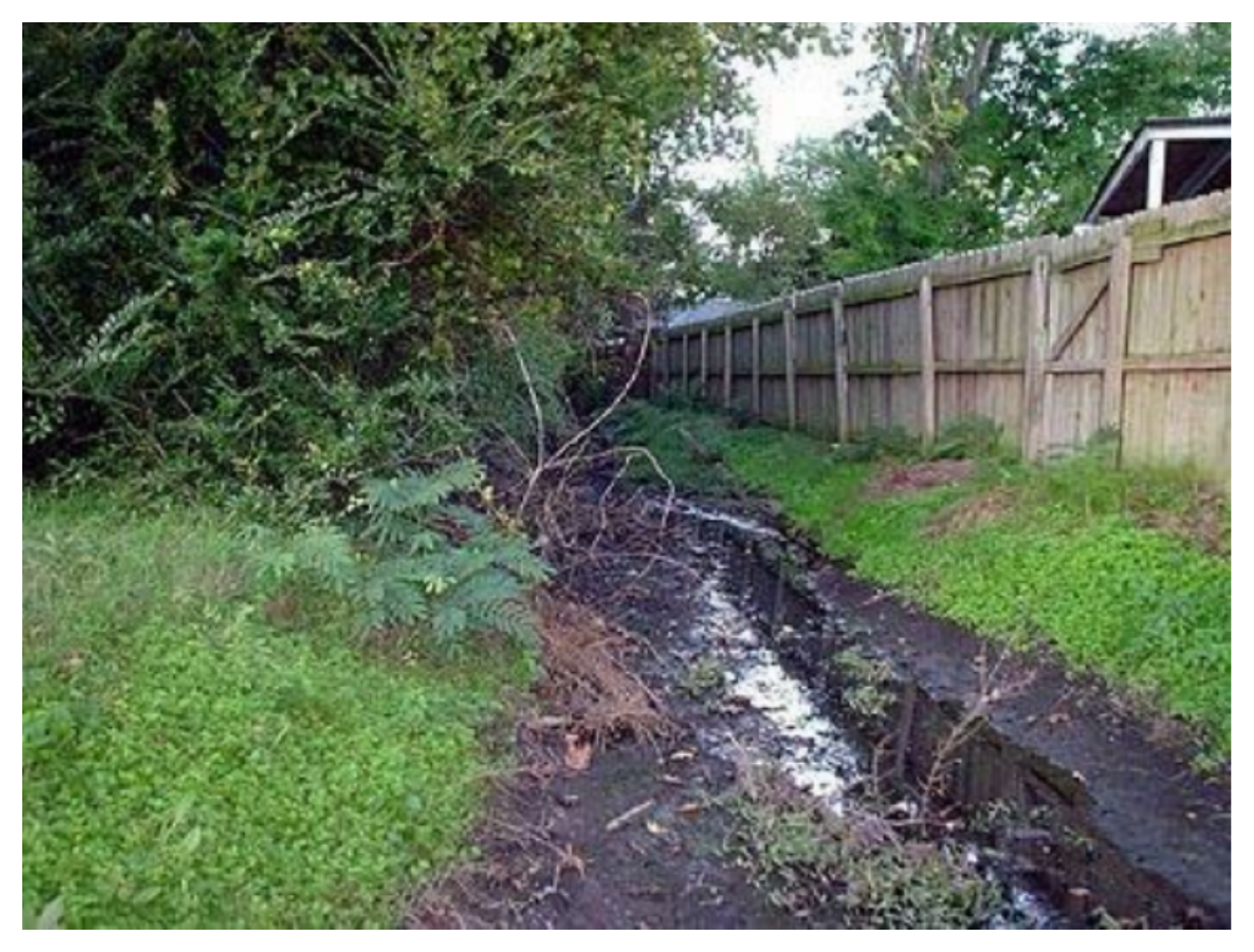}&%
\includegraphics[height=0.23\linewidth]{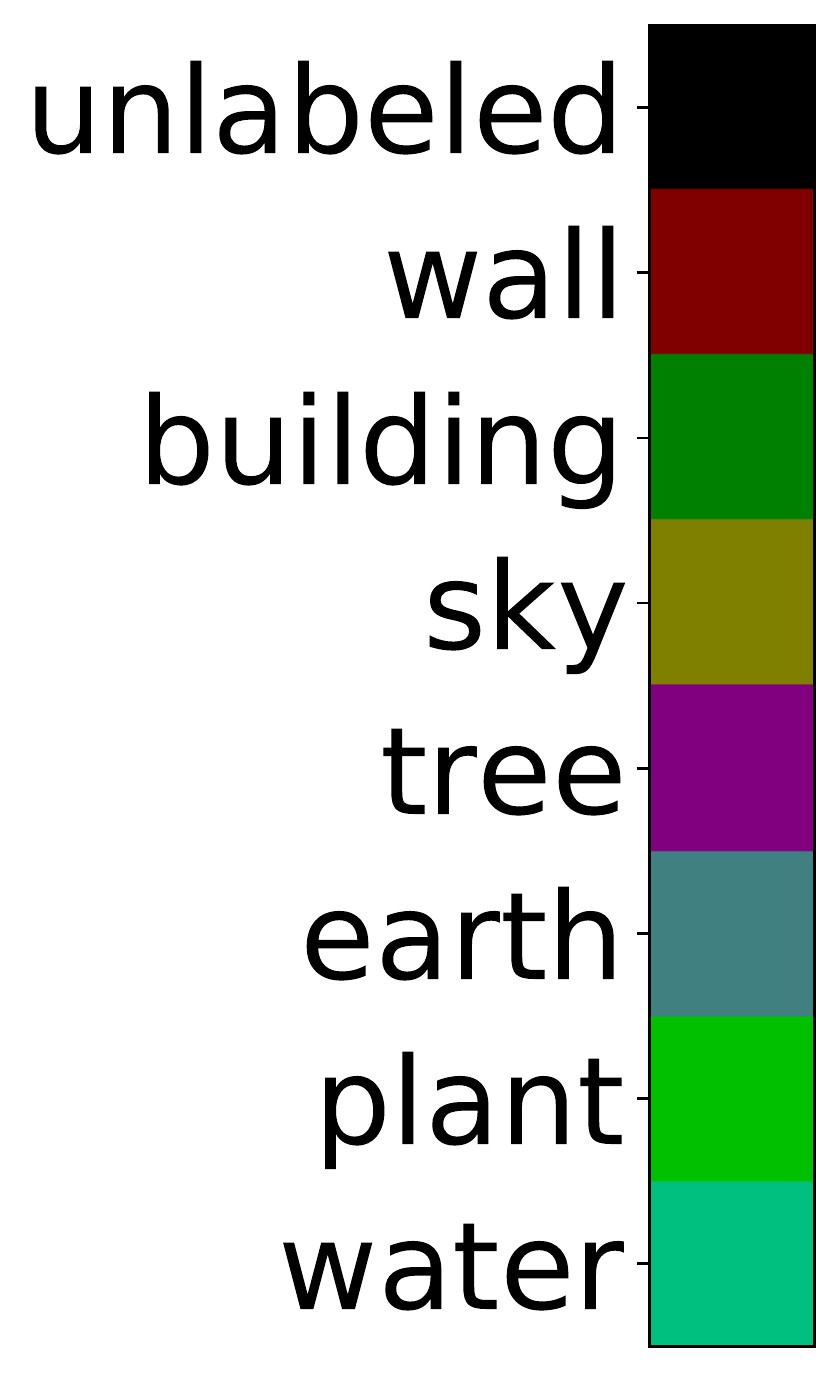}&%
\includegraphics[height=0.27\linewidth]{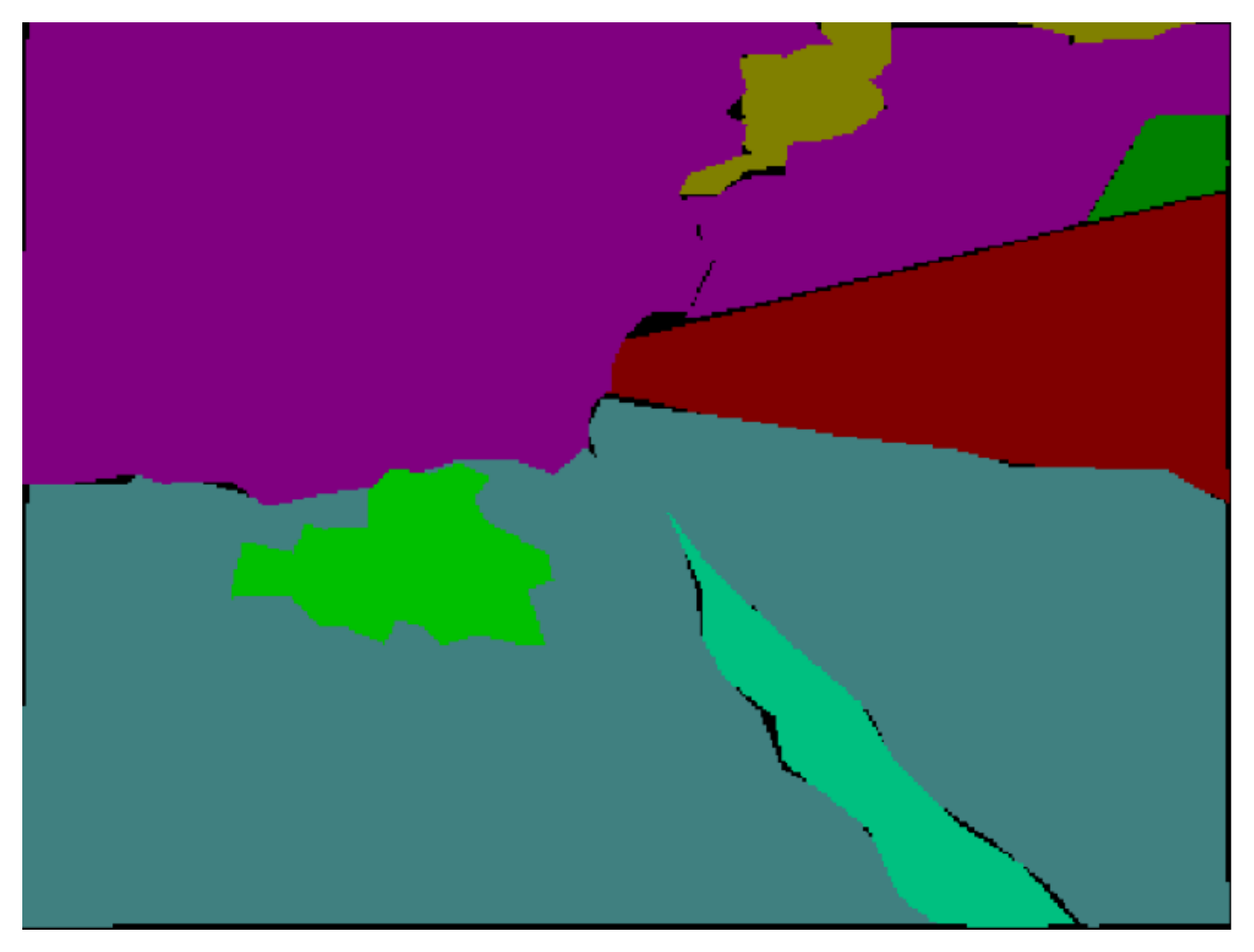}\\%
\includegraphics[height=0.27\linewidth]{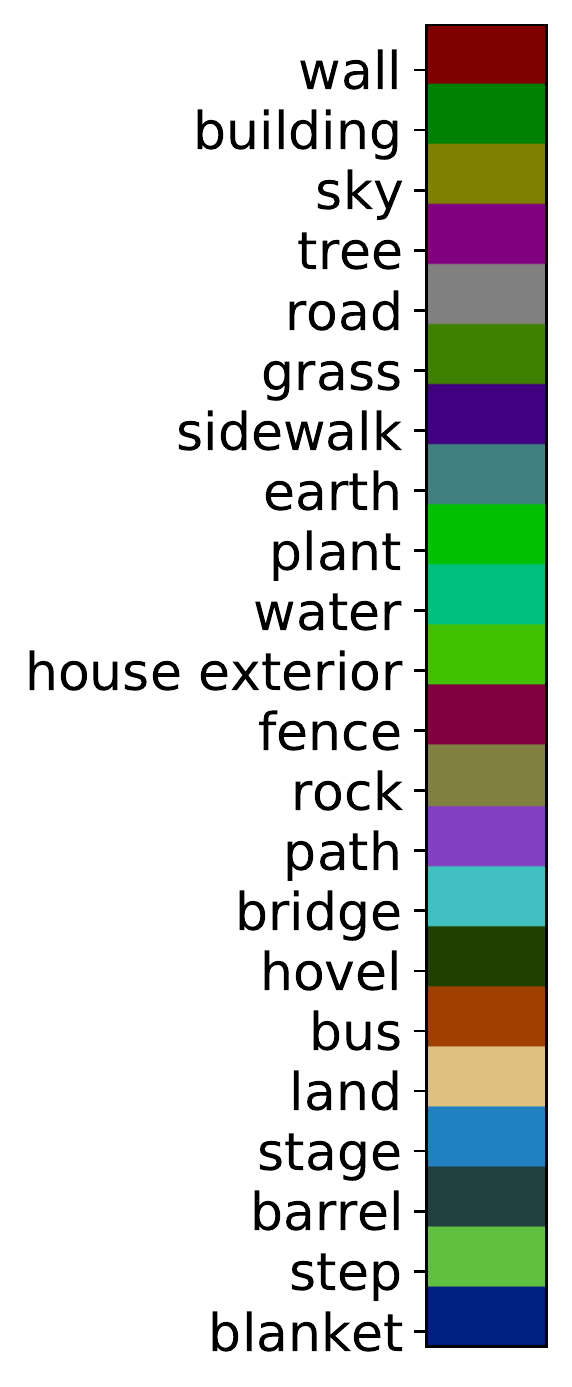}&%
\includegraphics[height=0.27\linewidth]{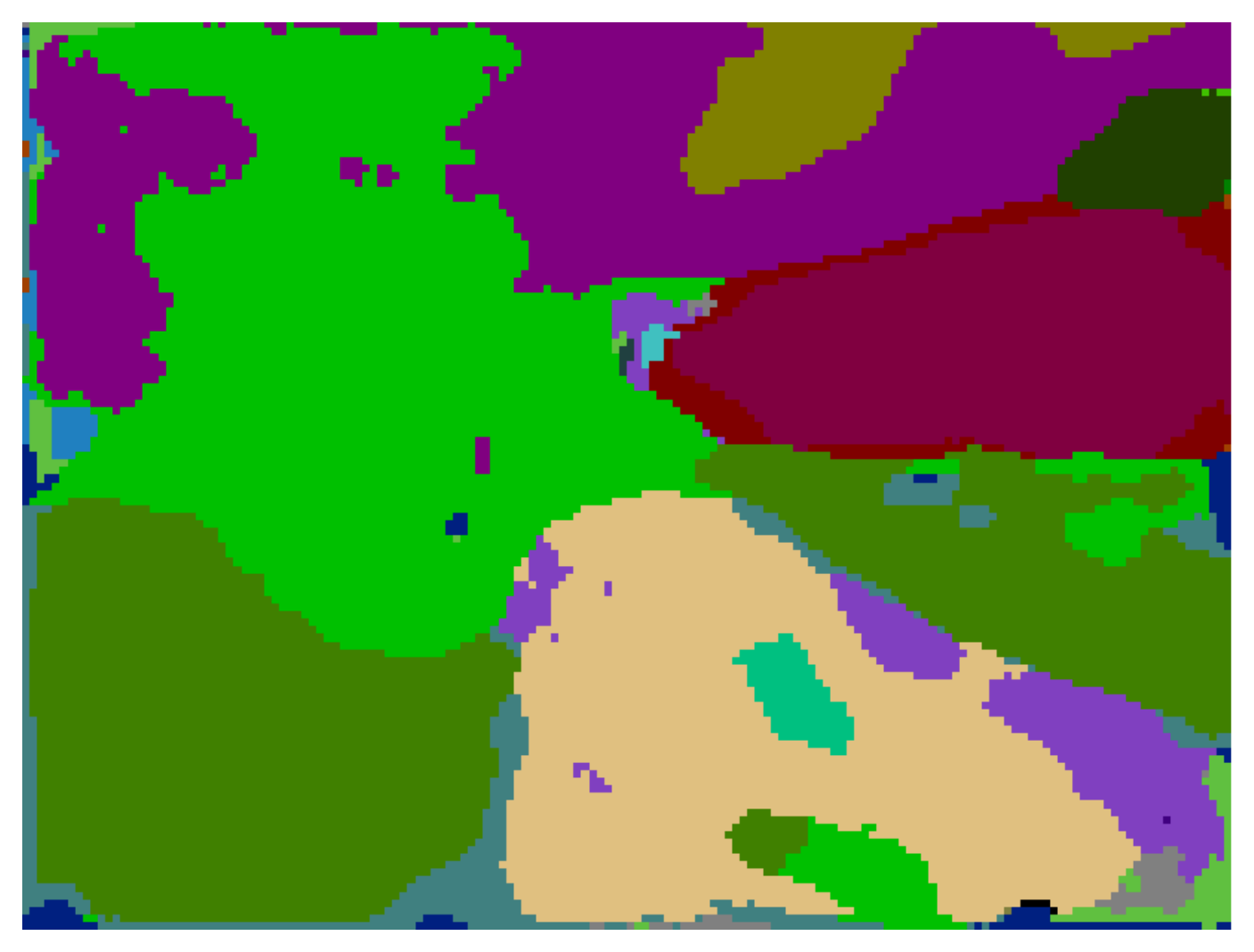}&
\includegraphics[height=0.27\linewidth]{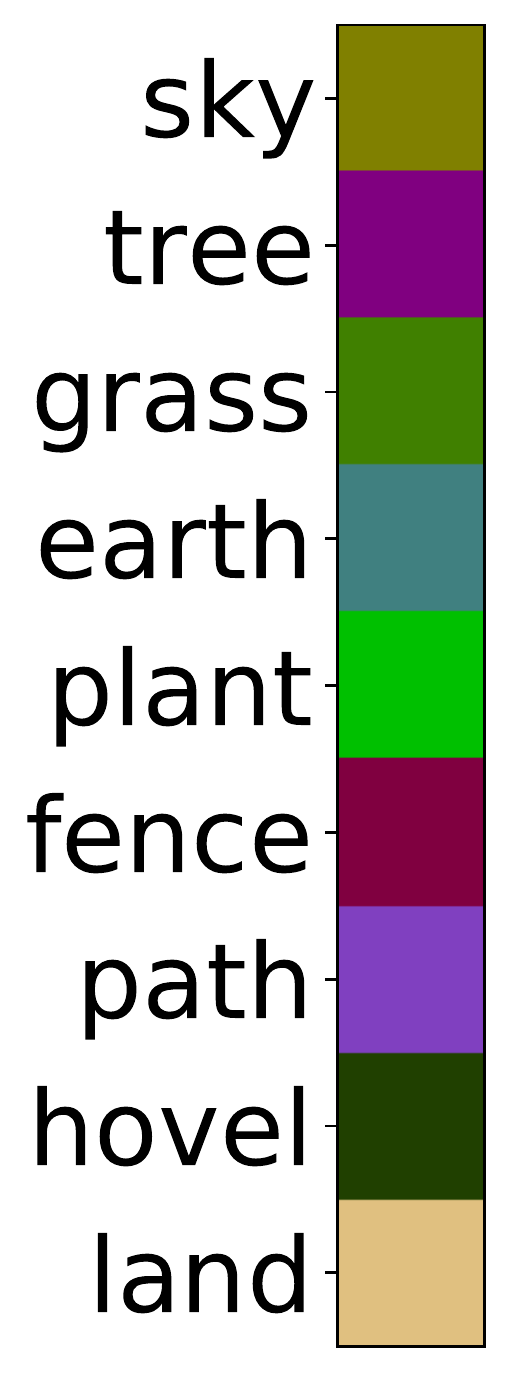}&%
\includegraphics[height=0.27\linewidth]{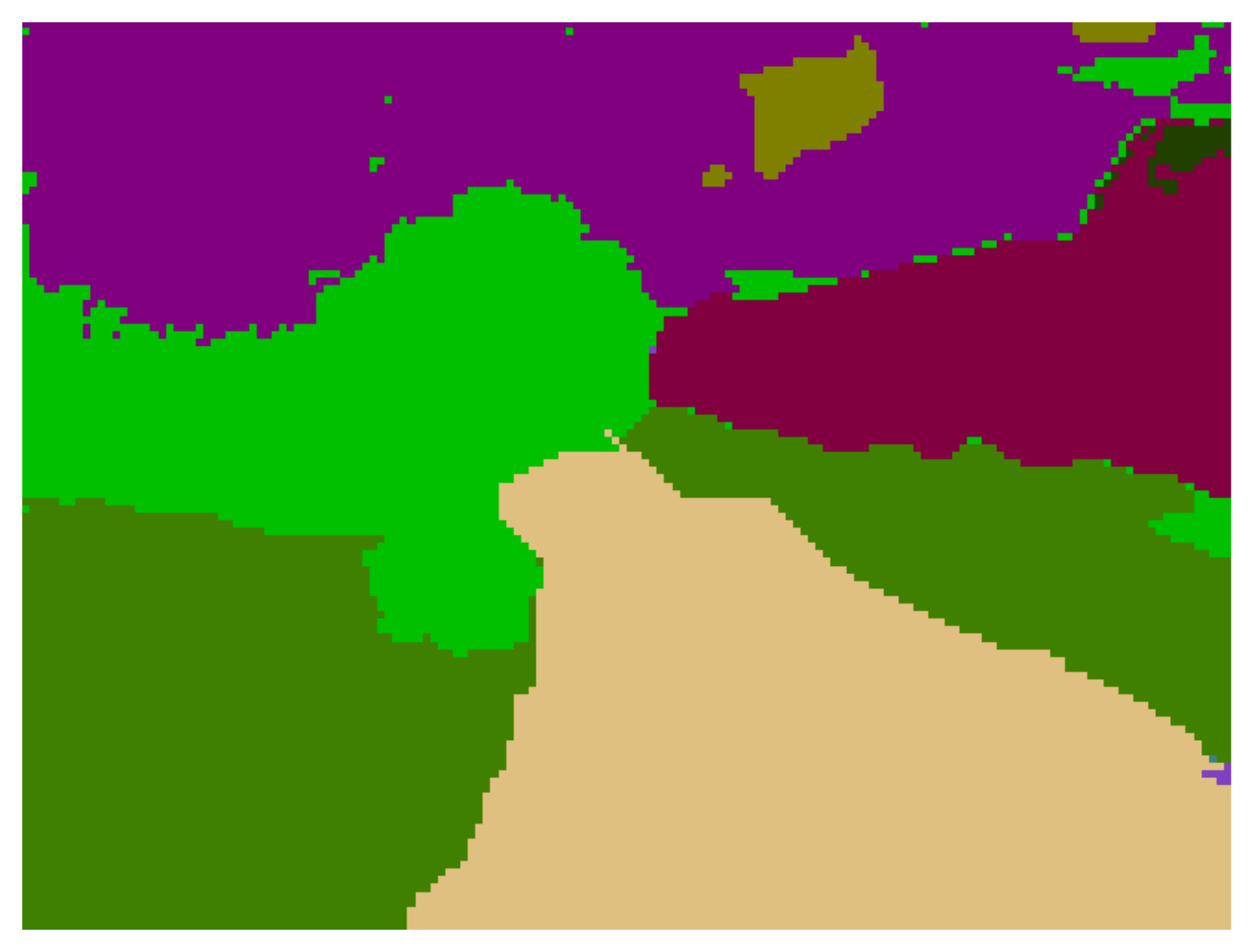}\\%
\vspace{1cm}\\
&%
\includegraphics[height=0.27\linewidth]{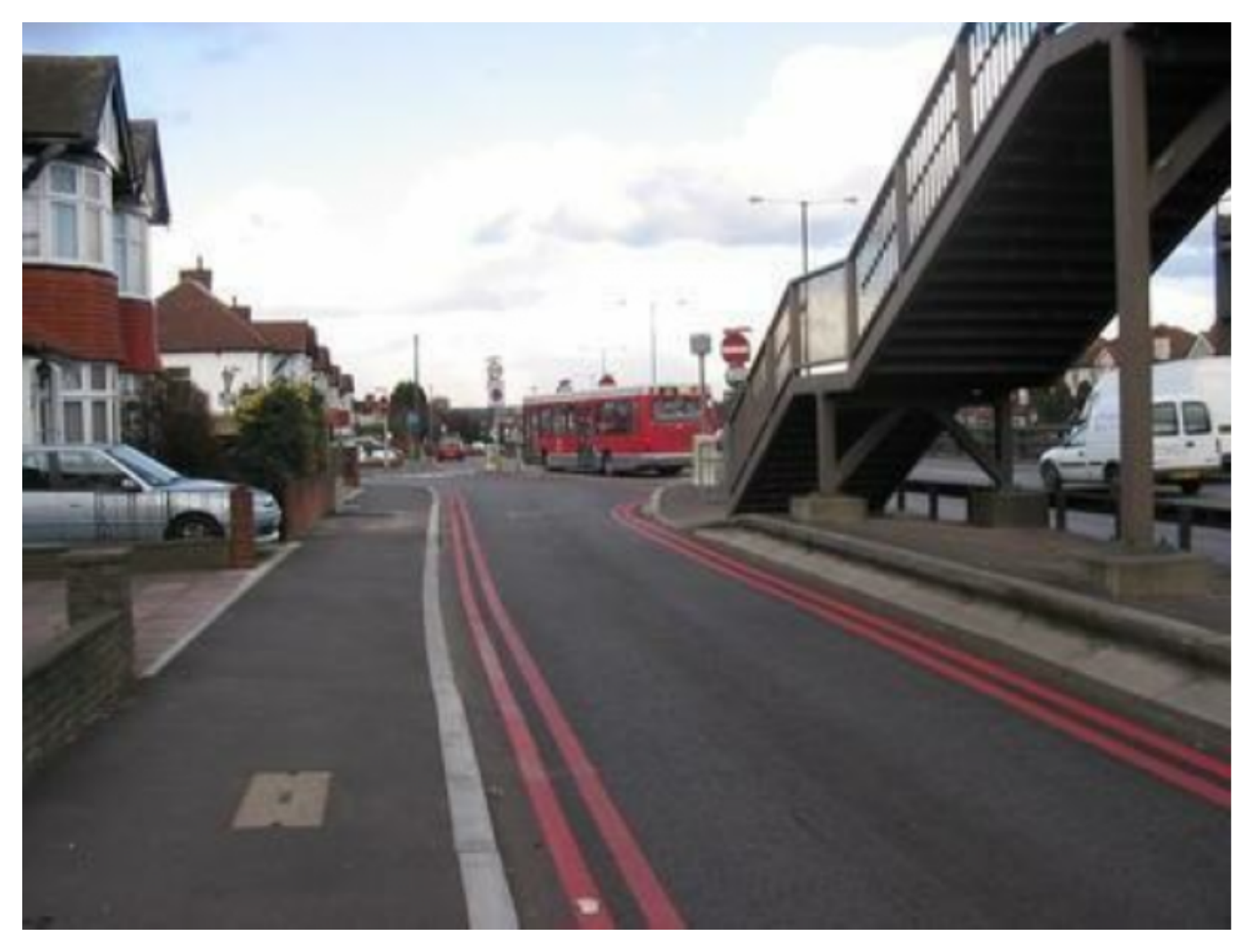}&%
\includegraphics[height=0.25\linewidth]{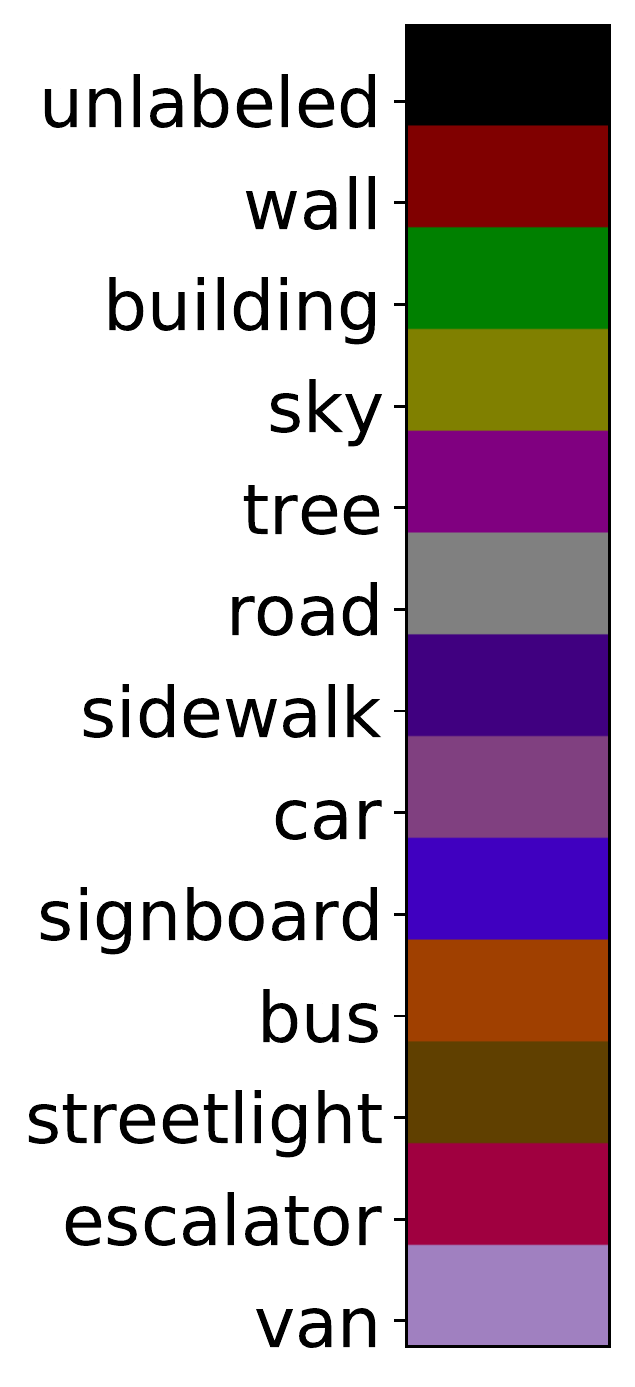}&%
\includegraphics[height=0.27\linewidth]{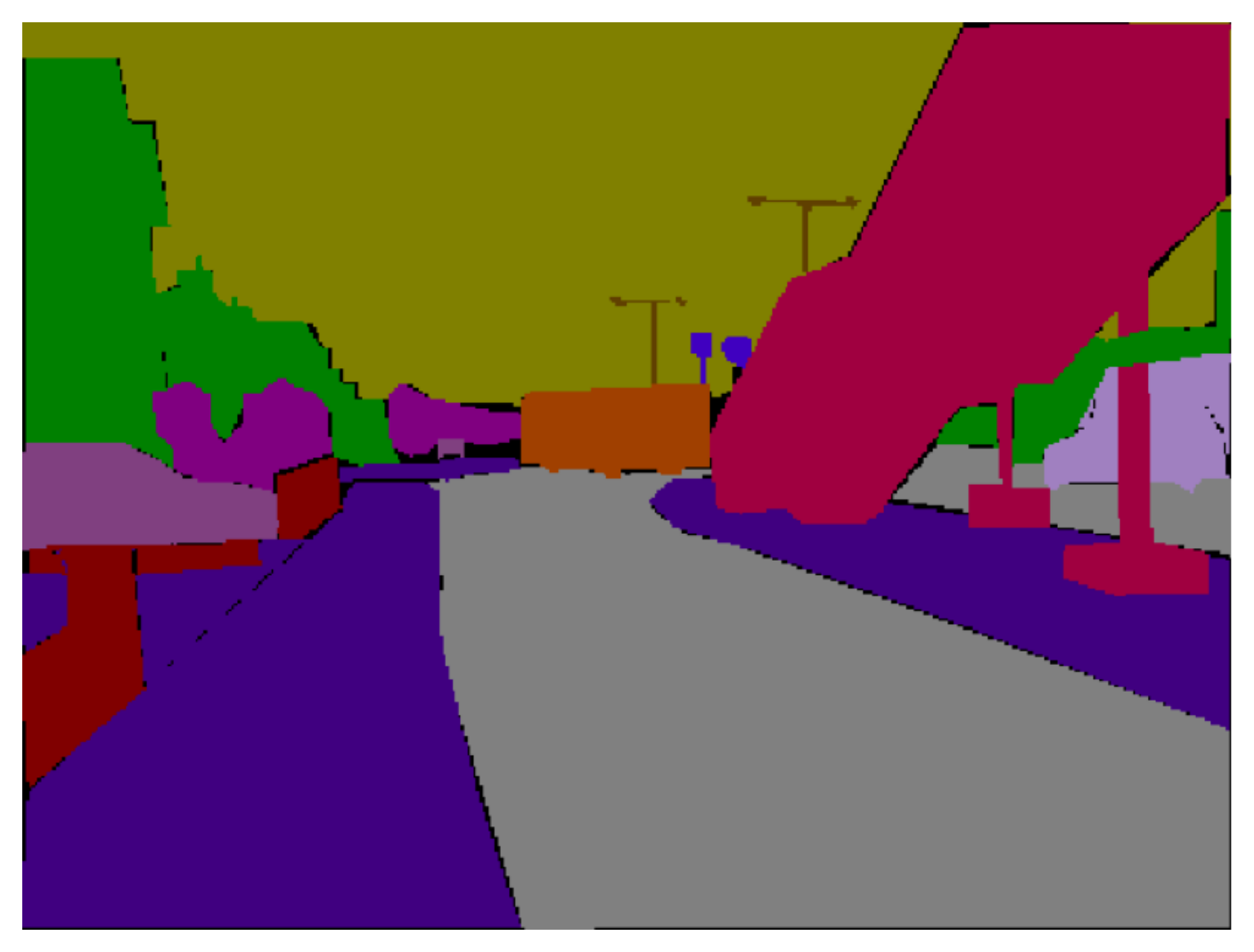}\\%
\includegraphics[height=0.27\linewidth]{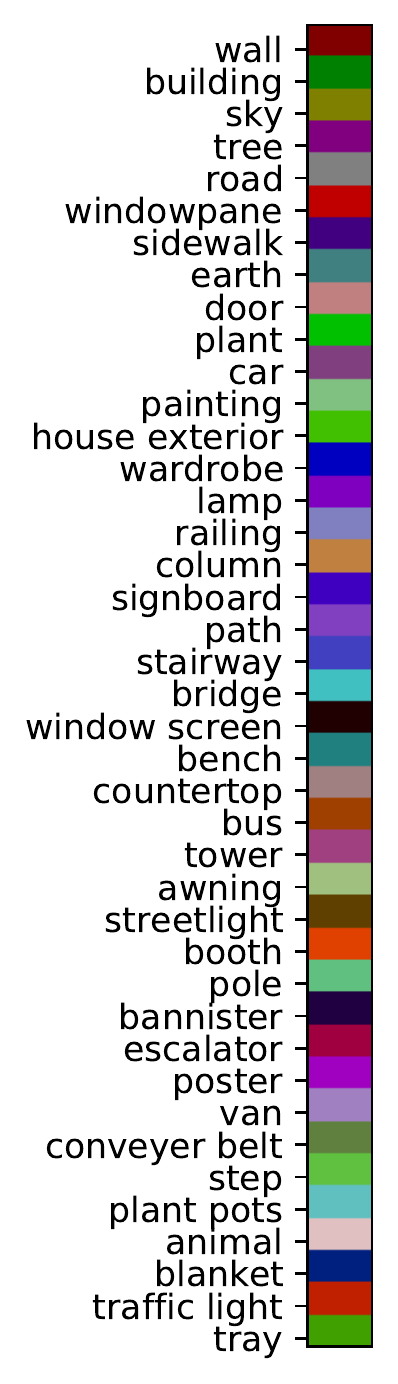}&%
\includegraphics[height=0.27\linewidth]{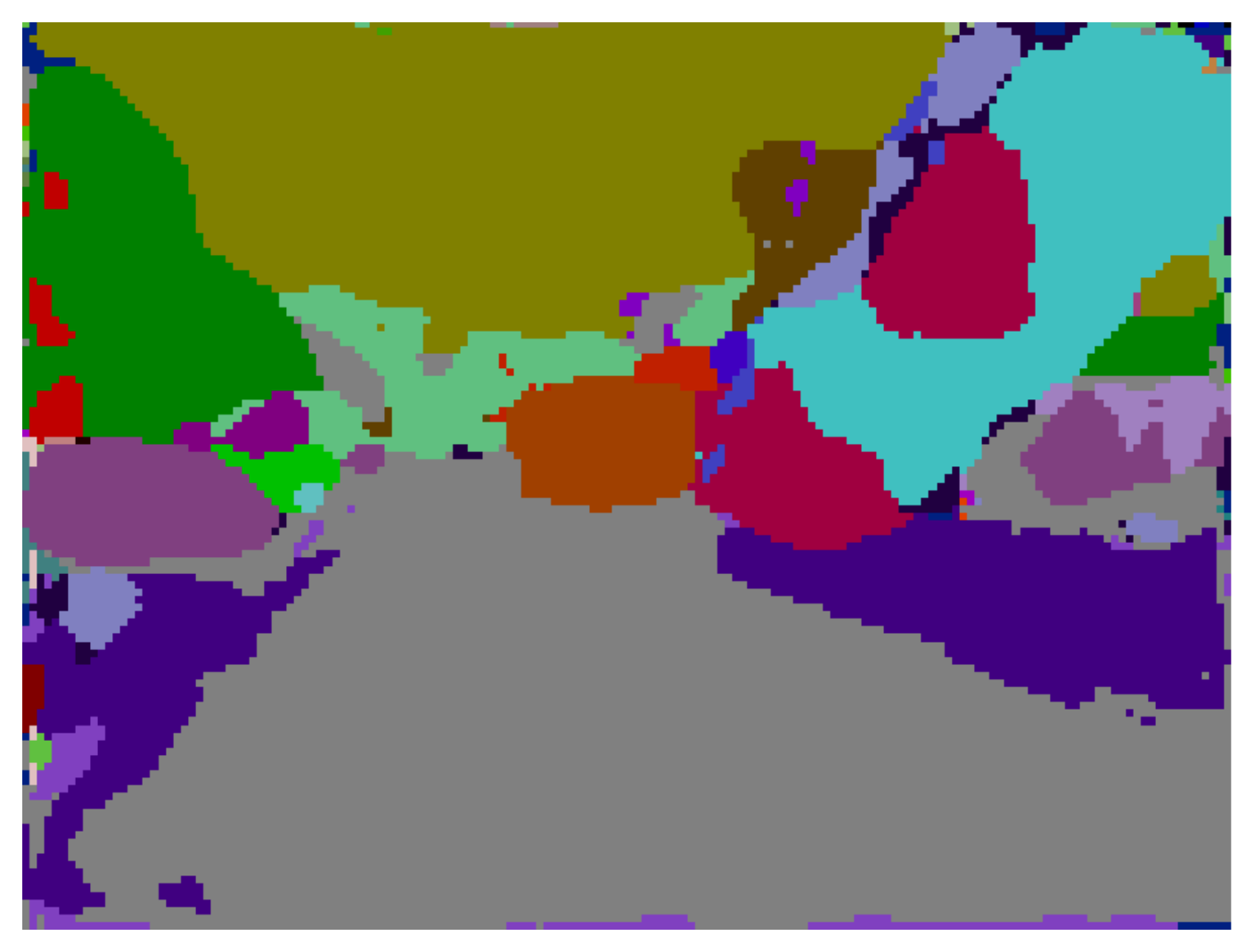}&
\includegraphics[height=0.25\linewidth]{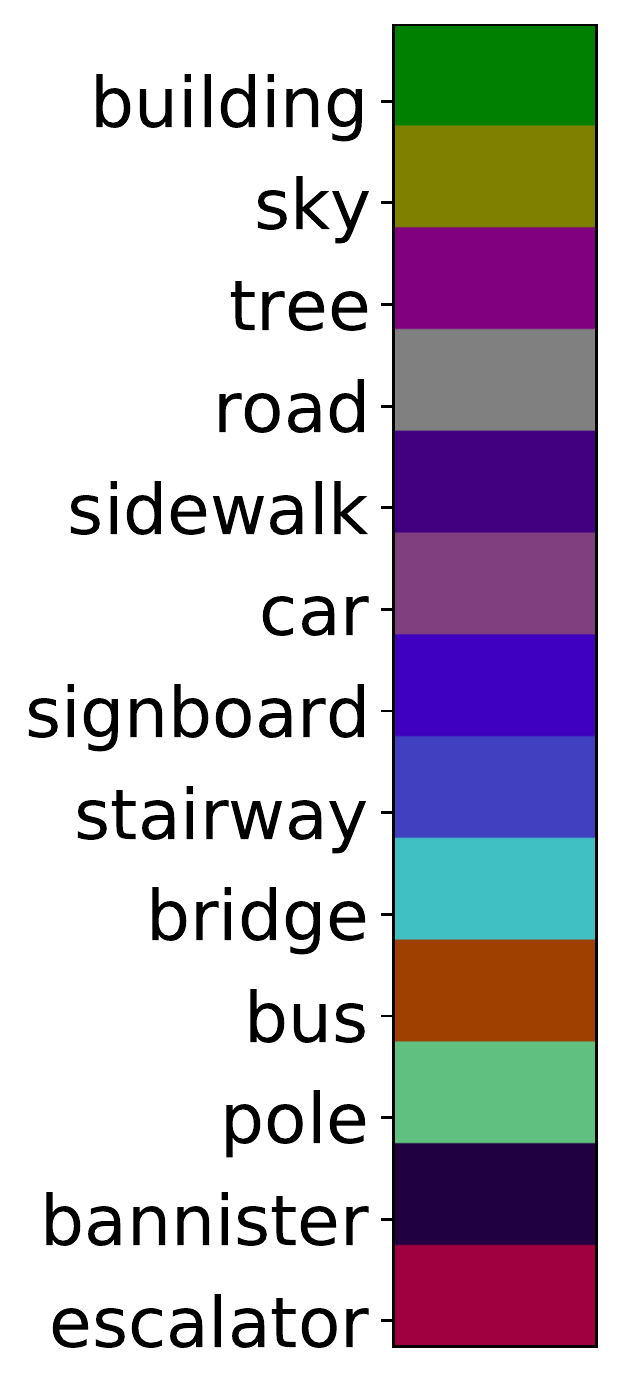}&%
\includegraphics[height=0.27\linewidth]{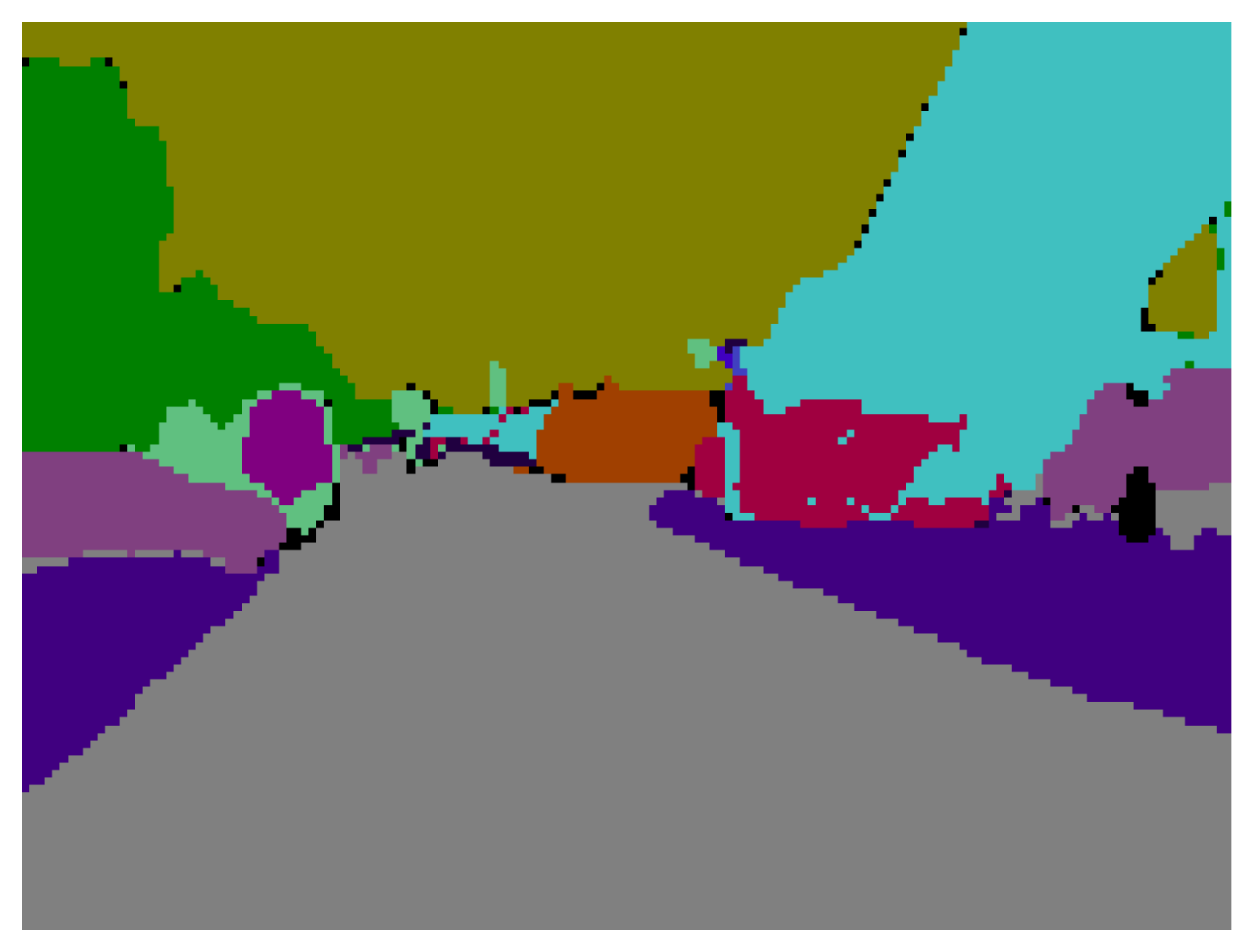}\\%
\end{tabular}
\caption{\textbf{Predictions of \ours~on random examples in the \ade~dataset (Part4).}
}
\label{fig:ade20k_examples_4}
\end{figure*}

In Figures~\ref{fig:ade20k_examples_1}-\ref{fig:ade20k_examples_4}, we present the predictions of \ours~on random images in the \ade~dataset, where the list of dataset categories are used as the query. For each image, we visualize the output of \ours{}. We also show the per-pixel prediction without incorporating the mask proposals (see Section~\ref{sec:proposals_at_inference}) for comparison.




\end{appendix}

\end{document}